\documentclass[twoside,11pt]{article}

\usepackage{jmlr2e}
\usepackage{booktabs} % For formal tables
\usepackage{subcaption}
\usepackage{adjustbox}
\usepackage{tabularx,ragged2e,booktabs}
\usepackage{amsmath}
\usepackage[most]{tcolorbox}
\usepackage{amssymb}
\usepackage{array,multirow}
\usepackage{graphicx}
\usepackage[colorinlistoftodos]{todonotes}
\usepackage[graphicx]{realboxes}
\usepackage{tikz}
\usetikzlibrary{positioning,fit,calc}
\usetikzlibrary{shadows}
\usepackage{mathrsfs}
\usepackage{graphicx}
\usepackage{breqn}
\usepackage[section]{placeins}
\usepackage[a4paper,bindingoffset=0.2in,%
            left=0.75in,right=0.75in,top=1in,bottom=1in,%
            footskip=.25in]{geometry}

\newtheorem{prop}{Proposition}
\newtheorem{assumption}{Assumption}
\newtheorem{observe}{Observation}
\newtheorem{prob}{Problem}

\ShortHeadings{GANs: The Progress So Far}{Manisha and Gujar}
\firstpageno{1}

\begin{document}

\title{Generative Adversarial Networks (GANs): The Progress So Far In Image Generation}

\author{\name Padala Manisha \email manisha.padala@research.iiit.ac.in \\
       \addr Machine Learning Lab\\
       International Institute of Information Technology\\
       Hyderabad, \\Gachibowli, Hyderabad, India 500032
       \AND
       \name Sujit Gujar \email sujit.gujar@iiit.ac.in \\
       \addr Machine Learning Lab \\ 
       International Institute of Information Technology\\
       Hyderabad, \\Gachibowli, Hyderabad, India 500032}

\editor{}%Kevin Murphy and Bernhard Sch{\"o}lkopf}

\maketitle

\begin{abstract}
In recent years, Generative Adversarial Networks (GANs) have received significant attention from the research community. With a straightforward implementation and outstanding results, GANs have been used for numerous applications. Despite the success, GANs lack a proper theoretical explanation. These models suffer from issues like mode collapse, non-convergence, and instability during training. To address these issues, researchers have proposed theoretically rigorous frameworks inspired by varied fields of Game theory, Statistical theory, Dynamical systems, etc. 

In this paper, we propose to give an appropriate structure to study these contributions systematically. We essentially categorize the papers based on the issues they raise and the kind of novelty they introduce to address them.  Besides, we provide insight into how each of the discussed articles solves the concerned problems. We compare and contrast different results and put forth a summary of theoretical contributions about GANs with focus on image/visual applications. We expect this summary paper to give a bird's eye view to a person wishing to understand the theoretical progress in GANs so far.
\end{abstract}

\begin{keywords}
  Generative Adversarial Networks, Neural Networks, Two-Player Zero Sum Games, Generative Models, Divergence Minimization  
\end{keywords}

\section{Introduction}

Generative Adversarial Networks (GANs) \citep{ganGoodfellow} are generative neural network models which aim to produce images that look like real data. Besides images, GANs can be trained to generate text, speech, or any sequences, although we here restrict our discussion on images. Given a data-set consisting of real-world images, we assume it follows a distribution referred to as the \emph{data distribution} $p_d$. In general, any generative model aims to learn $p_d$. Broadly there are two classes of generative models.

The first class consists of explicit models like Restricted Boltzmann Machines (RBM) \citep{hinton2006} and Variational Auto-encoders (VAE) \citep{kingma2014} which use latent variables as a hidden representation of the data samples.  These models specify an explicit parameterized log-likelihood functions to represent the data. We can learn the parameters from the data. Estimating the maximum likelihood of the parameters requires integrating over the entire space of latent variables, which is intractable. Hence approximation techniques are used which may not always yield the best results. GANs, on the other hand, belong to the second class, i.e., implicit models, which do not provide any distribution but generate the images which are sampled from the distribution it learns. As mentioned in \citep{review}, it follows the likelihood-free inference approach. The paper shows that in a GAN, the probability density function is estimated using density comparison techniques. GANs generate sharper images for even highly complex data-sets like CIFAR, SVHN, etc. as compared to VAE's and RBM's. With a straightforward implementation and striking results, GANs have caught the attention of the  research community.

In the entire set-up of design and training of GANs, there are three main components: (i) its architecture, (ii) loss function, and (iii) optimization technique. The typical architecture of a GAN consists of two different neural networks: (a) a \emph{generator} and (b) a \emph{discriminator} (Figure \ref{fig:linear}).  An input to the generator is a low dimensional noise vector. It transforms the noise into a data vector which forms a potential data sample. The discriminator takes this data vector as input and assigns it a score based on how likely the data vector is from the original data distribution. The data, sampled from both the real distribution and from the generator, is used to train the discriminator. Based on the score, the generator learns how to produce vectors such that the discriminator is confused. The loss function is a min-max loss where the generator minimizes the loss which the discriminator tries to maximize. In the landmark paper, the authors model the problem as a two-player zero-sum game \citep{ganGoodfellow}. They prove the convergence of this loss to the saddle point where $p_d = p_g$ in the space of distribution, i.e., on assuming that the generator and discriminator have infinite capacity. They propose to use \emph{simultaneous gradient descent} to optimize over the loss.

The assumptions made for the convergence analysis do not hold in practice. It is widely observed that these models tend to learn only a single mode of data distribution. This major challenge faced is referred to as \emph{mode collapse}. As a result of this phenomenon, the generator produces the same set of images every time. Although the images are sharp, they lack variation. Another major challenge is that the training of GANs is highly \emph{unstable} and the loss plots obtained during training are not indicative of convergence. These practical issues show the lack of sound theoretical understanding of GANs. It can happen that while training a GAN, though the loss is diverging, the network is producing realistic images and there is no explanation for this. It has resulted in plethora of research papers around GAN set-up, especially to address the above two issues. %Our goal is to give a birds eye view of this growing field.

\bigskip
\noindent\emph{Our Contributions.} In this paper, our goal is to give a framework to the discussions in the  recent papers.  We hope to give the reader an overview of the vital research so far and at the same time give the progress a structure to simplify further analysis. Researchers have rigorously analyzed the models in various lights obtaining concepts from fields of game theory, statistical learning theory, optimization, online learning, dynamical system, etc. In this work, we focus only on theoretical contributions and summarize the recent work based on the issues each address and novel approaches proposed. A related work by Hitawala \citep{hitawala} also presents a comparative study of models with different modifications over vanilla GANs. The author does not focus on the issues related to the model and its training. Hence the models the author chooses to discuss are entirely different from ours. 

\bigskip
\noindent\emph{Organization.} We now discuss the layout for the rest of the paper. We begin with a technical description of GAN set-up in Section \ref{sec:set_up}, followed by some of the success stories in Section \ref{sec:success}. In Section \ref{sec:challenges}, we explain the two practical challenges that the model faces: 

\bigskip
\begin{tcolorbox}[colback=gray!10!white,colframe=black!75!black]
\begin{itemize}
    \item[C$_1$] Mode collapse
    \item[C$_2$] Non-convergence and instability
\end{itemize}
\end{tcolorbox}

\bigskip

We identify certain questions that one must address in order to completely solve the above two challenges (Section \ref{sec:challenges}). In the literature, the researchers used mainly the following four types of solution techniques to resolve the questions. 
\bigskip
\begin{itemize}
    \item[S$_1$] \emph{Modify the loss function}: There are papers which modify the loss based on heuristics and also raise a fundamental issue of \emph{vanishing gradients} which lead to non-convergence in training. We further categorize the papers which modify the loss for the specific motives as listed below,
    \begin{itemize}
        \item[i.] Resolve vanishing gradient problem
        \item[ii.]Regularization for vanishing gradients
        \item[iii.]Regularization for non-convergence
        \item[iv.] Overcome the problem of Biased Gradient Estimator
        \item[v.] Resolve non-convergence
    \end{itemize}
    \item[S$_2$] \emph{Modify the architecture}: There are many papers which primarily propose an architectural change. These changes sometimes lead to change in the loss function too. We list the different ways in which these changes have been proposed. 
    \begin{itemize}
        \item[i.] Auto-encoder based architectures
        \item[ii.] Using a mix of discriminators or generators (Ensemble method)
        \item[iii.] Introduce memory within the network
    \end{itemize}{}
    \item[S$_3$] \emph{Modify the optimizer}: There are some papers which suggest to use a different optimizer other than gradient descent or modify the optimizer backed with the rigorous theoretical analysis.
    \item[S$_4$] \emph{Provide convergence and equilibrium analysis}: Theoreticians published papers that do not aim to propose solutions but rather build a theoretical framework to explain the convergence of loss, instability in training, and mode collapse. These papers also give generalization bounds for sample complexity. The papers are subdivided based on questions (given in Section \ref{sec:challenges}) they address
    \begin{itemize}
        \item[i.] Addressing Q$_1$
        \item[ii.] Addressing Q$_3$
        \item[iii.] Addressing Q$_3$
    \end{itemize}{}
\end{itemize}

In Section \ref{sec:novel}, we discuss each of the papers in further depth. For a better organization, we categorize the papers based on the novelty that they introduce as listed above. Mostly, papers make changes to one of the primary components of GAN set-up which we discussed above. According to our observation, there are papers which change the same component with differing motivations. In Section \ref{sec:vis_summary}, we compare the images generated by approaches followed in few of the papers. In Section \ref{ssec:tab_summary}, we summarize papers in tabular form \footnote{Disclaimer: The list is not exhaustive. The papers which we are describing are chosen based on our understanding of
the importance of the contribution in the article.} based on the issues they raise and address. We conclude the paper in Section \ref{sec:con}. In summary, the paper is organized as follows. 
%%% %%%%%%%%%%%%%%%%%%%%%%%%%%%%%%%%%%%
%%% Sujit Added for time being%%%%%%%%%
%%% %%%%%%%%%%%%%%%%%%%%%%%%%%%%%%%%%%%

\tableofcontents

% I wish you the best of success.

 

%%%%%%%%%%%%%%%%%%%%%%%%%%%%%%%%%%%%%%%%%%%%%%%%%%%%%%%%%%%%%%%%%%%%%%%%%%%%%%%%%%%%%%%%%%%%%%%%%%%%%%%%%%%

\section{Primary GAN Set-Up}
\label{sec:set_up}
In this section we provide the key elements of GANs as proposed in \citep{ganGoodfellow}. It includes the typical architecture\footnote{architecture and model are used interchangebly} used, the loss function and the optimizer followed by the convergence analysis. Before we begin, we list the notations that we would be using henceforth.
\subsection{Notation}
Given below in Table \ref{tab:not} is a list of the basic notations related to GANs .

%\begin{tcolorbox}[colback=gray!03!white,colframe=black!75!black]

\begin{table}[!htb]
    \centering
\begin{tabular}{ c|c }
\hline
 Symbol & Parameter \\ 
  \hline
  $D$ & Discriminator \\
  $G$ & Generator \\
  $D_{\theta}$ & $D$ parameterized by $\theta$\\
  $G_{\phi}$ & $G$ parameterized by $\phi$ \\ 
  $p_{d}$ & Data distribution \\
  $p_{g}$ & Model($G$) distribution\\
  $p_{z}(z)$ & Random noise distribution\\
  $x \sim p_{d}$ & real data sample \\
  $\hat{x} \sim p_{g}$ & generated sample \\
  $V$ & Utility function for $D$ Equation (\ref{eq:gan_loss1}) \\
  $U$ & Utility function for $G$ \\
 \hline
\end{tabular}
\caption{Notation} \label{tab:not} 
\end{table}

%\end{tcolorbox}

We list some useful definitions which are used later.
\begin{definition}[KL Divergence]
Given two probability distributions $p_d(x), p_g(x)$ with positive support $\forall \ x$. 
The kl divergence is given by,
\begin{equation}
    \label{eq:kl}
    KL(p_d \parallel p_g) = \int_{-\infty}^{\infty} p_d(x) \log \frac{p_d(x)}{p_g(x)} dx 
\end{equation}{}

\end{definition}{}
Note that in the above defintion the difference between data distrbution $p_d$ and model distribution $p_g$ is weighted by $p_d$. KL divergence is not symmetric hence we also define the following.
\begin{definition}[Reverse KL Divergence]
Given two probability distributions $p_d(x), p_g(x)$ with positive support $\forall \ x$. 
The reverse kl divergence is given by,
\begin{equation}
    \label{eq:rvkl}
    KL(p_g \parallel p_d) = \int_{-\infty}^{\infty} p_g(x) \log \frac{p_g(x)}{p_d(x)} dx
\end{equation}{}

\end{definition}{}
According to \citep{ganGoodfellow}, the following divergence is minimized during the training of GAN,
\begin{definition}[Jenson Shannon Divergence (JSD)]
The Jenson Shannon Divergence is a symmetric distance metric between the two distribution $p_d(x), p_g(x)$ given by,
\begin{equation}
    \label{eq:jsd}
    JSD(p_d\parallel p_g) = \frac{1}{2} KL(p_d \parallel \frac{p_d + p_g}{2}) + \frac{1}{2} KL(p_g \parallel \frac{p_d + p_g}{2})
\end{equation}{}

\end{definition}{}

We state the definition of the standard $L_p$ distance metric
\begin{definition}[$L_p$ distance]
\label{def:lp}
The $L_p$ distance between any two $n$-dimensional vectors $x = (x_1, \ldots, x_n), y = (y_1, \ldots, y_n)$ for $1 \leq p \leq \infty$ is given by,
$$\parallel x -y \parallel = (\sum_{i=1}^n |x_i - y_i|^p )^{1/p}$$
\end{definition}{}
Given a machine learning classification task with two classes, the loss between the predicted probability $p$ and the target probability $y$ is given by,
\begin{definition}[Binary Cross Entropy]
\label{def:bce}
Given $p$ is the predicted probability for one class and $y$ is the target probability for the same. 
$$BCE(p, y) = - y \log  p - (1-y)\log (1 - p)$$
\end{definition}{}

\begin{definition}[Integral Probability Metric]
\label{def:IPM}
Let $\mathscr{F}$ be a set of measurable, symmetric and bounded real valued functions on $\mathcal{X}$. Given $\mathbb{P}, \mathbb{Q} \in \mathscr{P}(\mathcal{X})$,  
    $$
d_{\mathscr{F}}(\mathbb{P}, \mathbb{Q})=\sup _{f \in \mathscr{F}}\{\underset{x \sim \mathbb{P}}{\mathbb{E}} f(x)-\underset{x \sim \mathbb{Q}}{\mathbb{E}} f(x)\}
$$

\end{definition}

%%%%%%%%%%%%%%%%%%%%%%%%%%%%%%%%%%%%%%%%%%%%%%%%%%%%%%%%%%%%%%%
\subsection{Architecture}
The model is primarily set up as a two player game, where the players are the neural networks. One player is the \emph{generator} $G$ which takes noise $z \in Z$ as the input and transforms it into a vector denoted by $\hat{x} \sim p_g(z)$. The generated $\hat{x}$ has similar dimension as that of a sample from the real data $x \sim p_d$. The generator is typically a multi-layered perceptron with transpose convolutional layers especially when generating images. It typically maps a lower dimensional noise vector $z$ to a higher dimensional data vector $\hat{x}$. 

The other player is called the \emph{discriminator} $D$. It takes as an input, a vector of dimension equivalent to the data sample i.e, $x$ or $\hat{x}$. Then the network $D$ finally gives a scalar output, which represents its confidence about the input being a real sample $x$ or a fake sample $\hat{x}$. $D$ typically is again a multi-layered perceptron with convolutional layers which at the end gives the image, a binary score.

The block diagram for the above architectures is given in Figure \ref{fig:linear}. Given the architectures of the players, the specific game set up between the two during training is further described below.

\tikzset{block/.style={draw,thick,text width=5em,minimum height=1mm,align=center}, line/.style={-latex}} \definecolor{mycolor}{rgb}{0.8,0.6,0.7} \definecolor{mycolor1}{rgb}{0.8, 0.6, 0.5} \definecolor{mycolor2}{rgb}{0.529,0.807,0.89} \definecolor{mycolor3}{rgb}{0.3, 0.5, 0.6} 
\begin{figure}[!h]
\centering 
\begin{tikzpicture} 
\node[block,fill=mycolor1, line width=0.2mm, rotate=90,drop shadow={color=mycolor1!60!black}] (a) {Input Noise ($z$)}; 
\node[block, fill=mycolor3, line width=0.2mm, minimum height=5em, minimum width=4em, right=of a, yshift = -3em,text=white,drop shadow={color=mycolor3!60!black}] (b) {Generator ($G$)}; 
\node[block, fill=mycolor2, line width=0.2mm, rotate=90, minimum height=1mm, minimum width=5em,right=of b,drop shadow={color=mycolor2!60!black}] (i) {Generated Data}; \node[block, fill=mycolor3, line width=0.2mm, minimum height=5em, minimum width=8 em, right= of i, text=white,drop shadow={color=mycolor3!60!black}] (c) {Discriminator ($D$)}; 
\node[block,fill=mycolor, line width=0.2mm, rotate = 90, right=of c, xshift=-3em,drop shadow={color=mycolor!60!black}] (d) {Accuracy}; 
\node[block, line width=0.2mm, minimum height=2em, fill=mycolor2,drop shadow={color=mycolor2!60!black}] (e) at ([yshift= 2.5cm]$(b)!0.3!(c)$) {Real Data}; 
\draw[line, line width=0.5mm] (a)-- (b); 
\draw[line, line width=0.5mm] (c)-- (d);
\draw[line, line width=0.5mm] (b)-- (i); 
\draw[line, line width=0.5mm] (i)-- (c); 
\draw[line, line width=0.5mm] (e)-- (c);
\end{tikzpicture}
\caption{\label{fig:linear} GANs Block Diagram } 
\end{figure}
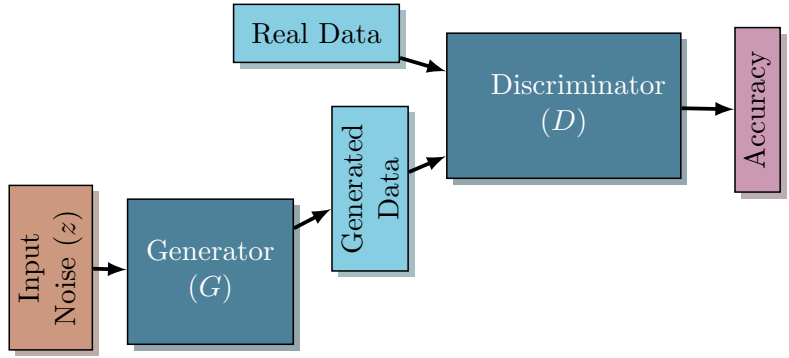

%%% No shadow
% \begin{figure}[!h]
% \centering 
% \begin{tikzpicture} 
% \node[block,fill=mycolor1, line width=0.2mm, rotate=90] (a) {Input Noise ($z$)}; 
% \node[block, fill=mycolor3, line width=0.2mm, minimum height=5em, minimum width=4em, right=of a, yshift = -3em,text=white] (b) {Generator ($G$)}; 
% \node[block, fill=mycolor2, line width=0.2mm, rotate=90, minimum height=1mm, minimum width=5em,right=of b] (i) {Generated Data}; \node[block, fill=mycolor3, line width=0.2mm, minimum height=5em, minimum width=8 em, right= of i, text=white] (c) {Discriminator ($D$)}; 
% \node[block,fill=mycolor, line width=0.2mm, rotate = 90, right=of c, xshift=-3em] (d) {Accuracy}; 
% \node[block, line width=0.2mm, minimum height=2em, fill=mycolor2] (e) at ([yshift= 2.5cm]$(b)!0.3!(c)$) {Real Data}; 
% \draw[line, line width=0.5mm] (a)-- (b); 
% \draw[line, line width=0.5mm] (c)-- (d);
% \draw[line, line width=0.5mm] (b)-- (i); 
% \draw[line, line width=0.5mm] (i)-- (c); 
% \draw[line, line width=0.5mm] (e)-- (c);
% \end{tikzpicture}
% \caption{\label{fig:linear} GANs Block Diagram } 
% \end{figure} 

%%%%%%%%%%%%%%%%%%%%%%%%%%%%%%%%%%%%%%%%%%%%%%%%%%%%%%%%%%%%%%%%
\subsection{Loss Function}
The loss function is designed such that the players $D$ and $G$ are pitted against each other. At a particular iteration, $D$ tries to get better at classifying $x$ and $\hat{x}$. Its parameters denoted by $\theta$ are trained to maximize the loss to distinguish between the real and generated samples. In the same iteration, $G$ is also trained. The parameters of $G$ denoted by $\phi$ are optimized such that the discriminator is not able to distinguish between $x$ and $\hat{x}$. $G$ is essentially trained to produce images which are more realistic such that the discriminator is confused. Ideally, $\phi$  is trained to minimize the same loss that $\theta$ is maximizing. Hence it is similar to a zero-sum game where the players have total competition (Loss of $D$ is gain for $G$ and vice versa).

Formally the loss is given by,
\begin{equation}
    \label{eq:gan_loss1}
    \begin{split}{}
    \underset{\phi}{min}\ \underset{\theta}{max}
\ V(D_{\theta},G_{\phi}) =  \  \mathbb{E}_{x \sim p_{d}(x)}[log D_{\theta}(x)] +  \mathbb{E}_{z \sim p_{z}(z)}[log(1-D_{\theta}(G_{\phi}(z)))]
    \end{split}
\end{equation}{}

This is a typical binary cross entropy loss (Definition \ref{def:bce}), where the real data samples $x$ are given label $1$ and generated samples $\hat{x} := G(z)$ are given label $0$. Early in the training, the discriminator is very powerful and is able to discriminate almost perfectly, given the generated images are far from realistic. In this phase, the gradients w.r.t. $\phi$ is very small, hence there is no strong signal for $G$ to improve. Authors  in \citep{ganGoodfellow} instead propose the following objective for $G$ which has better gradients when $D(G(z))$ takes low values.
\begin{equation}
\label{eq:gan_loss2}
\underset{{\phi}}{max} \ \log(D_{\theta}(G_{\phi}(z)))
\end{equation}

Ideally, at the end of training, we would like the generator to have learnt the data distribution i.e. The next challenge is to ensure that the optimizer over the loss actually converges to the desired global minima in finite time with finite samples.
 In the following subsection we discuss the first method proposed to optimize over the loss.

\subsection{Optimizer}
In a typical generative model the main goal is to ensure that $p_g = p_d$, i.e. the generated distribution has learned the data distribution. Given the min max objective, the above occurs at a saddle point $\phi^* , \theta^*$. In Algorithm 1 of \citep{ganGoodfellow}, the authors propose \emph{Simultaneous Gradient Descent} as the optimizer. In this method, we first fix $\phi$ and optimize over $\theta$ by maximizing Equation (\ref{eq:gan_loss1}). Then we fix $\theta$ and maximize the Equation (\ref{eq:gan_loss2}) over $\phi$. We have to check for the following aspects in order to prove that this method converges. 
\subsubsection{Existence of Saddle Point}
The authors prove that the objective given by Equation (\ref{eq:gan_loss1}) has a global optima at $p_g = p_d$.
 The optimal discriminator for a fixed generator is given by $D^*_{G}(x) = \frac{p_{d}(x)}{p_{d}(x) + p_{g}(x)}$. For the optimal discriminator as given above, the generator is shown to minimize the following Jenson Shannon Divergence(JSD) Equation (\ref{eq:jsd}), 
 
    \begin{equation}\label{eq:2}
         C(G) = -\log(4) + KL\bigg(p_{d} \parallel \frac{p_{d} + p_{g}}{2} \bigg) + KL\bigg(p_{g} \parallel \frac{p_{d} + p_{g}}{2} \bigg)
    \end{equation}
    \begin{equation}\label{eq:3}
    C(G) = -\log(4) + 2. JSD(p_{d} \parallel p_{g})
    \end{equation} 
    
\begin{tcolorbox}[colback=gray!03!white,colframe=black!75!black]
\begin{theorem} \citep{ganGoodfellow}
The global minimum of the virtual training criterion $C(G)$ is achieved if and only if $p_g=p_d$. At that point, $C(G)$ achieves the value $−\log 4$.
\end{theorem}{} 
\emph{Assumption: the theorem holds true only in the space of density functions. Given that $D, G$ have finite parameters, they may not model every distribution from the entire space of distributions possible.  }
\end{tcolorbox}
\subsubsection{Convergence to A Saddle Point}
Given that the global optima exists under certain assumptions, we proceed to reiterate the results from \citep{ganGoodfellow} that Simultaneous Gradient Descent converges to the optima. 

\begin{tcolorbox}[colback=gray!03!white,colframe=black!75!black]
\begin{prop} \citep{ganGoodfellow}
\label{prop:conv}
If $G$ and $D$ have enough capacity, and at each step of the algorithm, the discriminator is allowed to reach its optimum given $G$, and $p_g$ is updated so as to improve the criterion then $p_g$ converges to $p_d$
\end{prop}{}
\emph{Assumption: In the proof, $V(G, D)$ is set to $U(p_g, D)$ and the proposition holds true as $U(p_g,D)$ is convex in $p_g$. Given that $G_{\phi}$ cannot model all possible $p_g$ and $G$ is non-convex w.r.t. $\phi$ the results may not hold true. }
\end{tcolorbox}

\subsection*{Major Issues With the Analysis}
\begin{enumerate}
    \item There is no quantitative estimate of the required capacity of $G, D$.
    \item It is not possible to train the discriminator to convergence at every iteration in finite time. 
    \item The above proposition does not guarantee that the convergence will happen in finite iterations.
    \item We also do not have generalization bounds, which provides an estimate of how well the model generalizes for a given number of training samples.
\end{enumerate}

\vspace{\baselineskip} 
Now that we are familiarized with the architecture of vanilla GAN, the primary loss function and also the algorithm to optimize the loss, we proceed to brief about its actual performance. The model's performance with minor changes is beyond expectation despite the major challenges in its theory. This has led to its popularity and has attracted attention from theorists and also from those interested in applications of generative models. Before discussing the challenges in GANs, we briefly summarize the success of GANs compared to other generative models in various applications and 

%%%%%%%%%%%%%%%%%%%%%%%%%%%%%%%%%%%%%%%%%%%%%%%%%%%%%%%%%%%%%%%%%%%%%%%%%%%%%%%%%%%%%%%%%%%%%%%%%%%%%%%%%%
\begin{figure}
    \centering
    \includegraphics[width=15cm]{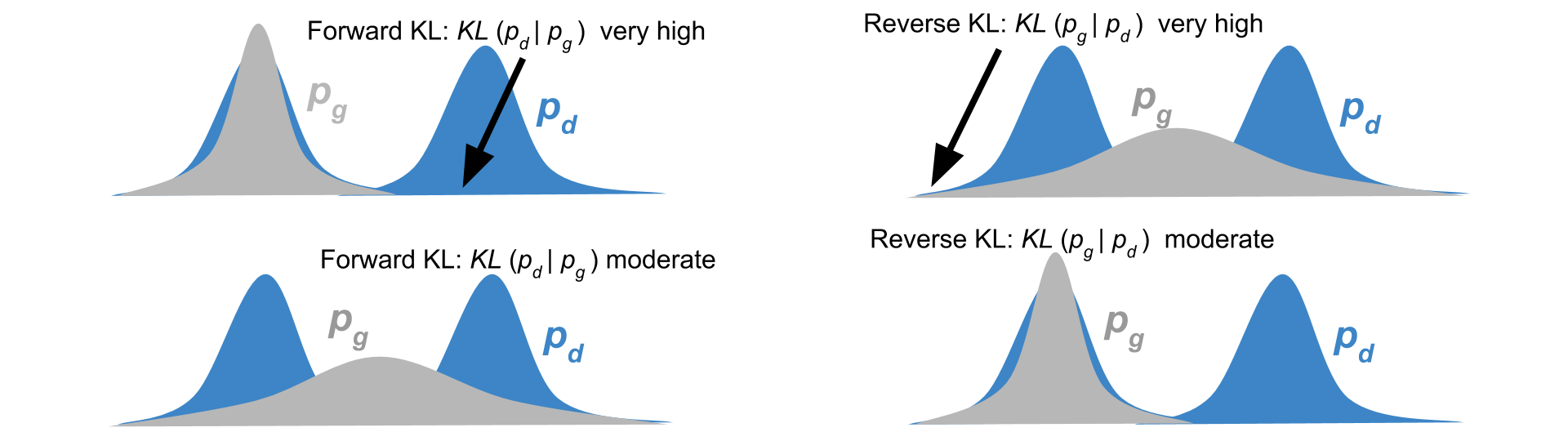}
    \caption{KL vs Reverse KL}
    \label{fig:frkl}
\end{figure}{}

\section{Success Stories}
\label{sec:success}
In this section, we discuss briefly about the advantages of GANs over two of the existing generative models. Then we also mention a few interesting applications of GANs in variety of areas including vision, text and other domains like music and even art.
\subsection{GANs vs Rest}
\begin{figure}[!ht]
\begin{subfigure}{.2\textwidth}
\centering
\includegraphics[width=3cm,keepaspectratio]{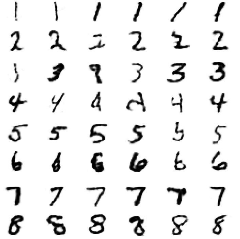}
\caption{RBM Generated MNIST}
\end{subfigure}
\begin{subfigure}{.25\textwidth}
\centering
\includegraphics[width=4cm, keepaspectratio]{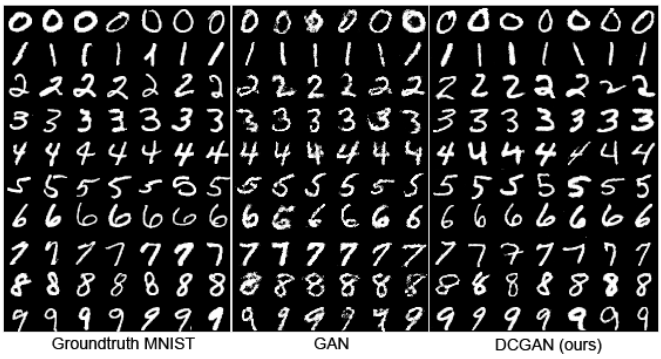}
\caption{DCGAN}
\end{subfigure}
\begin{subfigure}{.25\textwidth}
\centering
\includegraphics[width=3cm,keepaspectratio]{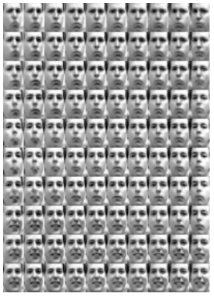}
\caption{VAE Learned Frey Face Manifold}
\end{subfigure}
\begin{subfigure}{.25\textwidth}
\centering
\includegraphics[width=3cm,keepaspectratio]{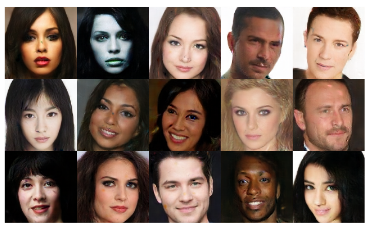}
\caption{Celeb A faces using BEGAN}
\end{subfigure}
\caption{Generative reults on MNIST digits and faces from the RBM, VAE, DCGAN and BEGAN paper to compare the quality of the results generated.}
\label{fig:1}
\end{figure}

Speaking of a generative model, we think of modeling the distribution explicitly by estimating it's parameters from the data. Given random variable $X$ which denotes our data, we model $p_{\theta}(x); x\sim X$. The objective is to maximize the log likelihood given by $\log p_{\theta}(x)$. Any generative model which at the end of training provides us the value of $p_{\theta}(x)$ is called an explicit model.
Restricted Boltzmann Machines (RBMs) \citep{hinton2006} , Deep Belief Networks (DBNs) \citep{dbn}, Variational Autoencoders (VAEs) \citep{kingma2014} are popular explicit models to name a few. These are latent variable models where estimating the maximum likelihood of the parameters requires integrating over the entire space of latent variables, making it intractable. RBMs maximize the likelihood using a procedure called \textit{contrastive divergence} which uses MCMC sampling. VAEs on the other hand minimize the variational lower bound to the negative log likelihood. The former faces the issues of mixing related to MCMC and the latter due to minimization of KL divergence as discussed further below. Although the theory is elegant and the models are simple to implement, the images generated are blurry (Figure \ref{fig:1}).  Besides they fail to produce complex images when trained on other datasets such as CIFAR, SVHN, etc.

GANs on the other hand are implicit models, which directly generate data without providing the density function. The images generated are sharper and realistic. It is believed that minimizing JSD (Equation (\ref{eq:jsd})) instead of KL(Equation (\ref{eq:kl})) might be one of the reasons for such good results. Maximizing likelihood technique is followed by VAEs which corresponds to minimizing KL. On examining Equation (\ref{eq:kl}), we can conclude that there is high penalty for whenever $p_d(x) > 0$ but $p_g(x) \rightarrow 0$  hence all data points must be fit into the model without missing modes. On the other hand penalty is low when $p_d(x) \rightarrow 0$ but $p_g(x) > 0$, implies generation of unrealistic samples. Similarly we can argue that reverse KL given by Equation (\ref{eq:rvkl}) would lead to better quality at the cost of missing samples from $p_d$. Refer to Figure (\ref{fig:frkl}) for better illustration. Unlike KL, JSD given by Equaiton (\ref{eq:jsd}) which GAN minimizes is symmetrical and ensures better quality images.

\subsection{Applications}
The follow-up research has vastly improvised upon the quality of results produced from vanilla GANs. Alec et al. DCGAN \citep{dcgan} have proposed a stable architecture and suitable values for the hyper-parameters for better training. Vision finds major application of GANs in super resolution \citep{progressive,srgan}, transferring domain knowledge from images of one domain to another \citep{cyclegan, discogan, pixeldtgan, pix2pix}, object detection \citep{perceptual}, image editing \citep{imageblend}, medical images \citep{anomaly}. Gans have also been used to generate music \citep{musegan} and paintings \citep{paint}. It is has also been used for generating text \citep{seqgan, relgan, advnl}. These are but a few of the applications that developed recently. The samples generated by these applications are exceptional.

Given the cool applications it is a wonder how effective GAN is even though it lacks a concrete theoretical analysis. Even after much progress, GANs are not without problems. Apart from the fact they have not been explained well theoretically, there are other issues with the generated samples which we list in the following section. Subsequently, we discuss the various mathematical approaches proposed towards characterization and mitigation of the issues.

%%%%%%%%%%%%%%%%%%%%%%%%%%%%%%%%%%%%%%%%%%%%%%%%%%%%%%%%%%%%%%%%%%%%%%%%%%%%%%%%%%%%%%%%%%%%%%%%%%%%%%%%%%

% \subsection{Different Views of GAN Optimization and Convergence}
% \subsubsection{Divergence Minimization}
% \subsubsection{Nash Equilibrium}
% \subsubsection{Regret Minimization}
%%%%%%%%%%%%%%%%%%%%%%%%%%%%%%%%%%%%%%%%%%%%%%%%%%%%%%%%%%%%%%%%%

\section{Primary Challenges in GAN Set-Up}
\label{sec:challenges}
In this section we discuss the key issues with GANs. Issues pertaining to the performance which are evident through experiments. There have been extensive papers trying to explain these issues with theoretical rigour. As a product of which we are introduced to other fundamental drawbacks of GANs. Typically the approach is to set up a generalized framework for the min-max objective given by Equation (\ref{eq:gan_loss1}) and then continue the analysis of convergence by borrowing the tools from the set framework. To be specific, authors in \citep{fgan} view the GAN objective as \textit{Divergence Minimization} and accordingly prove the convergence. There are papers which model it as a \textit{two-player zero sum game} and borrow the concept of Nash equilibrium to derive the optimal behaviour of the models. There is also a line of work which views it as a \textit{Regret Minimization} problem. Till date there has not been a satisfactory explanation not only for why the model fails but also for why it works sometimes when it does. 

There are three primary challenges we encounter while implementing the model: (i) \emph{mode collapse}, (ii) \emph{Non-Convergence and Instability}, and (iii) \emph{Evaluation of Generative Models}.
\subsection{Mode Collapse}
\label{subsec:mc}
\begin{figure}
    \centering
    \includegraphics[width=5cm,keepaspectratio]{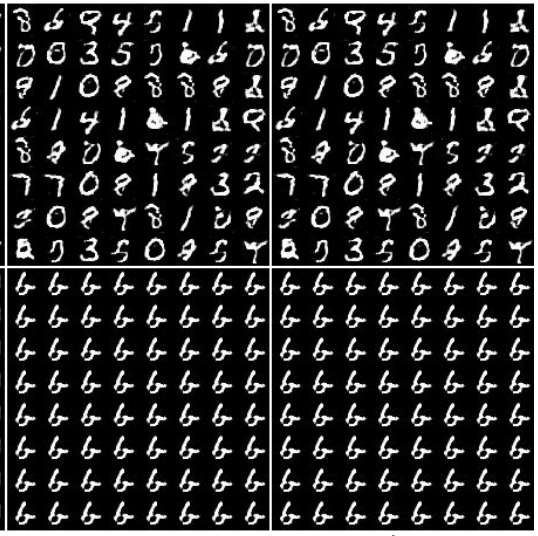}
    \caption{ In the second row, we see that vanilla GAN generates only one kind of image.  \citep{unrolled}}
    \label{fig:modecollapse}
\end{figure}

The most significant and widely discussed problem is \emph{Mode Collapse}. 
The data distribution $p_d$ generally is highly complex and spreads over many modes. These modes essentially represent the variation within the data. For example, considering the MNIST handwritten digit dataset, each mode could represent each type of digit. Ideally at convergence, $p_g = p_d$ i.e., the generated distribution should have equivalent number of modes. Unfortunately, it is found that the generated samples lack variation. This phenomenon which indicates that $p_g$ has just captured few modes is termed as mode collapse. Although the images generated are very sharp and realistic they do no have much variation. This phenomenon is clearly visible in Figure \ref{fig:modecollapse}. 
{}
Many researchers try to explain the problem and overcome it with novel approaches. In the subsequent sections we look into few key papers in further detail.

%%%%%%%%%%%%%%%%%%%%%%%%%%%%%%%%%%%%%%%%%%%%%%%%%%%%%%%%%%%%%%
\subsection{Non-Convergence And Instability}
\label{subsec:nci}
Training a GAN model is considered to be a precarious task. In a typical neural network training we look at the training and testing loss curves as an indication for stopping the training. It is assumed that when the training loss does not decrease any further and test loss is also at its lowest, we consider the loss has converged and model has attained the optimal parameters. Whereas in a GAN there are $G$ and $D$, $G$ tries to maximize the loss that $D$ minimizes. Hence the convergence is not evident through the training loss curves. 

\begin{figure}[!htb]
    \centering
    \includegraphics[width=6cm,keepaspectratio]{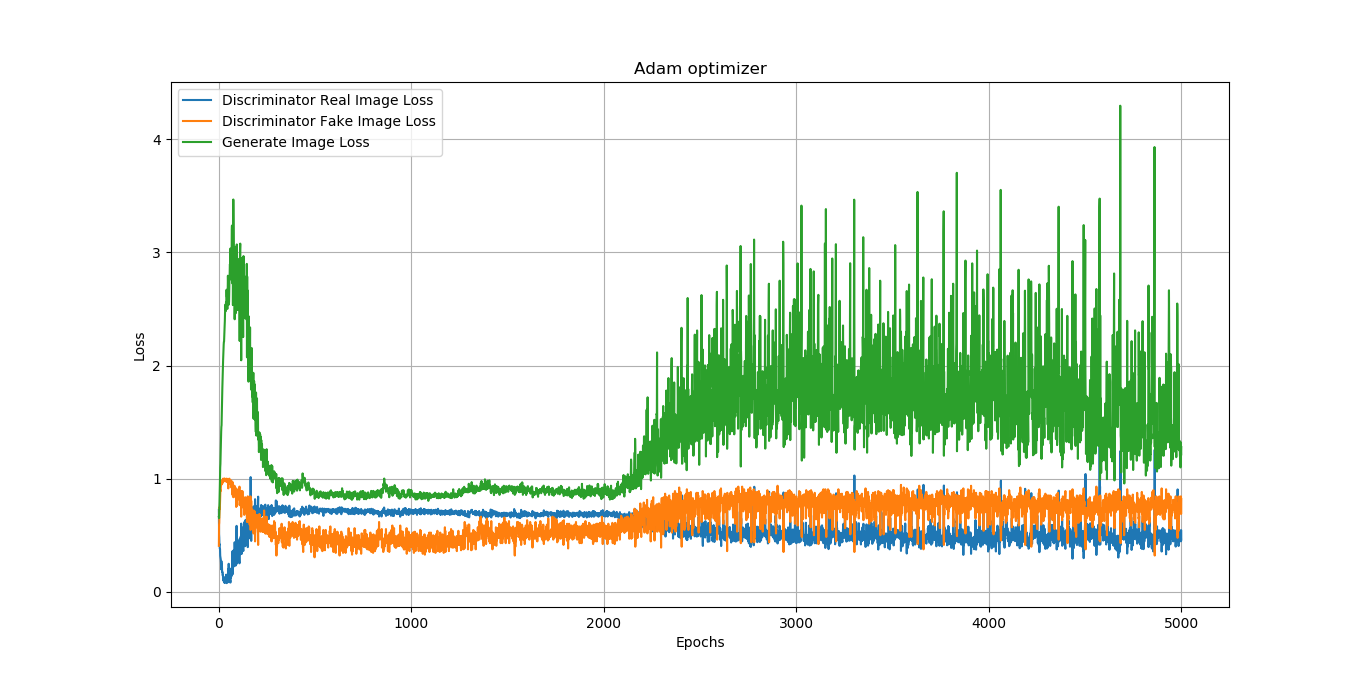}
    \caption{Training loss curves for GANs 
    (Fig 7 from \cite{loss_curves})}
    \label{fig:gan_loss}
\end{figure}

Generally, it is observed that the discriminator and generator losses converge to a particular value. This does not always imply that $p_g = p_d$, rather generator is always generating just few samples (mode collapse). It has also been observed that the generator and discriminator losses have not converged to any value, yet the generator is generating realistic samples. usually the losses are not smooth but have damped oscillations as in Figure \ref{fig:gan_loss}. These oscillations indicate that the training is highly unstable. Finally, sensitivity to hyper-parameters is a prevalent issue in deep learning models yet the issue is exaggerated within GANs.

{}

% While performing simultaneous gradient descent for training the generator and discriminator, it is tricky to decide upon the number of iterations for which the generator and discriminator have to be trained. Improper balance of the number of iterations results in \emph{Undamped Oscillations}. Ideally training the discriminator to optimality between every generator updates not only is computationally expensive but also results in a pessimistic discriminator leading to the problem of vanishing gradients. The convergence of the GAN game as proved in the original paper Goodfellow et al. \citep{ganGoodfellow} holds true if the function is convex, the model has infinite capacity, and enough training samples are available. These assumptions are not valid in practice.  Non-convergence of the simultaneous gradient descent has been discussed in many papers. The convergence is not guaranteed sometimes even in convex cases like $xy=0$. 

We can design a GAN set-up without mode collapse and unstable training only if we find positive answers for the three question given below.
\begin{tcolorbox}[colback=gray!03!white,colframe=black!75!black]
\begin{itemize}
    \item[Q$_1$] Does the model have enough capacity to capture the complex data distribution $p_d$?
    \item[Q$_2$] Given enough capacity, will minimizing the given objective function guarantee $p_g = p_d$ at optimality with considerable generalization bounds?
    \item[Q$_3$] Assuming there exists a global optima which corresponds to $p_g = p_d$, do we have an optimization algorithm which reaches it in finite iterations?
\end{itemize}
\end{tcolorbox}
 In the next subsection, we briefly discuss a third major challenge with GANs. Although in the rest of paper, our focus is restricted to the above two challenges.
%%%%%%%%%%%%%%%%%%%%%%%%%%%%%%%%%%%%%%%%%%%%%%%%%%%%%%%%%%%%%%%%
\subsection{Evaluation of Generative Models}

In many problems like classification, regression, object detection, segmentation etc., there is a precise quantitative way of testing a model's performance in terms of its accuracy or loss or mAP (mean Average Precision). In a generative model like GAN, measuring its performance is not an obvious task. Our aim is to generate realistic images similar to the data while not always generating the same images as in the data. Hence, mean square error loss between generated and real images or any such distance metric or similarity score is not a correct measure for realistic looking samples. Unless we have a human score the images, qualitative estimation is not defined well for such models. In the original paper \citep{ganGoodfellow} the authors use Parzen window to fit the generated samples and estimate the model's log-likelihood. It is also a common practice to evaluate based on the model's performance on some surrogate tasks like classification, de-noising or missing value imputation. The most widely accepted measure, for now, is the \emph{inception score}, proposed by Salimans et al. \citep{goodfellow16}. \citep{proCon} is a good survey discussing the pros and cons of the evaluation metrics. Ultimately, it is desirable to have a metric that evaluates both diversity and visual fidelity simultaneously.

In the past few years, researchers have actively worked towards proposing a solution for the above challenges. Besides providing new loss functions and architectures, significant amount of work has been dedicated towards building a theoretical framework to analyze the challenges. In the next section we categorize the recent work based on the kind of modifications they introduce in GAN set-up. We also briefly summarize each of these papers. 

%%%%%%%%%%%%%%%%%%%%%%%%%%%%%%%%%%%%%%%%%%%%%%%%%%%%%%%%%%%%%%%%%%%%%%%%%%%%%%%%%%%%%%%%%%%%%%%%%%%%%%%%%%%

\section{Progress So Far}
\label{sec:novel}
In this section, we discuss recent works which address the issues in GANs and build a theoretical framework to explain why the model works or why it fails. We have organized the section by categorizing the papers based on the kind of novelty they introduce. Primarily there are three ways in which researchers have tried to change the vanilla GAN set-up for resolving the issues discussed in the previous section. 
S$_1$) Proposing a new loss or introducing regularization in the existing loss (\ref{subsec:loss}). S$_2$) Changing the architecture (\ref{subsec:arch}). S$_3$) Changing the optimizer used (\ref{subsec:optim}). S$_4$) Finally, we also discuss the papers which provide rigorous theoretical analysis for a specific approach followed or for the existing issues (\ref{subsec:theory}).

\subsection{Loss Functions}
\label{subsec:loss}
We begin with the papers which modify the loss function  without changing the architecture. The original objective for the vanilla GAN is given by Equation (\ref{eq:gan_loss1}). The loss function can be modified to achieve better performance and resolve various issues as discussed by the following papers. We have further categorized the papers which modify the loss, in five groups. This is based on the rationale the papers follow behind introducing the loss.

%\bigskip

\subsubsection{Resolve Vanishing Gradients.} \label{subsubsec:loss_vg} The following  papers in this subsection introduce new loss function to overcome the problem of vanishing gradients as characterized in the first paper that we discuss \citep{arjovsky01}

\subsection*{Towards Principled Methods for Training GANs \citep{arjovsky01}}
In this paper the authors analyze the gradients for $D$ and $G$ to pin-point the reason for mode collapse and stability issues. The notion of \emph{perfect discriminator} is introduced. According to theory the discriminator will have a cost at most $2log2 - 2JSD(p_d \parallel p_g)$ given in Equation \ref{eq:3}. However when trained to convergence the error goes to 0 because of disjoint support of the distributions. Both $p_d$ and $p_g$ lie on low dimensional manifolds and hence unlikely to align perfectly. This results in the problem of  \emph{vanishing gradient}. When the discriminator is trained to convergence and becomes perfect, the gradients w.r.t. the generator parameters $\phi$ vanishes when using Equation (\ref{eq:gan_loss1}). If the two distributions $p_d$ and $p_g$ are on disjoint support then there will always exist a $D^*$ whose accuracy is $1$ and gradient is $0$ near the real samples, which results into vanishing gradients. \\
 To avoid vanishing gradients, Equation (\ref{eq:gan_loss2}) is minimized w.r.t. $G$ instead of Equation (\ref{eq:gan_loss1}). Let $D^* = \frac{p_{d}}{p_{g_{\phi_{0}}} + p_{d}}$, be the optimal $D$ when $G$'s parameters are fixed to $\phi_0$.
\begin{dmath*}
\mathbb{E}_{z\sim p(z)}[-\nabla_{\theta} \log D^*(g_{\phi}(z))|_{\phi = \phi_{0}} ] = \nabla_{\phi}[KL(p_{g_{\phi}}\parallel p_{d}) - 2 JSD(p_{g_{\phi}}\parallel p_{d})] |_{\phi = \phi_{0}}
\end{dmath*} 
Using the above, although reduces the problem of vanishing gradients yet it leads to mode collapse. The gradients for $G$ are such that, the JSD is in opposite direction, pushing for the distributions to be different. The inverted KL is not maximum likelihood, instead it assigns extremely high cost to generating fake looking samples, and extremely low cost to mode dropping. Moreover, the authors show results which show that if the $D$ above is not optimal i.e. it has not yet reached to being $D^*$ then the gradients will follow Cauchy distribution with infinite mean and variance causing unstable updates

Hence the authors discuss the following ways to mitigate the issue of vanishing gradients instead of using Equation \ref{eq:gan_loss2}
\begin{itemize}
    \item They propose to add continuous noise $\epsilon$ to the real data i.e, $x \sim p_d$ and model dat $\hat{x} \sim p_g$ hence the gradients w.r.t. $G$ will be $\nabla_{\phi} JSD(p_{d+\epsilon} || p_{g + \epsilon})$ which leads to matching of the noisy distributions. The noise is annealed over the time and hence the procedure leads to the matching of the actual distributions $p_d$ and $p_g$. Hence minimizing the JSD between the noisy variants mitigates the issue of mode collapse, but the training will become highly sensitive to the kind of noise used.
    \item The authors also introduce Wasserstein metric (Equation \ref{eq:em}) discussed in further detail in the next paper \cite{arjovskyWGAN}. They give insightful relation where this metric is upper bounded by $2 var(\epsilon)^{1/2} + 2\sqrt{ JSD(p_{d+\epsilon} || p_{g+\epsilon})}$
\end{itemize}{}

\subsection*{Wasserstein GAN  \citep{arjovskyWGAN}}  The authors propose a different distance metric to overcome vanishing gradients. The Earth Mover (EM) distance can be useful for learning distributions in lower dimensional manifold. The EM distance or Wasserstein distance is given by, 
\begin{equation}
    \label{eq:em}
    W(p_{d}, p_{g}) = \underset{\gamma \in \Pi(p_{d}, p_{g})}{inf} \mathbb{E}_{(x,y)\sim \gamma}[\parallel x - y \parallel]
\end{equation}{}
$\Pi(p_{d},p_{g})$ denotes the set of all joint distributions $\gamma(x, y)$ whose marginals are respectively $p_{d}$ and $p_{g}$. The Wasserstein distance is much weaker distance and is continuous in $\phi$ if $g$ is continuous in $\phi$. Equation (\ref{eq:em}) is intractable hence Kantorovich-Rubinstein duality is used to transform the objective into,
\begin{equation}
    \label{eq:em_dual}
    W(p_{d}, p_{g}) = \underset{\parallel f\parallel_{L}\leq 1}{sup} \mathbb{E}_{x\sim p_{d}} [f(x)] - \mathbb{E}_{x\sim p_{g}} [f(x)]
\end{equation}{}
where the supremum is over all the 1-Lipschitz functions. Thus given a parameterized family of 1-Lipschitz functions $\{f_{w}\} _{w\in \mathcal{W}}$,  solve the following problem which characterizes the Wasserstein GAN or WGAN objective, 
\begin{equation}
    \label{eq:wgan}
    \underset{w\in \mathcal{W}}{max} \ \mathbb{E}_{x\sim p_{d}} [f_{w}(x)] - \mathbb{E}_{z\sim p_{z}(Z)} [f_{w}(g_{\phi}(z))]
\end{equation}{}
The $f_{w}$ can be modeled as a neural network, where the fact that $f$ is 1-Lipschitz depends on $\mathcal{W}$ being compact. One way of enforcing the compactness is to clamp the weights to a fixed box. Using the above objective waives the need for balancing the generator and discriminator. In this case discriminator referred to as the critic could be trained till optimality without losing gradients.  

\subsection*{A two-step computation of the exact GAN Wasserstein distance \citep{wgan_ts}} WGAN uses weight clipping to ensure 1-Lipschitz condition in order to optimize over the dual formulation given by Equation (\ref{eq:em_dual}). But clipping weights causes vanishing and exploding gradients problem. Hence the authors propose a two-step formulation to compute Wasserstein distance, which is equivalent to the dual and does not need additional weight clipping or penalty.

To understand the proposed solution we would look at how the problem is set up such that it is equivalent to the Kantorovich duality. Then we would see how it is solved and finally we will see the modified objective of WGAN.
Problem Set Up
\begin{prob}
\label{prob:1}
Suppose $X$ and $Y$ are two bounded domains in $\mathbb{R}^n$. Given two probability measures $\mu \in \mathbb{P}(X), \nu \in \mathbb{P}(Y)$ and a cost function $c: X \times Y \rightarrow [0, +\infty]$. Find functions $\phi, \psi$ such that,
$$
C(\mu, \nu)=\sup _{\phi-\psi \leq c}\left\{\int \phi(y) d \nu(y)-\int \psi(x) d \mu(x)\right\}
$$
where $C(\mu, \nu)$ is the Wasserstein distance between $\mu$ and $\nu$.
\end{prob}
The problem is further transformed to the following,
\begin{prob}
\label{prob:2}
Find the function $\psi$ such that
$$
C(\mu, \nu)=\sup _{\psi}\left\{\int \psi^{c}(y) d \nu(y)-\int \psi(x) d \mu(x)\right\}
$$
where $C(\mu, \nu)$ is the Wasserstein distance between $\mu$ and $\nu$ and $\psi^c$ is the $c$-transform of the $\psi$ defined below:
$$
\forall y \in Y \qquad \psi^{c}(y)=\inf _{x \in X}(\psi(x)+c(x, y))
$$
\end{prob}{}
Since, we only have access to the samples so we need to discretize the problem as follows,
\begin{prob} Let,
\label{prob:3}
$$
\hat{d}(\psi)=\frac{1}{m} \sum_{i \in \mathcal{I}} \psi^{c}\left(y_{i}\right)-\frac{1}{n} \sum_{j \in \mathcal{J}} \psi\left(x_{j}\right)
$$
Find $\psi$ such that $\hat{C}(\mu, \nu) = sup_{\psi}\hat{d}(\psi))$ where $C(\mu, \nu)$ is the Wasserstein distance between $\mu$ and $\nu$ and $\psi^c$ is the $c$-transform of the $\psi$ defined below:
$$
\forall y_{i} \in \hat{Y} \qquad \psi^{c}\left(y_{i}\right)=\inf _{x \in \hat{X}}\left(\psi(x)+c\left(x, y_{i}\right)\right)
$$
\end{prob}

In order to make $\psi^c = \psi$ WGAN restricts the function to be 1-Lipschitz. The authors propose a new formulation to evade the above restriction.
\begin{prob}
\label{prob:4}
Solve the following problem,
$$
\begin{array}{ll}{\max _{f}} & {\hat{h}(f)=\left\{\frac{1}{m} \sum_{i \in \mathcal{I}} f\left(y_{i}\right)-\frac{1}{n} \sum_{j \in \mathcal{J}} f\left(x_{j}\right)\right\}} \\ {\text { s.t. }} & {f\left(y_{i}\right)-f\left(x_{j}\right) \leq c\left(x_{j}, y_{i}\right), \quad \forall j \in \mathcal{J}, \forall i \in \mathcal{I}}\end{array}
$$
\end{prob}{}
The authors go on to prove that Problem 3 and 4 are equivalent in the following theorem,
\begin{theorem}
If the cost function $c(·,·)$ satisfies the triangle inequality, then solving Problem \ref{prob:4} is equivalent to solving Problem \ref{prob:3}, i.e., the optimal objectives of Problem \ref{prob:4} and \ref{prob:3} are equal and $f^*(x_j) = \psi^*(x_j)$ and $f^*(y_i) = (\psi^c)^* (y_i)$, where $f^*, \psi^*$ are optimizers for Problem \ref{prob:4} and \ref{prob:3} respectively. 
\end{theorem}{}
Solving the dual formulation:
Step 1: Solve the following linear programming
$$
\begin{array}{cl}{\max _{T}} & {\frac{1}{m} \sum_{i \in \mathcal{I}} T_{i}-\frac{1}{n} \sum_{j \in \mathcal{J}} T_{j}} \\ {\text { s.t. }} & {T_{i}-T_{j} \leq c_{i j}, \forall i \in \mathcal{I}, \forall j \in \mathcal{J}}\end{array}
$$
The optimizer $T^*$ is unique upto a scalar hence we set $
 T_{t}^{*} \leftarrow T_{t}^{*}-\left(\sum_{k \in \mathcal{I} \cup \mathcal{J}} T_{k}^{*}\right) /(m+n) \  \forall t \in \mathcal{I} \cup \mathcal{J} $ .
The result obtained is exact Wasserstein distance but not differentiable hence we have the following step\\
Step 2: Optimize the following regression problem
$$
\min _{f} \quad \frac{1}{m+n}\left(\sum_{i \in \mathcal{I}}\left(f\left(y_{i}\right)-T_{i}^{*}\right)^{2}+\sum_{j \in \mathcal{J}}\left(f\left(x_{j}\right)-T_{j}^{*}\right)^{2}\right)
$$
This provides a differentiable approximation \\
WGAN-TS: Given $D$ is the discriminator and $G$ the generator, the new objective for WGAN is given as follows,
$$
\begin{array}{ll}{\min _{G} \max _{D}} & {\hat{C}(f)= \frac{1}{m} \sum_{i \in \mathcal{I}} D(y_{i})- \frac{1}{n} \sum_{j \in \mathcal{J}} D(G(z_{j}))} \\ 
{\text { s.t. }} &
{D\left(y_{i}\right)-D\left(G\left(z_{j}\right)\right) \leq c\left(y_{i}, G\left(z_{j}\right)\right), \quad \forall i, \forall j}\end{array} 
$$
where $c(y_i, G(z_j)) = \parallel y_i - G(z_j)\parallel_1$. The generator loss is computed as follows,
$$
\min _{G}-\frac{1}{n} \sum_{j \in \mathcal{J}} D\left(G\left(z_{j}\right)\right)$$

\subsection*{Improved Techniques for Training GANs \citep{goodfellow16}}

The authors propose different heuristics to deal with the issue of vanishing gradients, mode collapse and non-convergence. They introduce the notion of \emph{feature matching} which prevents the generator from getting over trained on the current discriminator by minimizing the following objective, given $f(x)$ is the feature obtained from the intermediate layer of the discriminator.
$$\parallel\mathbb{E}_{x \sim p_{d}}f(x) - \mathbb{E}_{z \sim p_{z}(z)} f(G(z)) \parallel _{2}^2$$
The discriminator is trained in the usual way which tries to find the features most discriminative of the real and fake data.

The authors introduce the technique of \textit{ Mini batch discrimination} for overcoming the issue of mode collapse. Instead of discriminating between one real sample and one generated sampled at a time, the idea is to discriminate between the representation of batch of samples, where the representation models the differences between the samples within a batch. The representation of a single sample includes the value of how different it is from every other sample within a batch. The difference is captured by the metric $c_b$ defined below. Let $f(x_{i}) \in \mathbb{R}^A$ denote a vector of features for input $x_{i}$ produced by some intermediate layer in the discriminator. $f(x_{i})$ is multiplied by a tensor $T \in \mathbb{R^{A \times B \times C}}$, which results in a matrix $M_{i} \in \mathbb{R}^{B \times C}.$ $$c_{b}(x_{i}, x_{j}) = exp(- \parallel M_{i, b} - M_{j, b} \parallel_{L_{1}} ) \in \mathbb{R} $$ $$o(x_{i})_{b} = \sum_{j=1}^n c_{b}(x_{i},x_{j}) \in \mathbb{R}$$ $$o(x_{i}) = [o(x_{i})_{1}, o(x_{i})_{2}, \ldots ,o(x_{i})_{B}] \in \mathbb{R}^B, \mbox{  } o(X) \in \mathbb{R}^{n \times B} $$ The $o(x_{i})$ is concatenated with $f(x_{i})$ and fed to the next layer of the discriminator.
The authors also introduce other heuristics to deal with the convergence of the training which we explain further in Subsection \ref{subsubsec:loss_nc}

\subsection*{Loss Sensitive GAN on Lipschitz Densities \citep{ls-gan}}
In this paper the authors introduce a margin between the generator loss and discriminator loss such that discriminator loss is always lower than the generator's loss by a margin. $$D_{\theta}(x) \leq D_{\theta}(G_{\phi}(z)) - \Delta(x, G_{\phi}(x)) $$ where $\Delta(x, G_{\phi}(x))$ is the margin and set to the $L_p$ distance with $p=1$. The above margin is relaxed using slack variable and finally, Loss Sensitive GAN (LS-GAN) optimizes $D_{\theta}$ and $G_{\phi}$ alternately. Loss for $D$ is,
$$ \underset{\theta}{min} \ \underset{x \sim p_d}{\mathbb{E}} [D_{\theta}(x) + \lambda \underset{\hat{x} \sim p_g}{\mathbb{E}} (\Delta (x, \hat{x}) + D_{\theta}(x) - D_{\theta}(\hat{x}))_{+}]
$$
with $(a)_{+} = max(a,0)$. $G$ minimizes the following,
$$\underset{\phi}{min} \underset{z \sim p_z(z)}{\mathbb{E}} D_{\theta}(G_{\phi}(z))$$
Comparison with WGAN: The WGAN objective given by Equation (\ref{eq:wgan}) it maximizes the first order moments of $f_w$. The second term in this equation can take very small values for generated samples, hence the loss can be arbitrarily high. In order to mitigate this, weight clipping is used. On the contrary in LS-GAN the loss is maximized till $D_{\theta}(\hat{x}) - D_{\theta}(x)$ exceeds $\Delta(x, \hat{x})$. At optimality they prove that non-parametric solution to their loss function has non vanishing gradient almost everywhere.\\

\subsection*{Least Squares GAN  \citep{lsgan}} The authors propose to use least square loss as opposed to binary cross entropy (Definition \ref{def:bce}) used in the original objective given by Equation (\ref{eq:gan_loss1}) for the discriminator. In GAN $D$ tries to learn the decision boundary between $x$ and $\hat{x}$. Given the loss is BCE, it assigns probability 1 to those fake samples which are in the right side of the decision boundary yet far away from the real data causing vanishing gradients. Hence they propose a loss which penalizes fake samples far from decision boundary hence forcing the generator to generate samples towards decision boundary.

Let $a,b$ be the labels for fake and real data respectively, then the objective function for Least Squares GAN (LSGAN) is given by the following. $c$ is the value that $G$ wants $D$ to believe for fake data.
\begin{equation}
    \label{eq:lsgan}
    \begin{aligned} \min _{D} V_{\mathrm{LSGAN}}(D) &=\frac{1}{2} \mathbb{E}_{\boldsymbol{x} \sim p_{\mathrm{d}}(\boldsymbol{x})}\left[(D(\boldsymbol{x})-b)^{2}\right]+  \frac{1}{2} \mathbb{E}_{\boldsymbol{z} \sim p_{\boldsymbol{z}}(\boldsymbol{z})}\left[(D(G(\boldsymbol{z}))-a)^{2}\right] \\ \min _{G} V_{\mathrm{LSGAN}}(G) &=\frac{1}{2} \mathbb{E}_{\boldsymbol{z} \sim p_{\boldsymbol{z}}(\boldsymbol{z})}\left[(D(G(\boldsymbol{z}))-c)^{2}\right] \end{aligned}
\end{equation}
The above objective is equivalent to minimizing Pearson $\chi^2$ divergence when $b-c=1,\ b-a=2$. \\

\subsection*{A Convex Duality Framework for GANs \citep{convDual}}
As shown in the original paper \cite{ganGoodfellow}, GANs are trained using a minmax objective which reduces to minimizing the JSD divergence if we assume the models to have infinite capacity and hence mimic all possible distributions (Section \ref{sec:set_up}). In practice $D$ is restricted to a smaller class of distributions denoted by $\mathcal{F}$ as shown in \citep{Arora03}. The authors in this paper propose to study the divergence minimization perspective in such a restricted setting by developing a convex duality framework.

Given an unrestricted $D$, the minmax objective is reduced to,
$$
\min _{\phi} \ \mathrm{JSD}\left(p_d(x), p_{g_{\phi}}(z)\right)
$$
The authors propose a duality framework, given a general divergence measure $d(P,Q)$ between two probability distributions $P, Q$ they define $d$'s
 conjugate over $P$ as follows
$$
d_{P}^{*}(D) :=\sup _{Q} \mathbb{E}_{Q}[D(x)]-d(P, Q)
$$

On restricting $D$ to belong to a convex class of functions $\mathcal{F}$, the Theorem 1 from \cite{} gives the following result,
$$
\begin{aligned}  \min _{\phi} \  \max _{D \in \mathcal{F}} \  \mathbb{E}_{P_{x\sim p_d}}[D(x)]-d_{p_{g_{\phi}}}^{*}(D) = \min _{\phi} \ \min _{Q}\ \left\{d\left(p_{g_{\phi}}, Q\right)+\max _{D \in \mathcal{F}}\left\{\mathbb{E}_{x\sim p_d}[D(x)]-\mathbb{E}_{\hat{x}\sim Q}[D(\hat{x})]\right\}\right\} \end{aligned}
$$

In other words the objective is searching for the generative model which is closest to the distribution $Q$ that shares the same moments as $p_d$. Further restricting $\mathcal{F}$ to a linear space, i.e., for any $D_1, D_2 \in \mathcal{F}$ and $\lambda \in \mathbb{R}$, $D_1 + \lambda D_2 \in \mathcal{F}$. Then, 

$$
\min _{\phi} \  \max _{D \in \mathcal{F}} \ \mathbb{E}_{x\sim p_d}[D(x)]-d_{p_{g_{\phi}}(z)}^{*}(D)=\min _{\phi} \min _{Q \in \mathcal{P}_{\mathcal{F}}\left(p_{d}\right)} d\left(p_{g_{\phi}}(z), Q\right)
$$
where $
\mathcal{P}_{\mathcal{F}}(P) :=\left\{Q : \forall D \in \mathcal{F}, \mathbb{E}_{Q}[D(x)]=\mathbb{E}_{P}[D(x)]\right\}
$
Using this formulation the authors also go on to prove that the moment matching interpretation holds true for $f$-GAN and WGAN for convex set $\mathcal{F}$ of 1-Lipschitz functions. 

The authors also provide a hybrid loss to overcome the vanishing gradient issue of JSD, consider two distributions $P_1$ and $P_2$ and denote Equation \ref{eq:wgan} by $W_1$
$$
d_{\mathrm{JSD}, W_{1}}\left(P_{1}, P_{2}\right) :=\min _{Q} W_{1}\left(P_{1}, Q\right)+\operatorname{JSD}\left(Q, P_{2}\right)
$$ The above is a special case of the following framework defined by the authors,
$$
d_{f, W_{1}}\left(P_{1}, P_{2}\right) :=\inf _{Q} W_{1}\left(P_{1}, Q\right)+d_{f}\left(Q, P_{2}\right)
$$

The authors prove that the above hybrid divergence is continuous w.r.t. to $P_1$ hence avoids the problem of vanishing gradients which occurs in JSD.

\subsubsection{Regularization for Vanishing Gradients} 
\label{subsubsec:loss_reg}In this subsection, we discuss papers which introduce different kinds of regularization on weights or gradients and as a result try to deal with the key issues of mode collapse and non-convergence. 

\subsection*{Stabilizing GANs through Regularization \citep{fgan_2}} 
This is a follow up on the $f$-gan paper \citep{fgan} discussed in Section \ref{subsubsec:theory_q3}. The authors claim that the fragility of gan training is due to non-overlapping model distribution and data distribution manifolds in the high dimensional space, which is termed as dimensionality misspecification. $f$-GAN models fail under such conditions.
Usually, such issue is taken care by adding high-dimensional noise, which introduces significant variance in the parameter estimation hence making the solution impractical. Instead, the authors propose analytic convolution of the densities with the Gaussian noise which yields a weighted penalty function on the norm of the gradients w.r.t. the input.
The following noise induced regularization results in a stable GAN training procedure. 

Noise Induced Regularization: $f$-Gan Objective as in Equation (\ref{eq:5}), given two distributions $P,Q$ and discriminator $D$, $F(P,Q;D) = E_{P}[D] - E_{Q}[f^* \circ D] \ $ \ 
Noise convolution, adding white noise $\xi \sim \Lambda = \mathcal{N}(0, \gamma I)$ to samples $x \sim P, Q$ :
\begin{dmath*}
E_{P}E_{\Lambda}[D(x + \xi)] = \int
D(x) \int p(x - \xi) \lambda(\xi)d\xi dx = \int D(x)(p*\lambda)(x)dx = E_{P*\Lambda}[D]
\end{dmath*} 
$p$ and $\lambda$ are probability densities of $P$ and $\Lambda,$ $\lambda(x)>0$ and $(p*\lambda)(x) > 0\mbox{ }(\forall x)$.\\
Regularized $f$-GAN given that $f^*$ is twice differentiable $$V_{\gamma}(P, Q; D) = E_{P}[D] - E_{Q}[f^* \circ D] - \frac{\gamma}{2} \Omega_{f}(Q;D)$$ $$\Omega_{f}(Q;D) := E_{Q}[({f^*}^{''} \circ D) \parallel \nabla D \parallel^2] $$

According to our notations the $P$ is $p_d$ and $Q$ is the generator parameterized by $\phi$ i.e., $G_{\phi}$

\subsection*{Improved Training of WGAN \citep{gularajani}}
The problem of exploding or vanishing gradients may resurface even in a WGAN setting, because of the use of weight clipping to enforce Lipschitz constraints. The subsequent paper after WGAN, Guljarani \emph{et. al.} \citep{gularajani} addresses the issues related to weight clipping. Apparently, weight clipping leads to capacity under use, i.e., the critic is biased towards much simpler functions. They introduce an alternative way of maintaining the Lipschitz constraints by introducing a gradient penalty term. They prove that the optimal critic has unit norm gradients everywhere, hence their penalty term constrains the gradients of the critic to be 1. The objective for WGAN with gradient penalty or WGAN-GP is,

\begin{equation}
\label{eq:wgan_gp}
    \begin{aligned}
        L = \mathbb{E}_{\tilde{x}\sim p_{g}}[D(\tilde{x})] - &  \mathbb{E}_{x\sim p_{d}}[D(x)] \\
        & + \lambda\mathbb{E}_{\hat{x}\sim p_{\hat{x}}}[(\parallel\nabla_{\hat{x}}D(\hat{x}) \parallel_{2} - 1)^2]
    \end{aligned}{}
\end{equation}{}

$p_{\hat{x}}$ sampling uniformly along straight lines between pairs of points sampled from the data distribution $p_d$ and the generator distribution $p_{g}.$ The authors claim an increase in sample quality and training speed.

\subsection*{Improving the Improved Training of WGANs \citep{imp_wgan_gp}}
Propose a novel way of imposing the Lipschitz condition on WGAN, in addition to gradient penalty introduced in Equation (\ref{eq:wgan_gp}) by \citep{gularajani}, to generate better photo-realistic samples. The authors claim that gradient penalty introduced in WGAN-GP cannot ensure Lipschitz condition everywhere in the support in finite training iterations. During the initial iterations, the generated samples maybe far from the actual manifold hence, the domain especially near the real data may not follow the Lipschitz condition. \\
To mitigate the above issue the authors propose to lay the Lipschitz constraint over the real data by perturbing it twice. The perturbation is implicit, since for every $x$ instead of perturbing the $x$ the authors perturb the $D(x)$ by introducing dropout in the discriminator network. Finally, they add the $CT$ defined below to the GAN objective. 
\begin{itemize}
    \item $D(x')$ : the discriminator output when the dropout rate applied in the hidden layers is small
    \item $D(x'')$ : the discriminator output after applying stochastic dropout again.
    \item Consistency Term: there exists a real constant $M \geq 0$ such that 
    $$CT|_{x', x''} = \mathbb{E}_{x \sim p_d}\bigg[ max\bigg( 0, \frac{d(D(x'), D(x''))}{d(x', x'')} - M'\bigg) \bigg]$$
    since it is difficult to compute $d(x', x'')$ as they are virtual points, hence it is assumed that the Lipschitz constant absorbs it.
\end{itemize}{}

\subsection*{On Regularization of WGANs \citep{reg_wgan}}
There are two ways of enforcing Lipschitz constraint in WGANs, through weight clipping \citep{arjovskyWGAN} and through a regularization term that penalizes the deviation the gradient norm of the critic from one \citep{gularajani}. The aim is to present a theoretical argument why the latter is harmful for training and propose a less restrictive regularization. 

The latter technique WGAN-GP requires the data and model samples to be drawn from a certain joint distribution and requires the optimal critic to be differentiable. In practice though, it is drawn independently from the marginal distributions as given by 
\begin{observe}
Suppose $f^* \in Lip_1$ is an optimal critic and $\gamma^*$ the optimal coupling determined by Equation (\ref{eq:em}). Then the optimal critic $f^* $ has unit gradients i.e. $\left|f^{*}(y)-f^{*}\left(x_{t}\right)\right|=\left\|x_{t}-y\right\|_{2}$ on the line $x_{t}=t x+(1-t) \hat{x}, 0 \leq t \leq 1, \text { for }(x, \hat{x})$ sampled from $\gamma^*$ but not when $x$ and $\hat{x}$ are sampled from their marginals $p_d$ and $p_g$ respectively as in \citep{gularajani}
\end{observe}{}
The assumption of differentiability of the optimal critic is not valid at points of interest as characterized by the following observation supported by proofs,
\begin{observe}
The assumption of differentiability of the optimal critic is not valid at points of interest
\end{observe}{}
Hence they propose a less restrictive penalization for violating the Lipschitz constraint; given $x\sim\mu$ and $y \sim \nu$
the following is added to the GAN loss leading to WGAN-LP,
$$
\left(\max \left\{0, \frac{|f(x)-f(y)|}{\|x-y\|_{2}}-1\right\}\right)^{2}
$$
The above regularization is also shown to be less sensitive to the penalty weight $\lambda$
\subsection*{Fisher GAN \citep{fisher}} 
Weight clipping in WGAN results in reducing the capacity of the discriminator and high sensitivity to the choice of hyper-parameters of clipping. WGAN-GP has high computational cost. The authors introduce a data-dependent constraint which maintains the capacity of the critic while ensuring the stability of training.  Based on Definition \ref{def:IPM}, the authors propose Fisher IPM, which is normalized IPM, given by,
$$
d_{\mathscr{F}}(\mathbb{P}, \mathbb{Q})=\sup _{f \in \mathscr{F}} \frac{\mathbb{E}_{\mathbb{P}}[f(x)]-\mathbb{E}_{x \sim \mathbb{Q}}[f(x)]}{\sqrt{1 / 2 \mathbb{E}_{x \sim \mathbb{P}} f^{2}(x)+1 / 2 \mathbb{E}_{x \sim \mathbb{Q}} f^{2}(x)}}
$$
Standardizing this discrepancy introduces as we will see a data dependent constraint, that controls the growth of the weights of the critic $f$ parameterized by $p$ and ensures the stability of the training while maintaining the capacity of the critic.
Learning GAN with Fisher IPM (empirical version):
\begin{equation}
    \label{eq:fishergan}
    \begin{aligned}
         \min _{g_{\theta}} \sup _{f_{p} \in \mathscr{F}_{p}} \hat{\mathcal{E}}\left(f_{p}, g_{\theta}\right) :=\frac{1}{N} & \sum_{i=1}^{N} f_{p}\left(x_{i}\right)-\frac{1}{M} \sum_{j=1}^{M} f_{p}\left(g_{\theta}\left(z_{j}\right)\right) \\ &\text { Subject to } \hat{\Omega}\left(f_{p}, g_{\theta}\right)=1
    \end{aligned}{}
\end{equation}{}
where $\hat{\Omega}\left(f_{p}, g_{\theta}\right)=\frac{1}{2 N} \sum_{i=1}^{N} f_{p}^{2}\left(x_{i}\right)+\frac{1}{2 M} \sum_{j=1}^{M} f_{p}^{2}\left(g_{\theta}\left(z_{j}\right)\right)$ Augmented Lagrangian Method for the final objective,

\begin{equation}
    \begin{aligned}{}
    \mathcal{L}_{F}(p, \theta, \lambda)=\hat{\mathcal{E}}\left(f_{p}, g_{\theta}\right)+  \lambda\left(1-\hat{\Omega}\left(f_{p}, g_{\theta}\right)\right)
     -\frac{\rho}{2}\left(\hat{\Omega}\left(f_{p}, g_{\theta}\right)-1\right)^{2}
    \end{aligned}
\end{equation}{}
In the final objective $f_p$ corresponds to $D_{\theta}$ and $g_{\theta}$ corresponds to $G_{\phi}$.
Fisher IPM will give rise to a whitened mean matching interpretation, or equivalently to mean matching with a Mahalanobis distance. Fisher IPM corresponds to Chi-squared distance when the critic has unlimited capacity.
\subsection*{Spectral Normalization for GANs \citep{spectral}}
In this paper the authors address the issue of vanishing gradient due to perfect discriminator by restricting the possible discriminators. The technique weight clipping in WGAN reduces the rank of the weight matrix and hence reduces the features used by the discriminator to distinguish the distributions. They also claim that in WGAN-GP, the support of the model distribution changes with training hence the effect of the previous regularization based on model samples is destabilized. Moreover it requires a lot of computation. The authors propose spectral normalization technique which does not effect the rank of the weight matrix. Unlike WGAN-GP, the regularization is not in the space of model samples. 

The authors introduce a sample dependant spectral normalization. Normalize the weight matrix in each layer $g^l: h_{in} \rightarrow h_{out}$ of the discriminator as follows, 
$$\hat{W}_{SN}^l(W^l) := W^l/\sigma(W^l)$$ where, 
$$\sigma(W^l) = \underset{\parallel h \parallel_2 \leq 1}{max} \parallel W^l h\parallel_2 = \text{largest singular values of} \  W^l $$

Power iteration method is used to estimate the singular values at each iteration to reduce the time complexity. The modified gradients w.r.t. to the objective $V(G, D)$ for the algorithm, given $u_1, v_1$ are the first left and right singular vectors respectively for the matrix $W^l$ 
\begin{equation}
    \label{eq:sn}
    \frac{\partial V(G, D) }{\partial W^l} = \frac{1}{\sigma(W^l)} (\hat{E} [\delta h_{in}^T] - \lambda u_1 v_1^T)
\end{equation}
where $\delta := (\partial V(G,D)/\partial(\hat{W}_SN h))^T, \ \ \lambda := \hat{E} [\delta^T (\hat{W}_{SN} h)]$
The first term in the Equation (\ref{eq:sn}) is same as the gradient of $W^l$ and the second term is as a result of normalization. $\lambda$ is positive when $\delta$ and $\hat{W}_SN^l h$ point in the same direction and then penalizes the first singular components hence prevents the transformation from becoming sensitive in one direction.
\\
\subsubsection{Regularization for Non-Convergence} We discuss works which modify the loss function by introducing a regularizer such that the new objective function converges to the global optima. The papers discussed provide rigorous theoretical analysis hence, would be a dealt with in further detail in Section \ref{subsec:theory}
\subsection*{The Numerics of GANs \citep{geiger06}} The authors investigate the non-convergence of simultaneous gradient descent based on the Jacobian of the gradients for both the $D$ and $G$. In Section \ref{subsec:optim} we discuss the further details. In order to over the non-convergence, they propose to add the following term as the regularizer to the loss. Given $ \begin{bmatrix}
\nabla_{\phi} G(\phi, \theta) \\
\nabla_{\theta} D(\phi, \theta) \\
\end{bmatrix}$
$$L(v(\theta, \phi)) = - \frac{1}{2} \parallel v(\theta, \phi) \parallel^2  $$
\subsection*{Gradient Descent GAN Optimization is Locally Stable \citep{CMU17}}
According to the convergence propoerties discussed in this paper, the authors propose the following reglarization penalty for the generator update.  Given $\theta_G = \phi$ and $\theta_D = \theta$ 
$$\theta_{G} := \theta_{G} - \alpha \nabla_{\theta_{G}} (V(D_{\theta_{D}}, G_{\theta_{G}})) + \eta \parallel \nabla_{\theta_{D}} V(D_{\theta_{D}}, G_{\theta_{G}}) \parallel $$
\subsection*{Which Training Methods for GANs do actually Converge? \citep{mescheder18icml}}

The authors claim the non-convergence of unregularized GANs. They claim WGAN-GP doesn't converge but noise induced regularizer as proposed in \citep{fgan_2} converges. The details are in Section \ref{subsubsec:theory_q3}. The authors suggest the following simplified gradient
penalty which is a simplified version as proposed in \citep{fgan_2}. This would ensure non-zero loss if there is non-zero gradients w.r.t. the discriminator in the orthogonal direction to the data manifold 

\begin{equation}
    R_{1}(\theta) :=\frac{\gamma}{2} \mathrm{E}_{p_d (x)}\left[\left\|\nabla D_{\theta}(x)\right\|^{2}\right]
\end{equation}{}
 To penalize the discriminator on the generator distribution $p_g$ obtained by $G_{\phi}$,
 \begin{equation}
R_{2}(\theta, \phi) :=\frac{\gamma}{2} \mathrm{E}_{p_g(\hat{x})}\left[\left\|\nabla D_{\theta}(\hat{x})\right\|^{2}\right]
\end{equation}  
Then they prove that with small learning rates, applying simultaneous gradient descent on the GAN objective with above regularizer is locally convergent.
\subsection*{On Convergence and Stability of GANs \citep{DRAGAN}}
The authors  view GANs objective as Regret minimization as opposed to divergence minimization. They make a connection between no regret algorithms and alternating SGD and prove it's convergence in convex-concave case as further discussed in Section \ref{subsubsec:theory_q3}. Besides they introduce an additional term in the objective which they refer to as local smoothing. According to their findings mode collapse is often accompanied by the discriminator function having sharp gradients around some real data points. Hence they introduce the following penalty term in the overall GAN loss and name the corresponding objective as DRAGAN, $$\lambda. \mathbb{E}_{x \sim p_{d}, \delta\sim N_{d}(0, cI)} [\parallel \nabla_{x}D_{\theta}(x + \delta) \parallel - k]^2 $$
\subsubsection{Biased Gradient Estimator} 
\label{subsubsec:loss_bge} In the previous sections papers introduced loss functions without any analysis for the estimated gradients of the loss. In this section, we discuss papers which claim that the previous loss (WGAN) would lead to biased gradients with finite samples. Hence, the following papers propose new loss to overcome this issue and ensure better convergence.
\subsection*{The Cramer Distance as a Solution to Biased Wasserstein Gradients \citep{cramer}} Wasserstein metric yields, from samples,  biased gradients for a fixed number of samples hence may not lead to convergence or may lead to wrong minimum. A very powerful critic is required to approximate the Wasserstein distance well. At the same time, a powerful critic would over-fit the empirical distribution, which is undesirable. The authors propose the Cramér distance or energy distance which has unbiased sample gradient. Moreover, energy distance enables learning with imperfect critic by combining with a transformation function $h$ as described below.

The authors first give a concept of \textit{Ideal divergence}, consider a divergence $d$
\begin{itemize}
    \item Scale sensitivity: if there exists a $\beta > 0$ such that    $\forall \ X, Y, \ c >0$ $$
d(c X, c Y) \leq|c|^{\beta}d(X, Y)
$$
\item Sum Invariant: if $A$ is independent from $X, Y$ then, $$
d(A+X, A+Y) \leq d(X, Y)
$$
\item Unbiased gradient estimator: Let $X_m =  X_1, X_2, \ldots X_m$ be samples from $P$ and $\hat{P}_m := \frac{1}{m}\sum_{i=1}^m \delta_{X_i}$
$$
\underset{\mathbf{X}_{m \sim P}}{\mathbb{E}} \nabla_{\theta} d\left(\hat{P}_{m}, Q_{\theta}\right)=\nabla_{\theta} d\left(P, Q_{\theta}\right)
$$
\end{itemize}{}
Cramer or energy distance for multivariate case. Let $X, X'$ and $Y, Y'$ be independent random variables distributed according to $P, Q$. Let $h$ be the transformation function
\begin{equation}
\label{eq:eg}
\begin{aligned}
\mathcal{E}(P, Q) &:=\mathcal{E}(X, Y) \\
&:=2 \mathbb{E}\|X-Y\|_{2}-\mathbb{E}\left\|X-X^{\prime}\right\|_{2}-\mathbb{E}\left\|Y-Y^{\prime }\right\|_{2}
\end{aligned}{}
\end{equation}
The Cramer GAN objective
$$\underset{G}{min} \ \underset{D_h}{max}\ \mathcal{E}(h(X), h(Y))  $$

In our notations, $h$ is $D$ parametrized by $\theta$ and $Y \sim G_{\phi}$, the objective is given by,
The Cramer GAN objective
$$\underset{\phi}{min} \ \underset{\theta}{max}\ \mathcal{E}(D_{\theta}(X), D_{\theta}(Y))  $$

\subsection*{Learning Generative models with Sinkhorn Divergences \citep{sinkhorn}}
The authors propose an optimal transport based Sinkhorn Divergence which is differentiable and tractable. They claim that estimating distances between two distributions with non-overlapping support is difficult with MLE. Hence weaker metrics are derived through duality. Given the dual norm $\mathcal{L}(\mu, \nu)=| \mu-\nu \|_{B}^{*}$ , $B$ a unit ball of continuous functions
    $$
\|\xi\|_{B}^{*}=\sup \left\{\int_{\mathcal{X}} h(x) \mathrm{d} \xi(x) ; h \in B\right\}
$$
There are two instances of this,
\begin{itemize}
    \item Wasserstein GAN : $B$ is the set of 1-Lipschitz functions.
    \item Maximum Mean Discrepency Losses
    $$
\begin{aligned} | \mu, \nu \|_{k} &=\mathbb{E}_{\mu \otimes \mu}\left[k\left(X, X^{\prime}\right)\right]+\mathbb{E}_{\nu \otimes \nu}\left[k\left(Y, Y^{\prime}\right)\right] \\ &-2 \mathbb{E}_{\mu \otimes \nu}[k(X, Y)] \end{aligned}
$$
\end{itemize}{}

The authors propose a different divergence based on Optimal Transport (OT) Metrics. OT supported on two metric spaces $(\mu, \nu) \in \mathcal{M}_{+}^{1}(\mathcal{X}) \times \mathcal{M}_{+}^{1}(\mathcal{X})$ is given by,
   $$
\mathcal{W}_{c}(\mu, \nu) \stackrel{\mathrm{def.}}{=} \min _{\pi \in \Pi(\mu, \nu)} \int_{\mathcal{X} \times \mathcal{X}} c(x, y) \mathrm{d} \pi(x, y)
$$
Challenges with OT:
    \begin{itemize}
        \item Computational burden of evaluating OT losses
        \item Lack of smoothness
        \item difficult to estimate their gradients in high dimension
    \end{itemize}{}
To overcome the challenges they define Sinkhorn distance which includes the good properties of both OT and MMD-GAN.
\begin{itemize}
    \item They introduce entropic smoothing which makes the loss differentiable
\item Compute the Sinkhorn distance using Sinkhorn fixed point iterations with GPU execution
\item By changing the smoothing parameter $\epsilon$ from $0$ to $\infty$ the proposed loss transforms from pure OT loss loss to MMD like loss. 
\item Good properties from MMD: i) favourable sample complexity
ii) unbiased gradient estimates (empirically)
\item good properties from OT: i) can be defined for any $c$ whereas MMD is defined for positive $k$
\end{itemize}
\noindent The authors introduce a regularized optimal transport problem
with cost $c$ and regularization parameter $\epsilon$
\begin{equation}
\mathcal{W}_{c, \varepsilon}(\mu, \nu)=\int c(x, y) \mathrm{d} \pi_{\varepsilon}(x, y)
\end{equation}

where $\pi_{\epsilon}$ is given by,

\begin{equation}
\min _{\pi \in \Pi(\mu, \nu)} \int c(x, y) \mathrm{d} \pi(x, y)+\varepsilon \int \log \left(\frac{\pi(x, y)}{\mathrm{d} \mu(x) \mathrm{d} \nu(y)}\right) \mathrm{d} \pi(x, y)
\end{equation}
The following theorem gives the Sinkhorn loss
\begin{theorem}
The Sinkhorn loss between two measures $\mu, \nu$ is 
$$
\overline{\mathcal{W}}_{c, \varepsilon}(\mu, \nu)=2 \mathcal{W}_{c, \varepsilon}(\mu, \nu)-\mathcal{W}_{c, \varepsilon}(\mu, \mu)-\mathcal{W}_{c, \varepsilon}(\nu, \nu)
$$
$$
\begin{array}{l}{\text { 1. as } \varepsilon \rightarrow 0, \quad \overline{\mathcal{W}}_{c, \varepsilon}(\mu, \nu) \rightarrow 2 \mathcal{W}_{c}(\mu, \nu)} \\ {\text { 2. } \text {as } \varepsilon \rightarrow+\infty, \quad \overline{\mathcal{W}}_{c, \varepsilon}(\mu, \nu) \rightarrow M M D_{-c}(\mu, \nu)}\end{array}
$$
where $MMD_{-c}$ uses the kernel as $c$
\end{theorem}{}
The generative model is interested in $$
\min _{\theta} E_{\varepsilon}(\theta) \quad \text { where } \quad E_{\varepsilon}(\theta) \stackrel{\text { det }}{=} \overline{\mathcal{W}}_{c, \varepsilon}\left(\mu_{\theta}, \nu\right)
$$
Estimating the gradients of $E_{\varepsilon}$ is difficult. Hence the authors approximate the loss by $L$ steps of Sinkhorn algorithm and obtain algorithmic loss $\hat{E}_{\varepsilon}^{(L)}(\theta)$

The choice of cost $c$,
$$
c_{\varphi}(x, y) \stackrel{\mathrm{def} .}{=}\left\|f_{\varphi}(x)-f_{\varphi}(y)\right\| \quad \text { where } \quad f_{\varphi} : \mathcal{X} \rightarrow \mathbb{R}^{p}
$$
Thus the final objective is given by,
$$
\min _{\theta} \max _{\varphi} \overline{\mathcal{W}}_{c_{\varphi}, \varepsilon} \left(\mu_{\theta}, \nu\right)
$$
We can view the $f_{\varphi}$ as the discriminator $D$ mapping the data to a feature vector of dimension $p$ and the generator $p_g$ is given by the distribution $\mu_{\theta}$
\subsection*{Improving GANs using Optimal Transport \citep{ot}}
Propose a new distance metric which measures the distance between model and data distribution. It is highly discriminative with unbiased mini-batch gradients. In WGAN, to compute the dual given by Equation (\ref{eq:em_dual}), the discriminator ideally has to optimize over all possible 1-Lipschitz functions which is not likely in finite steps. This leads to imperfect discriminator which would not approximate the actual distance given by Equation (\ref{eq:em}) well. In the paper \citep{sinkhorn} Sinkhorn distance is proposed, is fully tractable hence overcomes the above problem. Yet this results in an biased estimator of the actual distance between the two distributions. The authors here propose a new distance combines optimal transport in primal form with an energy distance defined in an adversarially learned feature space, resulting in a highly discriminative distance function with unbiased mini-batch gradients as described further below.
Propose a metric Mini-batch Energy Distance
\begin{itemize}
    \item Entropically smooth earth mover distance called Sinkhon distance
    \begin{equation}
        \label{eq:sinkhorn1}
        D_{\text { sinkhorn }}(p, g)=\inf _{\gamma \in \Pi_{\beta}(p, g)} \mathbb{E}_{x, x \sim \gamma} c(x, y)
    \end{equation}
    The set of allowed joint distributions are restricted to distributions with entropy at least $\beta$. This is evaluated for a mini-batch of $K$ data vectors.  The cost function $c$ gives rise to a cost matrix $C$, where $C_{i,j} = c(x_i, y_j)$. Similarly, $\gamma$ is replaced by $K \times K$ matrix $M$ of soft matchings, with sufficient entropy i.e., $-\operatorname{Tr}\left[M \log \left(M^{\mathrm{T}}\right)\right] \geq \alpha$ The resulting distance is evaluated as follows,
    \begin{equation}
    \label{eq:sinkhorn}
        \mathcal{W}_{c}(X, Y)=\inf _{M \in \mathcal{M}} \operatorname{Tr}\left[M C^{\mathrm{T}}\right]
    \end{equation}{}
The gradients of a fixed mini-batch for the Equation (\ref{eq:sinkhorn}) is not an unbiased estimator of the gradients of the Equation (\ref{eq:sinkhorn1})

    \item Generalized energy distance: The distance combines optimal transport in primal form, with an energy distance in adversarially learned feature space that has unbiased mini-batch gradients. Given a distance function $d$ and $X,X^{\prime} \in p$ and $Y, Y^{\prime} \in g$
    \begin{equation}
        \begin{aligned}
        D_{\mathrm{GED}}^2(p, g)=2 \mathbb{E}[d(X, Y)]- \mathbb{E}\left[d\left(X, X^{\prime}\right)\right] 
        -\mathbb{E}\left[d\left(Y, Y^{\prime}\right)\right]
        \end{aligned}
    \end{equation}
    \item Minibatch energy distance 
    The following is the distance metric proposed, the terms which are additional to Equation (\ref{eq:sinkhorn1}) makes the gradient estimator unbiased.
    \begin{equation}
        \begin{aligned}
        D_{\mathrm{MED}}^{2}(p, g)=2 \mathbb{E}\left[\mathcal{W}_{c}(X, Y)\right]-\mathbb{E}\left[\mathcal{W}_{c}\left(X, X^{\prime}\right)\right] 
        -\mathbb{E}\left[\mathcal{W}_{c}\left(Y, Y^{\prime}\right)\right]
        \end{aligned}
    \end{equation}

    \item Transport cost function which is learnt adversarially. Given $v_{\eta}$ is the discriminator which maps the samples to a latent space. The following is $1 -$ (cosine similarity between latent representation of data and generated sample)
    $$
c_{\eta}(x, y)=1-\frac{v_{\eta}(x) \cdot v_{\eta}(y)}{\left\|v_{\eta}(x)\right\|_{2}\left\|v_{\eta}(y)\right\|_{2}}
$$
\end{itemize}{}
\subsection*{Demystifying MMD GANs \citep{mmd}} Wasserstein distance can lead to biased gradients for the generator,and gave an explicit example where optimizing with these biased gradients leads the optimizer to incorrect parameter values, even in expectation. The authors show (Theorem 1) that the natural maximum mean discrepancy estimator, including the estimator of energy distance, has unbiased gradients when used “on top” of a fixed deep network representation. 
The MMD distance between two distributions $\mathbb{P}, \mathbb{Q}$ is given by,
$$
\operatorname{MMD}(\mathbb{P}, \mathbb{Q} ; \mathcal{H})=\sup _{f \in \mathcal{H},\|f\|_{\mathcal{H}} \leq 1} \mathbb{E}_{\mathbb{P}} f(X)-\mathbb{E}_{\mathbb{Q}} f(Y)
$$
where $f \in \mathcal{F}$ is a function class in a kernel Hilbert space $\mathcal{H}$. This situation is exactly analogous to WGANs: the generator’s gradients with a fixed critic are unbiased, but gradients from a learned critic are biased with respect to the supremum over critic. The authors clarify why MMD GANs are in some sense ``less biased'' than WGAN.

\subsubsection{Resolve Non-Convergence} 
\label{subsubsec:loss_nc}
In this subsection, the papers try to address Q$_2$ and Q$_3$ i.e., modify the loss function for better convergence. For the paper which introduce a generalized frameworks for the adversarial loss function with an aim to prove convergence, we refer the reader to Sections \ref{subsubsec:theory_q2} and \ref{subsubsec:theory_q3}.

\subsection*{Improved Techniques for Training GANs \citep{goodfellow16}}
As seen in Subsection \ref{subsubsec:loss_vg} in this paper authors introduce different heuristics like feature matching and minibatch discrimination. Here we are more interested in looking at the aspects of the paper that modifies the loss to prevent non-convergence. 
The introduce the notion of \emph{fictitious play} where each player's cost is modified by including a term $\parallel \theta - \frac{1}{t} \sum_{i=1}^{t} \theta[i] \parallel^2$, where $\theta[i]$ is the values of parameters at a past time $i$. This approach is inspired by fictitious play algorithm that can find equilibrium in different kinds of games including non-convex, continuous games. It has been observed that introducing batch normalization leads to stabler optimization in \citep{dcgan}. The authors claim that this results in the output corresponding to a particular input becomes dependant on the samples within the same batch. To overcome this they introduce \emph{virtual batch normalization} where an input sample is normalized based on a reference batch which is fixed at the start of the training itself. This procedure is computationally expensive hence used in $G$ only. Besides, the authors also use one-sided label smoothing by smoothing the positive labels only for better training.

\subsection*{Unrolled GANs \citep{unrolled}}
As in Proposition \ref{prop:conv}, the discriminator has to be optimal at every iteration for simultaneous gradient descent to converge which cannot be achieved in practice.
For a minimax loss as given in \citep{ganGoodfellow}, the optimal discriminator $D^*(x)$ is a known smooth function of the generator probability $p_{g}(x).$ These smoothness guarantees are lost when $D(x;\theta_{D})$ and $G(x;\theta_{G})$ are drawn from parametric families. Note here $\phi = \theta_G$ and $\theta = \theta_D$. Explicitly solving for the optimal discriminator parameters $\theta^*_{D}(\theta_{G})$ for every update step of the generator G is computationally infeasible. As a result GAN training suffers from mode collapse. A surrogate loss function $f_{K}(\theta_{G}, \theta_{D})$ is introduced for training the generator which more closely resembles the true generative objective $f(\theta_{G}, \theta^*{D}(\theta_{G}))$. $K=0$ (Normal Gan loss), $K \rightarrow \infty$ (True generative objective function). 
The gradient updates: $$\theta_{G} \leftarrow \theta_{G} - \eta \frac{df_{K}(\theta_{G}, \theta_{D})}{d \theta_{G}}, \mbox{ }\theta_{D} \leftarrow \theta_{D} + \eta \frac{df(\theta_{G}, \theta_{D})}{d \theta_{D}} $$

\begin{equation*}
    \begin{aligned}
    \frac{df_{K}(\theta_{G}, \theta_{G})}{d\theta_{G}} = \frac{\partial f(\theta_{G}, \theta^K_{D}(\theta_{G}, \theta_{D}))}{\partial \theta_{G}} 
    \frac{\partial f(\theta_{G}, \theta^K_{D}(\theta_{G}, \theta_{D}))}{\partial \theta^K_{D}(\theta_{G}, \theta_{D}) } \frac{ d\theta^K_{D}(\theta_{G}, \theta_{D})}{d\theta_{G}}
    \end{aligned}{}
\end{equation*}{}

The authors directly addresses the issue of mode collapse. They suggest a new loss for overcoming this problem. If one agent becomes more powerful than the other, the learning signal becomes useless. In a standard GAN the G tries to move as much mass to a single point that maximizes the ratio of the probability density. The D tracks the point and assigns lower probability to it and uniform elsewhere. This cycle will repeat forever. In this paper, however, using the surrogate loss, G's update takes into account the response of D before hand. This helps G to spread it's mass making the next D step less effective instead of collapsing to a point.
\subsection*{Coloumb GANs:Provably Optimal Nash Equilibria via Potential Fields \citep{coulomb}}
Propose Coulomb GANs which pose the learning problem as a potential field. Further it is proven to have one Nash equilibrium at which the model distribution would equal the data distribution. GANs convergence points are local Nash equilibria causing mode collapse. They propose Coloumb GAN which has only one Nash equilibria and the one which is optimal (data distribution $=$ model distribution)

\begin{itemize}
    \item Potential function: the influence of a potential at $\boldsymbol{b}$ on $\boldsymbol{a}$ given a kernel $k(\boldsymbol{a},\boldsymbol{b})$ and $\rho(\boldsymbol{a})=p_{y}(\boldsymbol{a})-p_{x}(\boldsymbol{a})$
    $$
\Phi(\boldsymbol{a})=\int \rho(\boldsymbol{b}) k(\boldsymbol{a}, \boldsymbol{b}) \mathrm{d} \boldsymbol{b}
$$
    In order for there to be single Nash Equilibrium, they use Plummer kernel, where $d \leq m- 2$ given by 
    $$
k(\boldsymbol{a}, \boldsymbol{b})=\frac{1}{\left(\sqrt{\|\boldsymbol{a}-\boldsymbol{b}\|^{2}+\epsilon^{2}}\right)^{d}}
$$
    \begin{itemize}
        \item Given that $\rho(\boldsymbol{b}) = 0$ then $\Phi(\boldsymbol{a}) = 0$ for all $\boldsymbol{a}$.
        \item $\Phi(\boldsymbol{a}) = 0$ should imply that $\rho(\boldsymbol{b}) = 0$ Given the energy function
 \begin{equation*}
     \begin{aligned}
     F(\rho)= \frac{1}{2} \int \rho(\boldsymbol{a}) \Phi(\boldsymbol{a}) \mathrm{d} \boldsymbol{a}  =\frac{1}{2} \iint \rho(\boldsymbol{a}) \rho(\boldsymbol{b}) k(\boldsymbol{a}, \boldsymbol{b}) \mathrm{d} \boldsymbol{b} \mathrm{d} \boldsymbol{a}
     \end{aligned}{}
 \end{equation*}{}

        \begin{theorem} (Convergence with low dimensional Plummer kernel) For $a,b \in \mathbb{R}^m$, $d \leq m-2$, $\epsilon > 0$ the densities $p_x$ and $p_y$ equalize over time when minimizing the energy $F$ with the low dimensional Plummer kernel by gradient descent. The convergence is faster for larger $d$
        \end{theorem}
    \end{itemize}
    \item The discriminator should learn $\hat{\Phi}(\boldsymbol{a})=\frac{1}{N_{y}} \sum_{i=1}^{N_{y}} k\left(\boldsymbol{a}, y_{i}\right)-\frac{1}{N_{x}} \sum_{i=1}^{N_{x}} k\left(\boldsymbol{a}, x_{i}\right)$ Hence the objective function is given by
   $$
\begin{aligned} \mathcal{L}_{D}(D ; G) &=\frac{1}{2} \mathrm{E}_{p_{a}}\left((D(\boldsymbol{a})-\hat{\Phi}(\boldsymbol{a}))^{2}\right) \\ \mathcal{L}_{G}(G ; D) &=-\frac{1}{2} \mathrm{E}_{p_{\boldsymbol{z}}}(D(G(\boldsymbol{z}))) \end{aligned}
$$
where $p(\boldsymbol{a})=1 / 2 \int \mathcal{N}(\boldsymbol{a} ; G(\boldsymbol{z}), \epsilon I) p_{z}(z) \mathrm{d} z+1 / 2 \int \mathcal{N}(\boldsymbol{a} ; y, \epsilon I) p_{y}(y) \mathrm{d} y$
    \item Thus we may not find the optimal $G^*, \ D^*$ , since  neural networks may suffer from capacity or optimization issues
    \item The main problem with learning Coulomb GANs is to approximate the potential function $\Phi$, which is a complex function in a high-dimensional space, since the potential can be non-linear and non-smooth.  When learning the discriminator,  we must ensure that enough data is sampled and averaged over.
\end{itemize}{}
%%%%%%%%%%%%%%%%%%%%%%%%%%%%%%%%%%%%%%%%%%%%%%%%%%%%%%%%%%%%%%%%%%
\subsection{Architecture} \label{subsec:arch}
In this section we discuss the papers which bring about major architectural changes. In some cases they also modify the loss and/or introduce regularizers or based on other heuristics. 
\subsection*{Unsupervised Representation Learning with Deep Convolutional GANs \citep{dcgan}} In this paper, authors introduced architectural changes to vanilla GAN and set other parameters of learning rate which helped stabilize training significantly. They primarily introduced convolutional layers. Other architectural guidelines are as follows,
\begin{itemize}
    \item The pooling layers are replaced with strided convolutions in $G$ and fractional-strided convolutions in $D$.
    \item Batchnorm layers are used both in $G$ and $D$.
    \item  ReLU is used in all layers of $G$ except the last output which uses tanh.
    \item LeakyReLU in the discriminator.
    \item Adam optimizer is used \citep{adam} with learning rate of $0.0002$ and a momentum $\beta_1$ of 0.9.
\end{itemize}{}
They name the model DCGAN.
Although, the above changes have no particular theoretical basis, they seem to work well in practice. In the rest of section, we will focus on papers which modify the architecture of GANs based on theoretical analysis.There are primarily three ways in which researchers have tried to modify the architecture as listed below. 
\subsubsection{Auto-encoder Architectures} \label{subsubsec:arch_ae} The most popular modification is based on having an auto-encoder architecture within the network.

\subsection*{Adversarially Learned Inference \citep{ali}}
The authors aim to incorporate the inference mechanism like that of a VAE in GANs. The approximate inference in VAE suffers from various drawbacks as disscussed in \ref{sec:success}. Hence the authors propose to use adversarial learning for inference which can be further used in other tasks like semi-supervised learning and inpainting. 

The set up the objective to match the following two joint distributions,
\begin{itemize}
    \item[i] $q(x, z) = p_d(x)\  q(z|x)$
    \item[ii] $p(x, z) = p_z(z)\  p(x|z)$
\end{itemize}{}
The $p(x|z)$ is learnt by the generator network $G_x$ which takes in $z \in \mathcal{N} (0, I)$ as input and gives $\hat{x}$ as output. $q(z|x)$ is learnt by the inference network $G_z$ which takes in $x\sim p_d$ and outputs a $\hat{z}$. In order to match the joint distributions, the $D$ is trained to discriminated between the joint $(x,z)$ samples while both the generator and inference network try to fool it.
The loss function used is given by,

\begin{equation}
\begin{aligned} \min _{G} \max _{D} V(D, G) &=\mathbb{E}_{q(\boldsymbol{x})}\left[\log \left(D\left(\boldsymbol{x}, G_{z}(\boldsymbol{x})\right)\right)\right]+\mathbb{E}_{p(\boldsymbol{z})}\left[\log \left(1-D\left(G_{x}(\boldsymbol{z}), \boldsymbol{z}\right)\right)\right] \\ &=\iint q(\boldsymbol{x}) q(\boldsymbol{z} | \boldsymbol{x}) \log (D(\boldsymbol{x}, \boldsymbol{z})) d \boldsymbol{x} d \boldsymbol{z} \\ &+\iint p(\boldsymbol{z}) p(\boldsymbol{x} | \boldsymbol{z}) \log (1-D(\boldsymbol{x}, \boldsymbol{z})) d \boldsymbol{x} d \boldsymbol{z} \end{aligned}
\end{equation}

Typically $q(z|x)$ is assumed to be $\mathcal{N}(\mu(x), \sigma^2(x)I)$ and to sample from this distribution the reparameterization trick \citep{kingma2014} is used,
$z=\mu(x)+\sigma(x) \odot \epsilon, \quad \epsilon \sim \mathcal{N}(0, I)$
 To avoid the vanishing gradient issue, generator maximizes the following loss,
 $$
V^{\prime}(D, G)=\mathbb{E}_{q(x)}\left[\log \left(1-D\left(x, G_{z}(x)\right)\right)\right]+\mathbb{E}_{p(z)}\left[\log \left(D\left(G_{x}(z), z\right)\right)\right]
$$
The authors also provide convergence analysis for the joint distributions similar to the analysis provided in \cite{ganGoodfellow}. They further prove that at optimality $G_x = G_z^{-1}$ and $G_z = G_x^{-1}$ almost everywhere.

\subsection*{Mode Regularized GANs \citep{modeReg}}
Che \textit{et al.} \citep{modeReg} give an intuition behind the problem of missing modes and also propose regularizers to circumvent this problem. 
It is usually the case that the data and the model distribution manifolds are disjoint. In such a case, the discriminator assigns zero probability to all the model points and one probability to all the data points. Thus, large modes usually have a much higher chance of attracting the gradient of the discriminator. For a typical GAN model, since all modes have similar D values, there is no reason why the generator cannot collapse to just a few major modes. For most $z$ the gradient of the generator pushes the generator distribution towards the major mode. It is highly unlikely to have $z$ which is close to the other minor modes, hence the problem of missing modes.
\begin{itemize}
\item Geometric Metric Regularizer - Having another similarity metric such as $L_{2}$-norm with nice geometric properties, in addition to the gradient information from the discriminator. Together with the $G$, they also have an Encoder $E(x) : X \rightarrow Z$. Assuming $d$ to be some similarity metric in the data space, the authors add the following term to the loss as a regularizer, $$\mathbb{E}_{x \sim p_{d}}[d(x, G\circ E(x))]$$
The encoder is trained by minimizing the reconstruction error.
\item Mode Regularizer - This is proposed to penalize the missing modes.
\begin{itemize}
\item The areas near the missing modes are rarely visited by the $G$
\item Both missing modes and non-missing modes correspond to high values of $D$. 
\end{itemize}
Consider a minor mode $M_{0}$. For $x\in M_{0}$, $G(E(x))$ will be located close to the mode $M_{0}$. They add the following to the loss,  $$\mathbb{E}_{x\sim p_{d}}[\log D(G \circ E(x))]$$
\end{itemize}
The overall loss for $G$ is given by, 
\begin{equation*}
    \begin{aligned}
    T_{G} = - \mathbb{E}_{z}[\log D(G(z))] 
    + \mathbb{E}_{x \sim p_{d}}[\lambda_{1} d(x, G\circ E(x)) + \lambda_{2}\log D(G \circ E(x))]
    \end{aligned}{}
\end{equation*}{}

The overall loss for $E$ is given by, 
$$T_{E} = \mathbb{E}_{x \sim p_{d}}[\lambda_{1} d(x, G\circ E(x)) + \lambda_{2}\log D(G \circ E(x))]$$

\subsection*{Energy-Based GAN \citep{eb-gan}}
Another encoder-decoder based approach for training GAN was put forth in the Energy-based GAN (EBGAN) paper by \citep{eb-gan} Zhao \emph{et. al.}. The paper views the discriminator as an energy function, which assigns low energy values to real data and high to fake data. The generator is a trainable parameterized function that produces samples in regions to which the discriminator assigns low energy. The objective function is given by, $$\mathcal{L}_{D}(x,z) = D(x) + [m-D(G(z))]^+$$$$\mathcal{L}_{G}(z) = D(G(z))$$ where, $[.]^+ = max(0,.)$; $m$ - positive margin; $\mathcal{L}_{D}$ - discriminator loss; $\mathcal{L}_{G}$ - generator loss
The discriminator is modeled as an auto-encoder  $$D(x) = \parallel Dec(Enc(x)) - x \parallel $$
 With the binary logistic loss, only two targets are possible, so within a minibatch, the gradients corresponding to different samples are most likely far from orthogonal. This leads to inefficient training, and reducing the minibatch sizes is often not an option on current hardware. According to the paper, the reconstruction loss introduced will likely produce very different gradient directions within the minibatch, allowing for larger minibatch size without loss of efficiency. When an EBGAN auto-encoding model is trained to reconstruct a real sample, the discriminator contributes to discovering the data manifold by itself without the need for explicit negative samples. To prevent the auto-encoder from learning identity function, the framework is regularized with the generator producing contrastive samples. A Repelling regularizer is introduced to prevent mode collapse used only with the generator loss, $S \in \mathbb{R}^{s\times N}$ where $S$ is a batch of sample representations taken from encoder output layer. $$f_{PT}(S) = \frac{1}{N(N-1)}\sum_{i} \sum_{j\neq i} \bigg( \frac{S_{i}^T S_{j}}{\parallel S_{i}\parallel \parallel S_{j}\parallel} \bigg)^2 $$ The $PT$ term attempts to orthogonalize the pairwise sample representations.

\subsection*{Boundary equilibrium GANs \citep{began}}
The authors propose an auto-encoder based GAN (BEGAN) with a loss derived from Wasserstein distance, in order to balance training between the generator and discriminator and provide new approximate convergence measure. They raise threefold issues, i) Balancing the training between discriminator and generator is difficult. ii)  Controlling the sample diversity is difficult. iii) Determining convergence in GANs is difficult just by the losses. The method proposed has three components to it, the architecture, the diversity ratio and global convergence measure to address the three issues,
\begin{itemize}
    \item The generator instead of trying to minimize the Wasserstein distance between data sample and model sample, it minimizes the difference in the auto-encoder loss corresponding to these samples. 
    The autoencoder loss is given by
    $$
\mathcal{L}(v)=|v-D(v)|^{\eta} \text { where }\left\{\begin{array}{ll}{D : \mathbb{R}^{N_{x}} \mapsto \mathbb{R}^{N_{x}}} \\ {\text { is the autoencoder function. }} \\ {\eta \in\{1,2\}} \\ {\text { is the target norm. }} \\ {v \in \mathbb{R}^{N_{x}}} \\ {\text { is a sample of dimension } N_{x}}\end{array}\right.
$$
    The objective is given by,
    \begin{equation}
    \label{eq:began}
\left\{\begin{array}{ll}{\mathcal{L}_{D}=\mathcal{L}\left(x ; \theta_{D}\right)-\mathcal{L}\left(G\left(z_{D} ; \theta_{G}\right) ; \theta_{D}\right)} & {\text { for } \theta_{D}} \\ {\mathcal{L}_{G}=-\mathcal{L}_{D}} & {\text { for } \theta_{G}}\end{array}\right.
\end{equation}
\item Diversity ratio is given by,
$$
\gamma=\frac{\mathbb{E}[\mathcal{L}(G(z))]}{\mathbb{E}[\mathcal{L}(x)]}
$$
$\gamma =1$ at equilibrium. Lower values of $\gamma$ will lead to lower image diversity since the discriminator focuses on auto-encoding images. The following procedure helps to balance the training between the generator and discriminator,
$$
\left\{\begin{array}{ll}{\mathcal{L}_{D}=\mathcal{L}(x)-k_{t}, \mathcal{L}\left(G\left(z_{D}\right)\right)} & {\text { for } \theta_{D}} \\ {\mathcal{L}_{G}=\mathcal{L}\left(G\left(z_{G}\right)\right)} & {\text { for } \theta_{G}} \\ {k_{t+1}=k_{t}+\lambda_{k}\left(\gamma \mathcal{L}(x)-\mathcal{L}\left(G\left(z_{G}\right)\right)\right)} & {\text { for each }} \\
{} & {\text{step}\  t}\end{array}\right.
$$
\item Convergence measure
$$
\mathcal{M}_{g l o b a l}=\mathcal{L}(x)+\left|\gamma \mathcal{L}(x)-\mathcal{L}\left(G\left(z_{G}\right)\right)\right|
$$
\end{itemize}

\subsection*{VEEGAN: Reducing Mode Collapse in GANs Using Implicit Variational Learning \citep{veegan}}
The authors address the major issue of mode collapse. They propose, having a reconstructor network which is trained to achieve two-fold objective i) Mapping samples from true distribution to a Gaussian ii) Mapping the generated samples also to the normal distribution (assuming $z$ is drawn from normal). Intuitively this is achievable only when the generated samples follow true distribution. The objective is given by,
$$\mathcal{O}_{entropy}(\gamma, \theta) = E[\parallel z - F_{\theta}(G_{\gamma}(z)) \parallel_2^2] + H(Z, F_{\theta}(X))$$
where $\gamma:$ parameters of the generator, $F_{\theta}:$ reconstructor network parameterized by $\theta$, $H:$ cross entropy loss.  \\
Tractable solution: 
\begin{equation*}
    \begin{aligned}
    \mathcal{O}_{entropy}(\gamma, \theta) =  KL [q_{\gamma}(x|z) p_0(z) || p_{\theta}(z|x) p(x)] 
    - E[\log p_0 (z)] + E[d( z , F_{\theta}(G_{\gamma}(z))]
    \end{aligned}{}
\end{equation*}{}

\subsection*{Variational approaches for auto-encoding GANs \citep{variational}}
The authors propose a model combining GANs and Variational Auto-encoders to help reduce mode collapse. The authors develop a principled approach for AE-GAN. They start with VAE objective and introduce adversarial loss as required.
\begin{itemize}
    \item Maximizing likelihood and the lower bound in VAE: given $q_{\eta}(z | x) $ is the variational distribution over the latent variables $z$
\begin{equation}
    \label{eq:vae}
    \begin{aligned}
    \log p_{\theta}(x)=\log \int p_{\theta}(x | z) p(z) d z 
    \geq \mathbb{E}_{q_{\eta}(z | x)}\left[\log p_{\theta}(x| z)\right]-\operatorname{KL}\left[q_{\eta}(z| x) \| p(z)\right]
    \end{aligned}{}
\end{equation}
\item Implicit variational distributions - in a VAE we need to typically assume the form of $q_{\eta}(z | x) $ like Gaussian etc, but with GANs we can learn the distribution implicitly, by replacing the second term in Equation (\ref{eq:vae}) with
\begin{equation*}
    \begin{aligned}
    -\mathrm{KL}\left[q_{\eta}(z | x) \| p(z)\right]  =\mathbb{E}_{q_{\eta}(z | x)}\left[\log \frac{p(z)}{q_{\eta}(z | x)}\right] 
     \approx \mathbb{E}_{q_{\eta}(z | x)}\left[\log \frac{\mathcal{C}_{\omega}(z)}{1-\mathcal{C}_{\omega}(z)}\right]
    \end{aligned}{}
\end{equation*}{}

\item Likelihood choice : Here again we can make an explicit choice of the likelihood or if intractable we introduce synthetic likelihood and discriminator
\begin{itemize}
    \item Explicit likelihood: zero mean Laplace distribution as in (AGE, BEGAN, cycleGAN and PPGN). Then replace the first term in Equation (\ref{eq:vae}) with
    $$
\mathbb{E}_{q_{\eta}(z | x)}\left[-\lambda\left\|x-\mathcal{G}_{\theta}(z)\right\|_{1}\right]
$$
\item The other option is to introduce the concept of synthetic likelihood by multiplying and dividing by $p^*(x)$ which gives us the following,
\begin{equation*}
    \begin{aligned}
    \mathbb{E}_{q_{\eta}(z | x)}\left[\log p_{\theta}(x | z)\right]=&\mathbb{E}_{q_{\eta}(z | x)}\left[\log \frac{p_{\theta}(x | z)}{P^{*}(x)}\right] \\
    & \mathbb{E}_{q_{\eta}(z | x)}\left[\log p^{*}(x)\right]
    \end{aligned}{}
\end{equation*}{}
Then we use the GAN loss as follows for the above term,
$$
\mathbb{E}_{q_{\eta}(z| x)}\left[\log \frac{\mathcal{D}_{\phi}\left(\mathcal{G}_{\theta}(z)\right)}{1-\mathcal{D}_{\phi}\left(\mathcal{G}_{\theta}(z)\right)}\right]
$$
\end{itemize}
\item The overall Hybrid loss is as follows,
\begin{equation}
\label{eq:ae_gan}   
\begin{aligned}
\mathcal{L}(\theta,\eta)= \mathbb{E}_{q_{\eta}(z | x)}\bigg[-\lambda\left\|x-\mathcal{G}_{\theta}(z)\right\|_{1} 
+ \log \frac{\mathcal{D}_{\phi}\left(\mathcal{G}_{\theta}(z)\right)}{1-\mathcal{D}_{\phi}\left(\mathcal{G}_{\theta}(z)\right)}+\log \frac{\mathcal{C}_{\omega}(z)}{1-\mathcal{C}_{\omega}(z)}\bigg]
\end{aligned}{}
\end{equation}
The algorithm has alternative updates between $\theta, \eta, \phi, \omega$
\item Modified Equation (\ref{eq:ae_gan}) for Non saturating generator loss,
\begin{equation*}
    \begin{aligned}
    \mathbb{E}_{q_{\eta}(z | x)}[\lambda\left\|x-\mathcal{G}_\theta(z)\right\|_{1}-\log \mathcal{D}_{\phi}\left(\mathcal{G}_{\theta}(z)\right)  
    + \log \left(1-\mathcal{D}_{\phi}\left(\mathcal{G}_{\theta}(z)\right)\right)]
    \end{aligned}{}
\end{equation*}{}
\end{itemize}

\subsection*{MMD GAN: Towards Deeper Understanding of Moment Matching Network \citep{mmd-gan}}
The authors improvise on the Generative Moment Matching Network (GMMN) proposed in \citep{gmmn} by introducing adversarial loss leading to MMD-GAN objective. The authors claim that the empirical results of GMMN are not comparable with GANs and requires large mini-batch for training. The convergence results of GMMN may not hold given the loss is empirically estimated.

The primary difference between GANs and GMMN is that in GMMN instead of a  discriminator there is a two sample test based on kernel maximum mean discrepancy (MMD) where the kernel is fixed to be Gaussian. Given two distributions $\mathbb{P}, \mathbb{Q}$ the MMD loss is given by 
\begin{equation}
    \begin{aligned}
    M_{k}(\mathbb{P}, \mathbb{Q})= \left\|\mu_{\mathbb{P}}-\mu_{\mathbb{Q}}\right\|_{\mathcal{H}}^{2}
    = \mathbb{E}_{\mathrm{P}}\left[k\left(x, x^{\prime}\right)\right]-2 \mathbb{E}_{\mathrm{P}, \mathbb{Q}}[k(x, y)]+\mathbb{E}_{\mathbb{Q}}\left[k\left(y, y^{\prime}\right)\right]
    \end{aligned}
\end{equation}
The following theorem guarantees convergence of the above loss.
\begin{theorem}
Given a kernel $k$, if $k$ is a characteristic kernel, then $M_{k}(\mathbb{P}, \mathbb{Q}) = 0$ iff $\mathbb{P} = \mathbb{Q}$
\end{theorem}
The above theorem may not hold true when using sample estimator because of the variance of the sample estimator.
The authors propose to overcome the issues within a GMMN network by introducing adversarial kernel learning.They show WGAN is a special case of MMD under certain conditions.
\begin{itemize}
    
\item They propose the following objective they prove to be differentiable,
$$
\min _{\theta} \max _{\phi} M_{k \circ D_{\phi}}\left(p_d, G_{\theta}\right)
$$
The authors use Gaussian kernel over $D_{\phi}$ which are injective functions and claim the resulting kernel is characteristic hence the above theorem hold true.
$$
\tilde{k}\left(x, \hat{x}\right)=\exp \left(-\left\|D_{\phi}(x)-D_{\phi}(\hat{x})\right\|^{2}\right)
$$
\item The $D$ has to be a injective function which is realized using auto-encoder architecture, its are $
\phi=\left\{\phi_{e}, \phi_{d}\right\}
$ which consists of the encoder and decoder parameters 
\begin{equation}
    \begin{aligned}
    \min _{\theta} \max _{\phi} M_{D_{\phi_e}}  \left(p_d, G_{\theta}(p_z(z)\right))
    -\lambda \mathbb{E}_{y \sim p_d \cup p_g}\left\|y-D_{\phi_{d}}\left(D_{\phi_{e}}(y)\right)\right\|^{2}
    \end{aligned}
\end{equation}

\item Comparison with WGAN: The $D_{\phi}$ above uses Gaussian kernel, replacing it with linear and restricting the output of $D_{\phi}$ to have 1 dimension would reduce the above objective to WGAN
\item The MMD distance is more stable as it correlates well with the quality of samples generated as experimentally shown.
\end{itemize}

\subsubsection{Ensemble Method} \label{subsubsec:arch_em} The second most significant one is having a mixture of generators and one discriminator or the other way round which we refer to as \emph{ensemble method}. 
\subsection*{Multi-Agent Diverse GANs \citep{madgan}} The authors propose to resolve the issue of mode collpase, by having mutliple generators and enforcing each to learn a different mode calling it MAD-GAN. We look into proposed architecture and loss function,
\begin{itemize}
    \item k generators and 1 discrimnator
    \item the generators may share parameters for initial layers depending on the dataset
    \item To enforce each generator to learn a different mode, the discriminator not just minimizes the loss but it also identifies which generator has generated the sample.
    \item the output of the discriminator is $k+1$ probability values which includes the k discriminators and belong to the real data-set.
    \item Given that $\delta \in \{ 0, 1\}^{k+1}$ for $j \in \{1, \ldots, k\}, \ \delta(j)=1$ if sample belongs to $j^{th}$ generator otherwise $\delta(k+1)=1$. The objective for discriminator:
    $$ \underset{\theta_d}{max} \ \mathbb{E}_{x \sim p} H(\delta, D(x;\theta_d)) $$
    \item The objective for the generator:
    \begin{equation*}
        \begin{aligned}
        \underset{\theta_g}{min}\ \mathbb{E}_{x \in p_d} \log D_{k+1}(x; \theta_d) 
         + \mathbb{E}_{z\in p_z} \log(1- D_{k+1}(G_i(z;\theta_g^i);\theta_d))
        \end{aligned}{}
    \end{equation*}{}
    The generators are updated in parallel
\end{itemize}
The authors do not provide on the number of generators required w.r.t. the complexity of the data leaving $k$ as an hyper-parameter to be tuned experimentally.
\subsection*{AdaGAN: Boosting Generative Models \citep{adagan}} Propose a GAN based on boosting algorithm which guarantees convergence in finite steps if each step is optimal else exponential convergence. It also addresses reduces mode collapse. The proposed algorithm trains a weak generative model at every iteration such that the samples are re-weighted, giving more weightage to hard (from missed modes) samples. The major components and theorems are discussed below,
\begin{itemize}
    \item Multiple weak generators as a mixture is the overall model for generating images
    \item The generators are added in a sequential manner such that at every step the added model covers the modes which were missed by models trained thus far.
    \item Minimizing $f$-divergence over mixture models: 
    Given $Y_1, \ldots, Y_n \sim Q$ and $X_1, \ldots, X_n \sim P$ where $Y_i = G(Z_i)$ we can find the optimal $Q$ such that,   
    $$
\min _{Q \in \mathcal{G}} D_{f}(Q \| P)
$$
Given $P_{g} :=P_{m o d e l}^{t}$ and new model $Q$ at $t+1$ 
the new model is defined as follows,
$$
P_{\text { model }}^{t+1} :=\sum_{i=1}^{t}(1-\beta) \alpha_{i} P_{i}+\beta Q
$$
Hence they optimize w.r.t. $Q$ and $\beta \in [0,1]$ given by the following objective,
$$
\min _{Q, \beta} \ D_{f}\left((1-\beta) P_{g}+\beta Q \| P_{d}\right)
$$
Finding the optimal $Q$ at every step is not required but they must ensure the following for $c < 1$
$$D_{f}\left((1-\beta) P_{g}+\beta Q \| P_{d}\right) \leq c. D_{f}\left(P_g \| P_{d}\right)$$
As the training proceeds, the new model would have lesser information to add hence the value of $\beta$ should reduce. But with decreasing $\beta$ tuning $Q$ becomes harder at every step, given the samples from mixture distribution would rarely belong to $Q$. Hence the authors minimize the upper bound.
\item  Minimizing upper bound of $f$-divergence over mixture models:
The upper bound is given as follows for some reference distribution $R$ such that $\beta dR \leq dP_d$
\begin{equation*}
    \begin{aligned}
    D_{f}\left((1-\beta) P_{g}+\beta Q \| P_{d}\right) \leq \beta D(Q \| R) 
    + (1-\beta) D_{f}\left(P_{g} \| \frac{P_{d}-\beta R}{1-\beta}\right)
    \end{aligned}{}
\end{equation*}{}

\item Solution for the minimizer of the upper bound is given by the following Theorems.
\begin{theorem}
For any $f$-divergence $D_f$ with $f\in \mathcal{F}$ and $f$ differentiale, any fixed distributions $P_g$ and $P_d$, and any $\beta \in [0, 1]$, the solution to the following minimization problem,
$$ \underset{Q \in \mathbb{P}}{min}D_{f}\left((1-\beta) P_{g}+\beta Q \| P_{d}\right) $$ where $\mathbb{P}$ is a class of all probability distributions is,
\begin{equation}
\label{eq:min1}
dQ^* = \frac{1}{\beta} (\lambda^* dP_d(x) - (1-\beta)dP_g(x))_{+}
\end{equation}
for some unique $\lambda^*$ satisfying $\int dQ^*_{\beta} = 1$. Furthermore, $\beta \leq \lambda^*\leq min(1, \beta/\delta)$, where $\delta := P_d(dP_g = 0)$. Also $\lambda^* = 1$ if and only if $P_d((1 - \beta)dP_g > dP_d) = 0$
\end{theorem}{}
\begin{theorem}
Given two distributions $P_g, P_d$ and some $\beta \in [0, 1]$, assume,
$$P_d(dP_g = 0) < \beta$$ Let $f\in F$. The solution to the minimization of the second term of the upper bound given by
$$\underset{Q:\beta dQ<dP_d}{min} D_f\bigg( P_g \parallel \frac{P_d - \beta Q}{1 - \beta}\bigg)$$
is given by the distribution
\begin{equation}
    \label{eq:min2}
  dQ_{\beta}^{\dagger}(x) = \frac{1}{\beta}(dP_d(x) - \lambda^{\dagger}(1 - \beta)dP_g (x))_+  
\end{equation}{}

for some unique $\lambda^{\dagger}$ satisfying $\int dQ^{\dagger}_{\beta} = 1$.
\end{theorem}{}
The above Theorems are independent of the distribution $f$ used.
\item Final Algorithm:
at each iteration they add $Q$ to the mixture $P_g$ with a weight $\beta$ and the optimal $Q^*$ as given by Equation (\ref{eq:min1}),
$$dQ^* = \frac{dP_d}{\beta} \bigg( \lambda^* - (1 - \beta) \frac{dP_g}{dP_d}\bigg)_+$$
\begin{itemize}
    \item Here thye use adversarial training given a corresponding function $h$
    $$\frac{dP_g}{dP_d}(X) = h(D(X)), \ \ h_{JSD} := \frac{1-D(X)}{D(X)}$$
    Using this they can estimate the weights of training sample when they compute $dQ^*$ as follows, given that $p_i = dP_d(X_i)$ and $d_i = D(X_i)$, $p_i = 1/N$
    $$w_i = \frac{p_i}{\beta} (\lambda^* - (1- \beta)h(d_i))$$
    \item They use an iterative algorithm to determine $\lambda^*$ such that $\sum_i w_i = 1$
    \item Choosing the value of $\beta$ based on heuristics.
\end{itemize}{}

\item Convergence analysis
Necessary and sufficient conditions for the iterative process to converge.
\begin{theorem}
Take any $f \in \mathcal{F}$ such that $f(x)\neq 0$ for $x\neq 1$. Starting from $P_{model}^1 = P_1$ update the model s.t. $P_{model}^{t+!} = (1-\beta)P_{model}^t + \beta Q_{\beta}^*$, where on every step $Q_{\beta}^* is given by Equation (\ref{eq:min1})$ with $P_g = P^T_{model}$. In this case $D_f(P^t_{model} \parallel P_d)$ will reach 0 in finite number of steps iff there exists $M>0$ s.t.
$$P_d((1 - \beta)dP_1 > MdP_d) = 0$$
it takes at most $-\ln max(M,1)/\ln (1-\beta)$ steps.
\end{theorem}{}
Otherwise the convergence is exponential
\end{itemize}{}
\subsection*{Evolutionary GANs \citep{egan}} The authors propose a framework where adversarial training is viewed as mutation operations which evolves a population of generators such that well-performing generators are preserved. Each of the previously proposed metrics like KL divergence has vanishing gradient issue and WGAN can have non-convergent limit cycles near equilibrium. Their method uses different metrics to jointly optimize over the generators. By preserving the best generator at every iteration, it overcomes the weakness of each of the metrics.
There are three steps followed
\begin{itemize}
    \item Given a Generator $G_{\theta}$, several copies are produced $\left\{G_{\theta_{1}}, G_{\theta_{2}}, \cdots\right\}$ which are modified by different mutations (i.e. different generator loss function)
    \item Evaluation: based on quality and diversity
    \begin{itemize}
        \item quality fitness score, the average discriminator value for each of the generator copies
    $$
\mathcal{F}_{\mathrm{q}}=\mathbb{E}_{z}[D(G(z))]
$$\item diversity fitness score: smaller discriminator gradients ensure higher score. Smaller discriminator gradients imply that the generated samples have spread out, to avoid mode collapse.
\begin{equation*}
    \begin{aligned}
    \mathcal{F}_{\mathrm{d}}=- \log \|  \nabla_{D}-\mathbb{E}_{x}[\log D(x)] 
     - \mathbb{E}_{z}[\log (1-D(G(z)))] \|
    \end{aligned}{}
\end{equation*}{}

\item Overall score:
$$
\mathcal{F}=\mathcal{F}_{\mathrm{q}}+\gamma \mathcal{F}_{\mathrm{d}}
$$
    \end{itemize}{}
    \item Selection: The worst performing generator based on the above fitness score is removed and the rest are carried forward to the next iteration.
\end{itemize}{}

\subsection*{Dual Discriminator GANs \citep{d2gan}} 
The authors propose dual discriminator GAN one trained on KL divergence and the other on reverse KL to reduce mode collapse. They cliam KL divergence given by Equaiton \ref{eq:kl} covers all modes but also produces unseen and undesirable samples. Reverse KL \ref{eq:rvkl} tries to fit to one single mode leading to mode collapse. JSD \ref{eq:jsd} minimization is empirically similar to reverse KL and hence results in mode collapse.
In the solution there are two discriminators trained as follows,
\begin{itemize}
    \item $D_1$ gives high score to samples from $P_{data}$ and low to samples from $P_g$
    \item $D_2$ gives high score to samples from $P_g$ and low to samples from $P_{data}$
\end{itemize}{}
Hence the overall objective is given by,
\begin{equation}
    \label{eq:d2gan}
    \begin{aligned} \min _{G} \max _{D_{1}, D_{2}} \mathcal{J}\left(G, D_{1}, D_{2}\right)= &\alpha \times \mathbb{E}_{\mathbf{x} \sim P_{\mathrm{data}}}\left[\log D_{1}(\mathbf{x})\right] 
    + \mathbb{E}_{\mathbf{z} \sim P_{\mathbf{z}}}\left[-D_{1}(G(\mathbf{z}))\right] \\ &+\mathbb{E}_{\mathbf{x} \sim P_{\mathrm{data}}}\left[-D_{2}(\mathbf{x})\right]
    +\beta \times \mathbb{E}_{\mathbf{z} \sim P_{\mathbf{z}}}\left[\log D_{2}(G(\mathbf{z}))\right] \end{aligned}
\end{equation}{}
Role of the hyper-parameters $\alpha, \beta$
\begin{itemize}
    \item stabilize the learning: Having $D_1(G(z))$ and $D_2(x)$ instead of their $\log$ has stronger impact on optimization but causes their output to be very large, hence $\alpha, \beta$ help stabilize that
    \item control the effect of KL and reverse KL
\end{itemize}{}
The convergence results are similar to original GAN

\subsection*{Generalization and Equilibrium in GANs \citep{Arora03}} 
In this paper, the authors provide generalization bounds for the new framework for GAN loss that they define as further discussed in Section \ref{subsubsec:theory_q1}. 
We in this section focus on the other important aspect that the paper addresses is the existence of equilibrium. Although it is unknown as to what equilibrium GAN converges, the authors prove the existence of a particular equilibrium. The main motivation is obtained from the min-max theorem \citep{neumann1928} which shows that if both players are allowed to play mixed strategies, then the game has an equilibrium which is the min-max solution. The paper models mixed strategies by considering a mixture of generators and discriminators.  As an infinite mixture is not possible; it admits an approximate solution with a finite mixture of generators. 

The architecture is called it MIX$+$GAN where they train a mixture of $T$ generators $\{ G_{\phi_i} i \in T \}$ and $T$ discriminators, $\{ D_{\theta_i} i \in T \}$. They maintain weight  $w_{\theta_{i}}=\frac{e^{\alpha_{\theta_{i}}}}{\sum_{k=1}^{T} e^{\alpha_{\theta_{k}}}}$ corresponding to each generator whose log probability predicted by the network is $\alpha_{\theta_i} $. The payoff function is given by,
Given 
$$
F(\theta, \phi)=\underset{x \sim \mathcal{D}_{\text {real}}}{\mathbb{E}}\left[\log\left(D_{\phi}(x)\right)\right]+\underset{x \sim \mathcal{D}_{G}}{\mathbb{E}}\left[\log\left(1-D_{\phi}(x)\right)\right) ]
$$
\begin{equation}
\label{eq:mix-gan}
\begin{array}{l}{\min _{\left\{\theta_{i}\right\},\left\{\alpha_{\theta_{i}}\right\}\left\{\phi_{j}\right\},\left\{\alpha_{\phi_{j}}\right\} i, j \in[T]} F\left(\theta_{i}, \phi_{j}\right)} \\ {=\min _{\left\{\theta_{i}\right\},\left\{\alpha_{\theta_{i}}\right\}} \max _{\left\{\phi_{j}\right\},\left\{\alpha_{v_{j}}\right\}} \sum_{i, j \in[T]} w_{\theta_{i}} w_{\phi_{j}} F\left(\theta_{i}, \phi_{j}\right)}\end{array}
\end{equation}

Similarly there can be an objective function corresponding to the WGAN \ref{eq:wgan}.

\subsubsection{Memory}\label{subsubsec:arch_mem}
Apart from the above methods, there is another paper which changes the architecture to introduce memory within the network 
\subsection*{Memorization precedes generation: Learning unsupervised GANs with Memory Networks \citep{memory}}
The authors raise two-fold issues with vanilla GANs, i)  GANs use unimodal continuous latent distribution to embed multiple classes hence the structural discontinuity between classes is not clear in the generated samples. ii) Discriminators forget about the previously generated samples which incurs instability and divergence in the training. Hence they propose memoryGAN which learns a joint distribution for which the the continuous variable $z$ and class variable $c$ are independent and modeled separately, hence enforcing discontinuity between classes. In the input to the generator $K_i$ represents the class and $z$ represents the variation within the class. It also memorizes the representation of clusters of real or fake samples in the form of key vectors.

Novel Architecture is proposed with the following components:
\begin{itemize}
    \item Discriminative memory network (DMN): Inference network $\mu$ takes in $x$ as input and returns $q$ with $\parallel q \parallel = 1$. The memory module takes $q$ as input and outputs whether $x$ is real or fake.
    \item Memory network: The memory network used above is $\mathcal{M}=(K, v, a, h), \  K \in \mathbb{R}^{N \times M}$ that is $N$ slots of dimension $M$. Each $v \in \{ 0, 1\}^N$ is a memory value vector. $a \in \mathbb{R}^N$ is a vector which tracks the age of each item in each memory slot. $h$ is the slot histogram where $h_i$ is the number of data points belonging to the $i^{th}$ memory slot.
    The posterior distribution over memory slots is given by,
    \begin{equation*}
        \begin{aligned}
        p(c=i | x)= &\frac{p(x | c=i) p(c=i)}{\sum_{j=1}^{N} p(x | c=j) }
        = \frac{\exp \left(\kappa K_{i}^{T} \mu(x)\right) p(c=i)}{\sum_{j=1}^{N} \exp \left(\kappa K_{j}^{T} \mu(x)\right) p(c=j)}
        \end{aligned}{}
    \end{equation*}{}
The categorical prior is given by,
$$
p(c=i)=\frac{h_{i}+\beta}{\sum_{j=1}^{N}\left(h_{j}+\beta\right)}
$$ The memory is updated using incremental EM algorithm
\item Discriminative probability: 

\begin{equation*}
    \begin{aligned}
    p(y=1 | x)= \sum_{i=1}^{N} p(y=1 | c=i, x) p(c=i | x)
    =  \sum_{i=1}^{N} v_{i} p(c=i | x)
    =\mathbb{E}_{i \sim p(c | x)}\left[v_{i}\right]
    \end{aligned}{}
\end{equation*}{}

\item Memory conditioned generator network: It samples a memory index $i$ $P\left(c=i | v_{c}=1\right)=\frac{h_{i} v_{i}}{\sum_{j}^{N} h_{j} v_{j}}$
Then the tuple $ [K_i, z]  $ is passed as the input to the generator. The objective is similar to infoGAN. The authors try to minimize the mutual information between $K_i$ and $G(z, K_i)$ to ensure the structural similarity between the sampled memory information and generated sample. Given $\hat{I}=-E_{x \sim G\left(z, K_{j}\right)}\left[\kappa K_{i}^{T} \mu(x)\right]$
\begin{equation}
\label{eq:memGAN}
\begin{aligned}
\mathcal{L}_{D}=&-E_{x \sim p(x)}[\log D(x)]
-E_{(z, c) \sim p(z, c)}\left[\log \left(1-D\left(G\left(z, K_{i}\right)\right)\right)\right]+\lambda \hat{I} \\ \mathcal{L}_{G}=&E_{(z, c) \sim p(z, c)}\left[\log \left(1-D\left(G\left(z, K_{i}\right)\right)\right)\right]+\lambda \hat{I}
\end{aligned}{}
\end{equation}
\end{itemize}

%%%%%%%%%%%%%%%%%%%%%%%%%%%%%%%%%%%%%%%%%%%%%%%%%%%%%%%%%%%%%%%%%
\subsection{Optimizers}
\label{subsec:optim}
In this section, we discuss papers which change the standard simultaneous gradient descent optimizer used and propose new methods for optimizing in a hope to address Q$_3$.
\subsection*{The Numerics of GANs \citep{geiger06}} In the paper Mescheder et al. \citep{geiger06}, the authors identify the main reason for non-convergence of GANs to local Nash equilibria. Let $\bar{x} = (\bar{\phi}, \bar{\theta})$ be a point of Nash equilibrium given by, $$\bar{\phi} \in \underset{\phi}{argmax}\mbox{  }f(\phi, \bar{\theta}) \quad \text{and} \quad \bar{\theta} \in \underset{\theta}{argmax}\mbox{  }f(\bar{\phi}, \theta) $$. Every differentiable two-player game defines a vector field $v(\phi, \theta) = \begin{bmatrix}
\nabla_{\phi} f(\phi, \theta) \\
\nabla_{\theta} g(\phi, \theta) \\
\end{bmatrix}$. $\bar{x}$ is a stationary point of $v(x)$ and $v'(\bar{x})$ is negative semidefinite iff $\bar{x}$ is a local Nash equilibrium. $v'(\bar{x})$ has eigen values with small real part and big imaginary part which results in slow convergence. This is in particular a problem of simultaneous gradient ascent for two-player games (in contrast to gradient ascent for local optimization), where the Jacobian $v'(x)$ is not symmetric and can therefore have non-real eigenvalues. 
Finding a stationary field is equivalent to solving the equation $v(x) = 0$. They define $L(x) = \frac{1}{2}\parallel v(x)\parallel^2$. Minimizing $L(x)$ directly leads to unstable stationary points, hence they consider a modified vector field $w(x) = v(x) - \gamma \nabla L(x) $ for some $\gamma > 0$. The modified utility functions for the two player game is now, 
$$
\hat{f}(\phi, \theta) = f(\phi, \theta) - \gamma L(\phi, \theta) $$ 
$$and \quad \hat{g}(\phi, \theta) = g(\phi, \theta) - \gamma L(\phi, \theta) $$ The $L(\phi, \theta)$ term encourages agreement between the two players, hence is called \textit{Consensus Optimization.}

\subsection*{Training GANs with Optimism \citep{optimism}}
In this paper the authors address the issue of limit cycling behaviour in WGAN by proposing Optimistic Mirror Descent (OMD).  In GANs, to solve the zero sum game, simultaneous SGD is used which is similar to running no-regret dynamics for each player.  From game theory it is known that this leads to limit oscillatory behaviour. Theoretical results show that no variant of GD can converge to an equilibrium in terms of the last iterate even in convex-concave setting.  It is only the average of the weights of the two players that constitutes an equilibrium. OMD converges to equilibrium in terms of the last iterate for bilinear functions.

Optimistic Mirror Descent: algorithm for zero-sum games which achieves faster convergence rate to equilibrium of $\epsilon = \mathcal{O}(\frac{1}{T})$  for the average of the parameters.

The algorithm uses the last iteration gradient as a predictor for the next iteration's gradient. The update rule is as follows,
$$
\begin{aligned} w_{t+1} &=w_{t}+2 \eta \cdot \nabla_{w, t}-\eta \cdot \nabla_{w, t-1} \\ \theta_{t+1} &=\theta_{t}-2 \eta \cdot \nabla_{\theta, t}+\eta \cdot \nabla_{\theta, t-1} \end{aligned}
$$
The stochastic OMD where the gradients are replaced by the unbiased estimators, estimated over a mini-batch of $B$ samples
$$ 
\begin{aligned} \hat{\nabla}_{w, t} &=\frac{1}{|B|} \sum_{i \in B}\left(\nabla_{w} D_{w_{t}}\left(x_{i}\right)-\nabla_{w} D_{w_{t}}\left(G_{\theta_{t}}\left(z_{i}\right)\right)\right) \\ \hat{\nabla}_{\theta, t} &=-\frac{1}{|B|} \sum_{i \in B} \nabla_{\theta}\left(D_{w_{t}}\left(G_{\theta_{t}}\left(z_{i}\right)\right)\right) \end{aligned}
$$
The above is claimed to be different from the variants of SGD.

\subsection*{GANs Trained by a Two Time-Scale Update Rule Converge to a Local Nash Equilibrium \citep{ttur}}
The authors raise the following concerns regarding C$_2$
\begin{itemize}
    \item The convergence of GAN training has not been proven. Only local Nash equilibrium are found because of gradient descent. 
    \item The authors of \citep{CMU17} prove the stability that is required for local equilibrium but with strong unrealistic assumptions and restricting $D$ to linear cases.
    \item Recent proofs  \citep{Arora03}, \citep{fisher}, \citep{app_conv}, \citep{mmd-gan}  do not consider minibatch training and instead provide convergence proofs for when the samples go to infinity
\end{itemize}{}
They propose the following optimization algorithm and prove its convergence
\begin{itemize}
    \item Propose Two time-scale update rule (TTUR). 
    ($G$ and $D$ have two different learning rates. $D$ converges with $G$ fixed. If $G$ changes slowly with small gradients, $D$ still converges.)
    \item TTUR converges to a stationary local Nash equilibrium under five assumptions \citep{ttur}. Given 
\begin{equation}
\label{eq:ttur}
\begin{aligned}
 &{\theta}_{n+1}={\theta}_{n}+b(n)\left(G\left({\phi}_{n}, {\theta}_{n}\right)+{M}_{n}^{(\theta)}\right), \\ &{\phi}_{n+1}={\phi}_{n}+a(n)\left({D}\left({\phi}_{n}, {\theta}_{n}\right)+{M}_{n}^{(\phi)}\right)
\end{aligned}{}
\end{equation}{}

   The learning rates are $b(n)$ for $D$ and $a(n)$ for $G$
    and ${M}_{n}^{(\theta)}$ and ${M}_{n}^{(\phi)}$ are the difference between true gradients and stochastic gradients estimated from minibatch.
    \begin{theorem}
    If the assumptions are satisfied then the updates given by Equation (\ref{eq:ttur}) converge to $\phi^*, \lambda(\phi^*)$, where $\lambda$ is the local asymptotically stable attractor
    \end{theorem}{}
    \item In practice $G$ is updated slowly such that $D$ converges faster
\end{itemize}{}

The authors also raise the issue of Mode Collapse C$_1$ and propose to use Adam Optimizer \citep{adam} to overcome the issue. They claim the update rule in Adam depends on the average over the past gradients, hence the update does not get pushed into smaller regions hence it seeks flat minima and avoids local minima. The authors also  characterize the hyper-parameters within the Adam which results in convergence

%%%%%%%%%%%%%%%%%%%%%%%%%%%%%%%%%%%%%%%%%%%%%%%%%%%%%%%%%%%%%%%%%%
\subsection{Theoretical Aspects}
\label{subsec:theory}
In this section we discuss the papers which provide rigorous theoretical analysis for analyzing the convergence and generalization properties of GANs. Apart from this, there are other papers which characterize mode collapse and other issues with the previous work. 
Let us first consider a paper which gives an idea how in GANs the theoretical analysis and the practical results do not agree at all times.

\subsection*{Many Paths to Equilibrium: Gans Do Not Need To Decrease Divergence At Every Step \citep{many}}
In this paper, the authors give two major claims and show them empirically. i) GANs training need not be guided by divergence minimization. Even when theoretically the problem of vanishing gradients should occur, the discriminator provides gradients. ii) Introducing Gradient penalty  as discussed in paper \citep{gularajani} is motivated from divergence minimization, but can be used independently.

The authors conduct the experiments using Equation (\ref{eq:gan_loss2}) which does not suffer from vanishing gradient problem, instead of Equation (\ref{eq:gan_loss1}).
To the Equation (\ref{eq:gan_loss2}) they add two different gradient penalty terms proposed in \citep{gularajani} (GAN-GP), \citep{DRAGAN} (DRAGAN-NS). The discriminator loss looks as follows,
\begin{equation*}
    \begin{aligned}
    V(D, G) = -\mathbb{E}_{x\in p_d}[\log(D(x))] - \mathbb{E}_{z \in p_z}[\log (1 - D(G(z)))] 
    +  \lambda \mathbb{E}_{\hat{x}\in p_{\hat{x}}} [(\parallel\nabla_{\hat{x}} D(\hat{x})\parallel_2 - 1)^2] 
    \end{aligned}{}
\end{equation*}{}

Where they use the following
$$
x \sim p_{\text { data }} ; \quad x_{\text { model }} \sim p_{\text { model }} ; \quad x_{\text { noise }} \sim p_{\text { noise }}
$$
$$
 \begin{array}{ll}{\text{DRAGAN}}{\ \tilde{x}=x+x_{\text { noise }}} \\ 
{\text{WGAN-GP} \ }{\tilde{x}=x_{\text { model }}}\end{array}
$$
$$
\begin{array}{c}{\alpha \sim U(0,1)} \\ {\hat{x}=\alpha x+(1-\alpha) \tilde{x}}\end{array}
$$
These are the following results observed from their experiments
\begin{itemize}
    \item Both GAN-GP and DRAGAN-NS stabilize training and improve convergence 
    \item Gradient penalty makes the training less sensitive to hyperparameter tuning.
    \item The non-saturating version of GAN given by Equation (\ref{eq:gan_loss2}) performs well on disjoint manifold as well. No issue of vanishing gradients since the generator loss is able to amplify the small differences in discriminator loss and obtain larger gradients. 
    \item JSD being parameterized by density functions suffers from the problem of vanishing gradients whereas in practice , it is parameterized by the samples from the two distributions.
    \item The above loss works well for over-capacity generator and lower dimension input.
\end{itemize}{}

\bigskip

For the rest of subsection, we divide the work based on which of the questions they address as discussed in Section \ref{sec:challenges}
\subsubsection{Addressing \emph{Q$_1$}} \label{subsubsec:theory_q1} We discuss the papers which relate the capacity of the network with the major challenges.
\subsection*{Generalization and Equilibrium in GANs \citep{Arora03}}
In the paper Arora et al. \citep{Arora03}, question the generalization of GAN objective as well as the existence of pure equilibrium in the two-player game.  Generalization in GANs as defined by the authors means that the population distance between the true and the generated distribution is close to the empirical distance between the empirical distribution. $$|d(\mathcal{D}_{real},\mathcal{D}_{G})-d(\mathcal{\hat{D}}_{real}, \mathcal{\hat{D}}_{G})| \leq \epsilon$$ where $\mathcal{\hat{D}}_{real}$ is the empirical version of $\mathcal{D}_{real}$ with polynomial number of samples. They prove that Jenson Shanon Divergence and Wasserstein distance do not generalize with a polynomial number of examples. Further analysis show that GANs actually minimize a surrogate distance called the Neural Network distance, \begin{definition}Let $\mathcal{F}$ be a class of functions from $\mathbb{R}^d$ to $[0, 1]$ such that if $f \in \mathcal{F}, 1-f \in \mathcal{F}.$ Let $\phi$ be a concave measuring function. Then the $\mathcal{F}$-divergence with respect to $\phi$ between two distributions $\mu$ and $\nu$ supported on $\mathcal{R}^d$ is defined as $$d_{\mathcal{F},\phi}(\mu, \nu) = \underset{D\in\mathcal{F}}{sup}\mbox{ }\underset{x\sim\mu}{\mathbb{E}}[\phi(D(x))] + \underset{x\sim\nu}{\mathbb{E}}[\phi(1 - D(x))] -  2\phi(1/2) $$    
\end{definition}
The major theorem stated in the paper claims that, since there are not infinitely many discriminators, given enough samples the expectation over the empirical distribution converges to the expectation over the true distribution for all discriminators. Although this analysis guarantees generalization, the assumption of finite discriminators results in lack of diversity in the generated distribution. For JS and Wasserstein distance, when the distance between two distributions $\mu$, $\nu$ is small would imply that the distributions are close. However $d_{NN}(\mu, \nu)$ can be small even if the distributions are not close. A neural network with p parameters cannot distinguish between a distribution $\mu$ and distribution with support $\tilde{O}(p/\epsilon^2)$. Such a limited capacity network cannot learn the distribution although it has access to a lot of samples from the distribution $\mu$.

The other important aspect that the paper addresses is the existence of equilibrium. Although it is unknown as to what equilibrium GAN converges, the authors prove the existence of a particular equilibrium. The main motivation is obtained from the min-max theorem \citep{neumann1928} which shows that if both players are allowed to play mixed strategies, then the game has an equilibrium which is the min-max solution.

\subsection*{Do GANs Learn the Distribution? Some Theory and Empirics \citep{doarora}}
In this paper the authors aim to quantify mode collapse. They claim there is no clear quantitative metric for sample diversity in the generated samples. With a discriminator size of $p$, the training objective could be $\epsilon$ close to the optimal even though the output distribution is supported on only $O\left(p \log p / \epsilon^{2}\right)$ images. \\
Hence they aim to validate mode collapse in different gans by conducting a birthday paradox test for gans. This gives the support size of the learned distribution. Using the metric they also provide limitation of encoder-decoder frameworks like BiGAN \citep{bigan}, ALI \citep{ali}. They propose the following \emph{Birthday Paradox Test}
 
\begin{itemize}
    \item Given a distribution of $N$, the $\sqrt{N}$ samples would have duplicates.
    \item In the proposed method, one searches for duplicates, if there are $s$ duplicates with high probability, the the distribution has a support of $s^2$
    \item The test is likely to fail, when few samples have high probability and rest have low, although the support is large, such non uniformity is a faliure of GAN too.
    \item In GANs the distribution is infinite, hence the $s$ where the probability to find duplicates should also be very large. But there is more than 50\% probability that there is a duplicate in 800 samples for DCGAN and MIX+DCGAN and 1200 samples for ALI. This shows the distribution is 6.4 lakhs and 1 million respectively.
\end{itemize}{}

They discuss the following limitations of encoder-decoder architecture: Given the BiGan objective function,
    $$
\min _{G, E} \max _{D}|\underset{x \sim \hat{\mu}}{\mathbb{E}} \phi(D(x, E(x)))-\underset{z \sim \hat{\nu}}{\mathbb{E}} \phi(D(G(z), z))|
$$
The following theorem concludes that given that the encoder has very small complexity and the support of the generated distribution is small still the objective function of BiGAN becomes very small.
\begin{theorem}
There exists a generator $G$ of support $\frac{p \Delta^{2} \log ^{2}\left(p \Delta L L_{\phi} / \epsilon\right)}{\epsilon^{2}}$ and and encoder $E$ with at most $\Tilde{d}$ non-zero weights, s.t. for all discriminators $D$ that are L-Lipschitz and have a capactity less than $p$, $$
|\underset{x \sim \mu}{\mathbb{E}} \phi(D(x, E(x)))-\underset{z \sim \nu}{\mathbb{E}} \phi(D(G(z), z))| \leq \epsilon
$$
\end{theorem}

\subsection*{Approximation and Convergence Properties of Generative Adversarial Learning \citep{app_conv}}
The authors find a relation between the parameters in the discriminator and the convergence of $p_g$ to $p_d$.
They define a notion of adversarial divergence as follows,
\begin{definition}[Adversarial Divergence]
\label{def:adv_div}
Let $X$ be a topological space, $\mathcal{F} \subseteq C_b(X^2), \mathcal{F} \neq \Phi$. ($C_b(X^2)$ is the set of bounded continuous functions on $X^2$) An adversarial divergence $\tau$ over $X$ is given by,
$$
\begin{aligned} \mathcal{P}(X) \times \mathcal{P}(X) & \longrightarrow \mathbb{R} \cup\{+\infty\} \\(p_d, p_g) & \longmapsto \tau(p_d \| p_g)=\sup _{f \in \mathcal{F}} \mathbb{E}_{p_d \otimes p_g}[f] \end{aligned}
$$

They show that $\tau$ defined above encompasses general class of objective functions which include GAN \ref{eq:gan_loss1}, f-GAN, MMD-GAN, WGAN, WGAN-GP, entropic regularized optimal transport problems.

They show that using a restricted class of discriminators, the adversarial divergence is equivalent to matching generalized moments. Hence they analyze the existence and properties of unique $p_g$ which minimizes the objective given a discriminator with restricted capacity.

\end{definition}{}

\subsection*{A Convex Duality Framework for GANs \citep{convDual}}
As discussed in Section \ref{subsubsec:loss_vg}, the authors in this paper propose to study the divergence minimization perspective in a setting where the discriminator has restricted capacity by developing a convex duality framework.
As shown in the original paper \cite{ganGoodfellow}, GANs are trained using a minmax objective which reduces to minimizing the JSD divergence if we assume the models to have infinite capacity and hence mimic all possible distributions (Section \ref{sec:set_up}). In practice $D$ is restricted to a smaller class of distributions denoted by $\mathcal{F}$ as shown in \citep{Arora03}.  The authors show that by restricting the discriminator to a class of convex functions the convex dual objective searches for the generative model which is closest to the distribution $Q$, such that $Q$ shares the same moments as $p_d$. Refer to Section \ref{subsubsec:loss_vg} for further details.

\subsubsection{Addressing \emph{Q$_2$}} \label{subsubsec:theory_q2} We discuss the papers which define new loss and prove its convergence in non-parametric space and also give generalization bounds.
\subsection*{Generalization and Equilibrium in GANs \citep{Arora03}} This paper, as discussed in Section \ref{subsubsec:theory_q1}, uses the fact that there always exists a mixed strategy Nash equilibrium. To simulate mixed strategies they consider a mixture of generators.  As an infinite mixture is not possible; it admits an approximate solution with a finite mixture of generators. Thus they show the existence of $\epsilon-$approximate equilibrium. 

\subsection*{Approximation and Convergence Properties of Generative Adversarial Learning \citep{app_conv}}
Besides what was discussed in Section \ref{subsubsec:theory_q1} the authors also raise the following issue of convergence and generalization. 
\begin{itemize}
    \item How well can GANs approximate the target distribution in the presence of large number of samples and perfect optimization?
    \item Does GAN set-up always converge under the standard notion of distributional convergence?
\end{itemize}{}
The authors show convergence of adversarial divergence defined in \ref{def:adv_div} implies a standard notion of topological convergence.

\subsection*{Loss Sensitive GANS on Lipschitz Densities \citep{ls-gan}}
The authors claim the non-parametric assumption that the model has infinite modeling capacity is too strong. Hence, they propose a new loss of which convergence results do not require the assumption. They also provide generalization bounds.

As discussed in Section \ref{subsubsec:loss_vg}, the authors introduce a new margin based loss which quantifies the quality of generated samples.
$$L_{\phi}(x) = L_{\theta}(G_{\phi}(z)) - \Delta(x, G_{\phi}(x)) $$ where $\Delta(x, G_{\phi}(x))$ is the margin between the losses is the difference between $x$ and $G_{\phi}(z)$. \newline The loss is data-dependent and vanishes for better samples generated. ($l_{p}$-distance). Given $\zeta_{x,z}$ is a slack variable, 
$$\underset{\theta }{min}\mbox{ }S(\theta) \triangleq \underset{x\sim p_{data}(x)}{\mathbb{E}} L_{\theta}(x) + \lambda \mathbb{E}_{\substack{{x\sim p_{data}(x)}\\{z\sim p_{z}(z)}}} \zeta_{x,z}$$
$$s.t., L_{\theta}(x) -  \zeta_{x,z} \leq L_{\theta}(G_{\phi}(z)) - \Delta(x, G_{\phi}(z)), \zeta_{x,z} \geq 0$$
$$\underset{\phi}{min} \underset{z\sim P_{z}(z)}{\mathbb{E}}L_{\theta^*}(G_{\phi}(z))$$
Convergence of Loss with underlying Lipschitz densities (no need of infinite capacity)
Given that the loss function $L, p_{G^*}, p_{data}$ are Lipschitz continuous. Then as $\lambda \rightarrow \infty, P_{G^*} \rightarrow P_{data}$ as $\int_{x} |P_{data} - P_{G^*}(x)|dx \leq \frac{2}{\lambda}$ 

\noindent Generalization ability
\begin{assumption}
\begin{itemize}
\item The loss function $L_{\theta}(x)$ is $\kappa_{L}$-Lipschitz in its parameter $\theta$, i.e., $|L_{\theta}(x) - L_{\theta'}(x)| \leq \kappa_{L} \parallel \theta - \theta' \parallel$ for any $x$;
\item $L_{\theta}(x)$ is $\kappa_{L}$-Lipschitz in $x$, i.e., $|L_{\theta}(x) - L_{\theta}(x')| \leq \kappa_{L} \parallel x - x' \parallel$
\item The distance between two samples is bounded, i.e., $|\Delta(x, x')| \leq B_{\Delta}$
\end{itemize}
\end{assumption}
\begin{theorem}
Under the Assumption 1 with probability $1 - \eta$, we have $|S_{m} - S| \leq \epsilon$ when the number of samples $$m \geq \frac{CNB^2_{\Delta}(\kappa + 1)^2 \log(\kappa_{L}N/\eta \epsilon) }{\epsilon^2}$$ where $C$ is a sufficiently large constant, and $N$ is the number of parameters in the loss function. 
\end{theorem}
Similarly, generalizability can be derived for $T(\theta, \phi)$ with assumptions on $G_{\phi}$. Sample complexity is dependent on parameter size and Lipschitz constants.

\subsection*{Coloumb GANs: Provably Optimal Nash Equilibria via Potential Fields \citep{coulomb}} As discussed in Section \ref{subsubsec:loss_nc}, the authors propose a new loss function. They show that there exists a unique Nash equlibrium in GAN set-up with the proposed loss function.

\subsection*{Dual Discriminator GANs \citep{d2gan}} As discussed in Section \ref{subsubsec:arch_em}, the authors have proposed a D2GAN and provide the following theorem for it's convergence,
\begin{theorem}
Given $D_1^*, D_2^*$, at the Nash equilibrium point $(G^*, D_1^*, D_2^*)$ for minmax optimization problem of their model named D2GAN, we have the following form for each component
$$\begin{aligned} \mathcal{J}\left(G^{\star}, D_{1}^{\star}, D_{2}^{\star}\right) &=\alpha(\log \alpha-1)+\beta(\log \beta-1) \\ D_{1}^{*}(x)
&=\alpha \text { and } D_{2}^{*}(x)=\beta, \forall x \text { at } p_{G^{*}}=p_{\text {data}} \end{aligned}$$
\end{theorem}

\subsubsection{Addressing \emph{Q$_3$}} \label{subsubsec:theory_q3} We discuss the papers which actually talk about the optimization algorithm converging to the equilibrium. Essentially they address the instability in the training algorithms.

\subsection*{$f$-GAN: Training Generative Neural Samplers using Variational Divergence Minimization \citep{fgan}}
Couple of years after the original GAN paper  Nowozin et al. generalized the idea of generative models which use probabilistic feed forward neural networks \citep{fgan}. They call it generative neural samplers. They even generalized the notion of statistical divergences which measure the distances between two distributions. Given two distributions $P$ and $Q$, they define $f$-divergence, $$D_{f}(P \parallel Q) = \int_{\mathcal{X}}q(x)f \bigg( \frac{p(x)}{q(x)} \bigg) \emph{dx} $$ where the function $f : \mathbb{R}_{+} \rightarrow \mathbb{R}$ is a lower semi continuous function satisfying $f(1) = 0$. They also mention the variational lower bound of the $f$-divergences.  

\noindent\textit{Fenchel Conjugate}$f^*$ defined as follows is used in the loss function.\\
 $$f^*(t) = \underset{u\in dom_{f}}{sup} \left \{ut - f(u) \right \}$$ 
Variational Divergence Minimization is the new method they suggest for estimating the parameters of the model $G_{\phi}$. Given that $D : \mathcal{X} \rightarrow \mathbb{R}$ is the variational function parameterized by $\theta$, the $f$-GAN objective is given by,
\begin{equation}\label{eq:5}
V(\phi, \theta) = \mathbb{E}_{x \sim p_d}[D_{\theta}(x)] - \mathbb{E}_{x \sim Q_{\phi}}[f^*(D_{\theta}(x))] 
\end{equation}

It is shown that GAN objective is a particular instance of the above loss function. They propose a single-step gradient descent algorithm and prove its convergence to the saddle point if there is a neighborhood around it in which $V$ is strongly convex in $\phi$ and strongly concave in $\theta$ .

\subsection*{On Convergence and Stability of GANs \citep{DRAGAN}}
The authors raise the following issues,
\begin{itemize}
\item In non-convex settings alternate gradient update is unstable and results in mode collapse by converging to a potentially bad local equilibrium. The issues of cycling not addressed
\item The divergences are supposed to be maxed out for disjoint distributions, yet the vanilla GAN is able to learn the swiss roll distribution which it should not by divergence minimization hypothesis. Hence such a theory is not suitable to discuss convergence or address instability.
\item Coupled smoothing in LS-GAN and WGAN-GP both regularize the discriminator's gradients in the domain space. LS-GAN : $D_{\theta}(x) - D_{\theta}(G_{\phi}(z)) \approx \parallel x, G_{\phi}(z) \parallel$, WGAN : $\parallel\nabla_{x}D_{\theta}(\hat{x}) \parallel$ where $\hat{x} = \epsilon x +(1-\epsilon )G_{\phi}(z) $  
\item WGAN-GP's penalty doesn't follow from KR duality as claimed.By Lemma 1 of \citep{gularajani}, the optimal discriminator D∗ will have norm-1 gradients (almost
everywhere) only between those $x$ and $ G_{\phi}(z)$ pairs which are sampled from the optimal coupling $\pi^*$ not for arbitrary samples of real and fake data.
\end{itemize}

Hence they propose an alternative perspective of regret minimization for convex-concave case,
If both players update their parameters using no-regret algorithms then it is easy to show that their averaged iterates will converge to an equilibrium pair.
\begin{definition}(No-regret algorithm). Given a sequence of convex loss functions $L_{1}, L_{2}, \ldots : K \rightarrow \mathbb{R},$ an algorithm that selects a sequence of $k_{t}$'s, each of which may depend only on previously observed $L_{1}, \ldots L_{t-1}$, is said to have no regret if $\frac{R(T)}{T} = o(1)$ where $$R(T) := \sum_{t=1}^T L_{t}(k_{t}) - min_{k \in K}\sum_{t=1}^T L_{t}(k) $$
\end{definition}
Guaranteed convergence under no-regret condition : $\bar{\phi_{T}} := \frac{1}{T}\sum_{t=1}^T \phi_{t},\mbox{ }\bar{\theta_{T}} := \frac{1}{T}\sum_{t=1}^T \theta_{t}$,  $V^*$ is the equilibrium and $R_{1}(T), R_{2}(T)$ : Regrets. Using standard arguments it can be shown, 

\begin{equation*}
    \begin{aligned}
    V^* - \frac{R_{2}(T)}{T} & \leq max_{\theta \in \Theta} J(\bar{\phi},\theta) - \frac{R_{2}(T)}{T} 
     \leq min_{\phi \in \Phi } J(\phi, \bar{\theta}) + \frac{R_{1}(T)}{T} 
      \leq V^* + \frac{R_{1}(T)}{T}
    \end{aligned}{}
\end{equation*}{}
 Under no regret $\bar{\phi_{T}}, \bar{\theta_{T}}$ are almost optimal. For the non-convex case they use the result that under the notion of local regret, if both the players used a smoothed variant of OGD to minimize this quantity then the non-convex game converges to some form of $\epsilon-$approximate local equilibrium.

\subsection*{GANs Trained by a Two Time-Scale Update Rule Converge to a Local Nash Equilibrium \citep{ttur}}
As discussed in Section \ref{subsec:optim} in this paper, the authors propose Propose  Two time-scale update rule (TTUR). ($G$ and $D$ have two different learning rates. $D$ converges with $G$ fixed. If $G$ changes slowly with small gradients, $D$ still converges.)
 TTUR converges to a stationary local Nash equilibrium under five assumptions stated in the paper. Given the Equation (\ref{eq:ttur}) they prove the following convergence theorem.
\begin{theorem}
If the assumptions are satisfied then the updates given by Equation (\ref{eq:ttur}) converge to $\phi^*, \lambda(\phi^*)$, where $\lambda$ is the local asymptotically stable attractor
\end{theorem}{}

\subsection*{Adagan: Boosting Generative Models \citep{adagan}} As discussed in Section \ref{subsubsec:arch_em} the authors prove convergence analysis for their method proposed. 
 
\subsection*{The Numerics of GANS \citep{geiger06}}
As discussed in Section \ref{subsec:optim} the authors identify when simultaneous gradient descent does not converge to local Nash equilibrium based on the Jacobian of the gradients. The introduce a penalty term which results in consensus optimization. The authors show that this converges to local Nash equlibrium.

\subsection*{Gradient Descent GAN Optimization is Locally Stable  \citep{CMU17}}
This paper is a follow-up work on the above paper \citep{geiger06}. 
They show that Equation (\ref{eq:gan_loss1}) although not convex-concave game, the optimization is locally asymptotically stable under proper conditions. WGAN has non-convergent limit cycles. In order to prove the above, they use ODE method for analyzing convergence properties of dynamical system. 

They also suggest the addition of a regularization term on the norm of the discriminant gradient. Besides they establish that under suitable conditions GAN optimization is locally exponentially stable.  WGAN although can perennially cycle around an equilibrium point without converging. The regularization that they propose enhances the local stability of the optimization procedure, for any general gan framework.
They suggest the following update of Generator,
$$\theta_{G} := \theta_{G} - \alpha \nabla_{\theta_{G}} (V(D_{\theta_{D}}, G_{\theta_{G}})) + \eta \parallel \nabla_{\theta_{D}} V(D_{\theta_{D}}, G_{\theta_{G}}) \parallel $$

\subsection*{Which Training Methods for GANs do Actually Converge? \citep{mescheder18icml}}
The authors introduce Dirac-GAN a simple example as follows,

\begin{definition}[Dirac-GAN]
It consists of a (univariate) generator distribution $p_\phi = \delta_\phi$ and a linear discriminator $D_{\theta} (x)  = \theta x$

Based on the above counterexample, they prove their further claims.

\end{definition}{}

1) Non-convergence of unregularized gans:
They analyze the unregularized objective given by,
$$
L(\theta, \phi)=f(\phi \theta)+f(0)
$$
As given in the Dirac-GAN, usually distributions lie on low-dimensional manifolds and they claim that for Dirac-GAN, alternating gradient descent on the above objective oscillates in stable cycles around the equilibrium. They also show that the non-convergence is backed by analyses given by \citep{geiger06,CMU17}.These instabilities arise due to the fact that, when the $p_g$ is far from $p_d$, the discriminator gradients push it closed while the discriminator itself is more certain. When $p_g = p_d$, the discriminator is most certain and pushes the distributions apart. Hence, there needs to be zero gradients for the discriminator in the orthogonal direction of the data-manifold.

2)Introducing Gradient Penalty as in \citep{gularajani} WGAN-GP and performing simultaneous gradient descent also does not converge on Dirac-GAN

3) Introducing instance noise i.e, adding Gaussian noise to input data led to convergence. Motivated from this, noise induced regularization was introduced in \citep{fgan_2} discussed in Subsection \ref{subsubsec:loss_reg}. The authors show, performing simultaneous gradient descent for this penalty converges for Dirac-GAN.

4)The authors suggest the following simplified gradient
penalty which is a simplified version as proposed in \citep{fgan_2}. This would ensure non-zero loss if there is non-zero gradients w.r.t. the discriminator in the orthogonal direction to the data manifold 

\begin{equation}
    R_{1}(\theta) :=\frac{\gamma}{2} \mathrm{E}_{p_d (x)}\left[\left\|\nabla D_{\theta}(x)\right\|^{2}\right]
\end{equation}{}
 To penalize the discriminator on the generator distribution,
 \begin{equation}
R_{2}(\theta, \phi) :=\frac{\gamma}{2} \mathrm{E}_{p_g(x)}\left[\left\|\nabla D_{\theta}(x)\right\|^{2}\right]
\end{equation}  
Then they prove that with small learning rates, applying simultaneous gradient descent on the GAN objective with above regularizer is locally convergent.

Given the extensive summary of each paper, we present a tabular summary highlighting the challenges addressed in each paper in the next section.

%%%%%%%%%%%%%%%%%%%%%%%%%%%%%%%%%%%%%%%%%%%%%%%%%%%%%%%%%%%%%%%%%%%%%%%%%%%%%%%%%%%%%%%%%%%%%%%%%%%%%%%%%%
\section{Comparison Based Summary}
\label{sec:comparison}
In this section we aim to present the comparison among all the papers discussed above. We provide the comparison in a tabular manner and also provide pictorial results from various papers.

\subsection{Visual Comparison}
\label{sec:vis_summary}
We provide some sample results for few of the papers discussed above. The image results we provide are for the following five data-sets, i) MNIST ii) CELEB A iii) CIFAR iv) LSUN v) IMAGENET.
Given the limitations on file size, we provide the link with the images.
\url{https://drive.google.com/file/d/1SZjzJfyN6wTrjwIt2ge6xjP6cLb-npUn/view} \footnote{Kindly enlarge on screen for better view}. MNIST and CIFAR has images of very low resolution hence accessing the quality of images generated is tough. It is evident from the images that few of the best results have been reported on Celeb A and LSUN dataset. Celeb A is specific to face and LSUN consists of bedroom images. On datasets like Imagenet, there are large variety of classes and the resolution is pretty high hence the model is not able to perform very well.  
\subsection{Tabular Summary}
\label{ssec:tab_summary}

In this section, we provide a tabular summary (Table \ref{tab:summary}. This is to give a bird's eye view over all the papers discussed so far. We aim to highlight each of its contributions and proposed solutions. The first column is the paper name, the second enlists the concerns raised in the paper. The third column specifies which of the challenges out of C$_1$ and C$_2$ does the paper address. The fourth column points out the novel approach followed and the final column specifies the category of the solution. 

From the Table \ref{tab:glance}, we can observe that while most of the papers aim to resolve the issue of non-convergence there a few which explicitly attempt to resolve mode collapse \footnote{In a way mode collapse can also be seen as an outcome of non-convergence. Yet we find most of the papers deal with these aspects separately and hence we find it convenient to categorize them into two different issues. }. We also find that researchers are keenly interested in combining VAEs and GANs by changing the architectures to build a model that includes the best of both models (S$_2$(i)). The increased popularity of WGAN has led to many papers which explore other distance metrics in the loss (S$_1$(i)(ii)). WGAN-GP has  popularized the notion of gradient penalty which significantly improves the performance (S$_1$(iii)(iv)). There has been many papers catching upto this notion and have introduced gradient penalties of different forms as the regularizers. While there has been few papers using ensemble methods S$_2$(ii) there is only one which uses the notion of memory in the network architecture S$_2$(iii). We find that there has been significantly less contribution in S$_3$ which is proposing new optimizers for better convergence and S$_4$(i) which is characterizing the quantitative relation between the capacity of the network and GAN related issues.

\begin{table}[!htb]
  \centering
\begin{adjustbox}{max width =0.95\textwidth}
  {\def\arraystretch{1.4}

\begin{tabular}{||l|l|c|l|c||}
\hline
\multicolumn{1}{||l|}{\emph{Paper}} & \multicolumn{1}{l|}{\emph{Concerns Raised}} & \multicolumn{1}{l|}{\emph{Challenge}} & \multicolumn{1}{l|}{\emph{Novel Approach}} & \emph{Solution} \\ \hline
\hline
 & {\color[HTML]{680100} Overtraining of the discriminator} &  & {\color[HTML]{680100} Feature Matching} &  \\
 & {\color[HTML]{CE6301} Mode collapse of Generator} &  & {\color[HTML]{CE6301} Mini-batch Discrimination} &  \\
 & {\color[HTML]{010066} Gradient descent may not converge} &  & {\color[HTML]{010066} Historical Averaging (Fictitious play)} &  \\
 & {\color[HTML]{34696D} Vulnerable to adversarial examples.} &  & {\color[HTML]{34696D} Label-smoothing} &  \\
\multirow{-5}{*}{\begin{tabular}[c]{@{}l@{}}Improved techniques for \\ training GANs\\ \citep{goodfellow16}\end{tabular}} & GAN outputs depend on the inputs & \multirow{-5}{*}{\begin{tabular}[c]{@{}c@{}}C$_1$\\ C$_2$ \end{tabular}} & Virtual Batch normalization & \multirow{-5}{*}{\begin{tabular}[c]{@{}c@{}}S$_1$(i)\\ S$_1$(v)\end{tabular}} \\ \hline \hline
 & {\color[HTML]{680100} \begin{tabular}[c]{@{}l@{}}Perfect Discriminator resulting in\\ zero grads when distributions are in\\ low dimensional manifolds (vanishing grad)\end{tabular}} &  & {\color[HTML]{680100} \begin{tabular}[c]{@{}l@{}}Softer Metrics-Adding Gaussian\\ for Training GANs [1]\\ Noise (Contrastive Divergence\end{tabular}} &  \\
\multirow{-4}{*}{\begin{tabular}[c]{@{}l@{}}Towards Principled Methods \\ for training GANs\\ \citep{arjovsky01}\end{tabular}} & {\color[HTML]{CE6301} \begin{tabular}[c]{@{}l@{}}The Equation \ref{eq:gan_loss2} alternative \\ causes unstable updates\end{tabular}} & \multirow{-3}{*}{C$_2$} & {\color[HTML]{CE6301} \begin{tabular}[c]{@{}l@{}}No need for the update with softer \\ metric\end{tabular}} & \multirow{-3}{*}{S$_1$} \\ \hline \hline
 & {\color[HTML]{010066} Vanishing Gradient} &  & {\color[HTML]{010066} Propose EM distance} &  \\
 & {\color[HTML]{34696D} } &  & {\color[HTML]{34696D} } &  \\
\multirow{-3}{*}{\begin{tabular}[c]{@{}l@{}}Wasserstein GAN\\ (WGAN) \citep{arjovskyWGAN}\end{tabular}} & \multirow{-2}{*}{{\color[HTML]{34696D} Require D to learn 1-Lipschitz functions}} & \multirow{-3}{*}{C$_2$} & \multirow{-2}{*}{{\color[HTML]{34696D} Weight Clipping}} & \multirow{-3}{*}{S$_1$(i)} \\ \hline \hline
\begin{tabular}[c]{@{}l@{}}A Two-Step Computation of\\ the exact GAN Wasserstein\\ Distance (WGAN-TS) \citep{wgan_ts}\end{tabular} & \begin{tabular}[c]{@{}l@{}}Vanishing and exploding gradient\\ due to weight clipping in WGAN\end{tabular} & C$_2$ & \begin{tabular}[c]{@{}l@{}}Two step formulation to compute \\ EM without weight clipping\end{tabular} & S$_1$(i) \\ \hline\hline

\begin{tabular}[c]{@{}l@{}}Least Squares GAN\\ (LSGAN) \citep{lsgan}\end{tabular} & {\color[HTML]{680100} \begin{tabular}[c]{@{}l@{}}Vanishing gradients due to \\ binary cross entropy loss\end{tabular}} & C$_2$ & {\color[HTML]{680100} \begin{tabular}[c]{@{}l@{}}Propose least square loss\\ or Pearson $\chi^2$ Divergence\end{tabular}} & S$_1$(i) \\ \hline \hline
 & {\color[HTML]{CE6301} \begin{tabular}[c]{@{}l@{}}Vanishing gradients due to over-\\ pessimistic $D$\end{tabular}} &  & {\color[HTML]{CE6301} \begin{tabular}[c]{@{}l@{}}Loss having a data-dependent\\ margin with gradients everywhere\end{tabular}} &  \\
 & {\color[HTML]{010066} \begin{tabular}[c]{@{}l@{}}Assuming infinite capacity for \\ convergence which leads\\ to mode collapse\end{tabular}} &  & {\color[HTML]{010066} \begin{tabular}[c]{@{}l@{}}Convergence proof without the  \\ assumption of infinite capacity.\\ Generalization bounds\end{tabular}} &  \\
\multirow{-5}{*}{\begin{tabular}[c]{@{}l@{}}Loss Sensitive GAN on \\ Lipschitz Densities \\ (LS-GAN) \citep{ls-gan}\end{tabular}} & {\color[HTML]{34696D} \begin{tabular}[c]{@{}l@{}}WGAN objective is unbounded\\ from above\end{tabular}} & \multirow{-5}{*}{C$_2$} & {\color[HTML]{34696D} \begin{tabular}[c]{@{}l@{}}Pairwise comparison unlike WGAN\\ where the loss is decomposed into\\ two first-order moments\end{tabular}} & \multirow{-5}{*}{\begin{tabular}[c]{@{}c@{}}S$_1$(i)\\ S$_4$(ii)\end{tabular}} \\ \hline\hline
 & {\color[HTML]{000000} Dimensionality misspecification} &  & {\color[HTML]{000000} Adding high dimensional noise} &  \\
\multirow{-2}{*}{\begin{tabular}[c]{@{}l@{}}Stabilizing GANs through\\ Regularization \citep{fgan_2}\end{tabular}} & {\color[HTML]{680100} Variance due to noise} & \multirow{-2}{*}{C$_2$} & {\color[HTML]{680100} Noise induced regularization} & \multirow{-2}{*}{S$_1$(ii)} \\ \hline \hline
\begin{tabular}[c]{@{}l@{}}Improved Training of\\ WGAN (WGAN-GP)\\ \citep{gularajani}\end{tabular} & {\color[HTML]{CE6301} \begin{tabular}[c]{@{}l@{}}Weight clipping in WGAN\\ causes vanishing and exploding \\ gradients and capacity underuse\end{tabular}} & C$_2$ & {\color[HTML]{CE6301} \begin{tabular}[c]{@{}l@{}}Introduce a penalty term on gradients\\ of $D$ w.r.t. the sample which lies\\ on a line between $x$ and $\hat{x}$\end{tabular}} & S$_1$(ii) \\ \hline
\begin{tabular}[c]{@{}l@{}}Improving the Improved\\ Training of WGANs\\ (CT-GAN) \citep{imp_wgan_gp}\end{tabular} & {\color[HTML]{010066} \begin{tabular}[c]{@{}l@{}}WGAN-GP regularization depends\\ on model samples which may not \\ lie close to the actual support initially. \\ Hence need lot of iterations \\ to ensure Lipschitz constraint\end{tabular}} & C$_2$ & {\color[HTML]{010066} \begin{tabular}[c]{@{}l@{}}Add a regularization through \\ Consistency term by perturbing \\ the real data sample itself, twice.\end{tabular}} & S$_1$(ii) \\ \hline \hline
 & {\color[HTML]{34696D} \begin{tabular}[c]{@{}l@{}}WGAN-GP requires the data and model \\ samples to be drawn from a certain \\ joint distribution\end{tabular}} &  & {\color[HTML]{34696D} \begin{tabular}[c]{@{}l@{}}In practice the data and models\\ samples are drawn from\\ marginal distributions\end{tabular}} &  \\
 & {\color[HTML]{000000} \begin{tabular}[c]{@{}l@{}}WGAN-GP assumes the optimal\\ critic to be differentiable\end{tabular}} &  & {\color[HTML]{000000} \begin{tabular}[c]{@{}l@{}}Prove that this assumption \\ does not hold true\end{tabular}} &  \\
\multirow{-5}{*}{\begin{tabular}[c]{@{}l@{}}On Regularization of \\ WGANs (WGAN-LP) \\  \citep{reg_wgan}\end{tabular}} & {\color[HTML]{680100} \begin{tabular}[c]{@{}l@{}}Weight clipping is also highly \\ restrictive strategy to ensure \\ Lipschitz condition\end{tabular}} & \multirow{-5}{*}{C$_2$} & {\color[HTML]{680100} \begin{tabular}[c]{@{}l@{}}Propose a less restrictive\\ regularization and also less\\ sensitive to hyperparameters\end{tabular}} & \multirow{-5}{*}{S$_1$(ii)} \\ \hline \hline
\begin{tabular}[c]{@{}l@{}}Fisher GAN\\ \citep{fisher}\end{tabular} & {\color[HTML]{CE6301} \begin{tabular}[c]{@{}l@{}}Weight clipping reduces the capacity\\ of the discriminator\\ WGAN-GP has high computational\\ cost\end{tabular}} & C$_2$ & {\color[HTML]{CE6301} \begin{tabular}[c]{@{}l@{}}Introduce data-dependent\\ regularization which maintains the\\ capacity of the critic while\\ ensuring stability\end{tabular}} & S$_1$(ii) \\ \hline \hline
 & {\color[HTML]{010066} \begin{tabular}[c]{@{}l@{}}Weight clipping reduced the rank \\ of the weight matrix\end{tabular}} &  & {\color[HTML]{010066} \begin{tabular}[c]{@{}l@{}}Regularization which performs \\ spectral normalization of weight \\ matrix and doesn't affect the rank\end{tabular}} &  \\
\multirow{-4}{*}{\begin{tabular}[c]{@{}l@{}}Spectral Normalization\\ for GANs (SN-GAN)\\ \citep{spectral}\end{tabular}} & {\color[HTML]{34696D} \begin{tabular}[c]{@{}l@{}}WGAN-GP introduces regularization\\ based on unreliable model samples\end{tabular}} & \multirow{-4}{*}{C$_2$} & {\color[HTML]{34696D} \begin{tabular}[c]{@{}l@{}}Not dependent on model samples\\ and less computationally complex\end{tabular}} & \multirow{-4}{*}{S$_1$(ii)} \\ \hline
\end{tabular}
}
\end{adjustbox}
 %\caption{Summary}
\end{table}

\newpage
\begin{table}[!htb]
  \centering
 \begin{adjustbox}{max width =0.95\textwidth}
  {\def\arraystretch{1.4}
  \begin{tabular}{||l|l|c|l|c||}
\hline
\multicolumn{1}{||l|}{\emph{Paper}} & \multicolumn{1}{l|}{\emph{Concerns Raised}} & \emph{Challenge} & \multicolumn{1}{l|}{\emph{Novel Approaches}} & \multicolumn{1}{c||}{\emph{Solution}} \\ \hline \hline
& {\color[HTML]{000000} } &  & \begin{tabular}[c]{@{}l@{}}Identify the cause based on the \\ Jacobian of gradients.\end{tabular} &  \\
 & {\color[HTML]{000000} } &  & {\color[HTML]{343434} \begin{tabular}[c]{@{}l@{}}Propose consensus optimization \\ based on regularization w.r.t. $\phi, \theta$\end{tabular}} &  \\
\multirow{-5}{*}{\begin{tabular}[c]{@{}l@{}}The Numerics of GANs\\ \citep{geiger06}\end{tabular}} & \multirow{-5}{*}{{\color[HTML]{000000} \begin{tabular}[c]{@{}l@{}}Non-convergence of simultaneous \\ gradient descent\end{tabular}}} & \multirow{-5}{*}{C$_2$} & {\color[HTML]{656565} Prove its convergence} & \multirow{-5}{*}{\begin{tabular}[c]{@{}c@{}}S$_1$(iii)\\ S$_3$\\ S$_4$(iii)\end{tabular}} \\ \hline \hline
 & {\color[HTML]{680100} \begin{tabular}[c]{@{}l@{}}GAN is not convex-concave \\ objective hence gradient descent\\ may not converge. WGAN has \\ non-convergent limit cycles\end{tabular}} &  & {\color[HTML]{680100} \begin{tabular}[c]{@{}l@{}}Use ODE method to prove that\\ GAN objective is locally\\ asymptotically stable under\\ certain conditions\end{tabular}} &  \\
\multirow{-5}{*}{\begin{tabular}[c]{@{}l@{}}Gradient Descent GAN \\ Optimization is Locally Stable \\ \citep{CMU17}\end{tabular}} & {\color[HTML]{CE6301} \begin{tabular}[c]{@{}l@{}}Local instability in a GAN \\ Framework\end{tabular}} & \multirow{-5}{*}{C$_2$} & {\color[HTML]{CE6301} \begin{tabular}[c]{@{}l@{}}Propose regularization on gradients\\ of discriminator for stability\end{tabular}} & \multirow{-5}{*}{\begin{tabular}[c]{@{}c@{}}S$_1$(iii)\\ S$_4$(iii)\end{tabular}} \\ \hline \hline
 & {\color[HTML]{010066} } &  & {\color[HTML]{010066} \begin{tabular}[c]{@{}l@{}}Noise induced regularization\\ \citep{fgan_2} converges.\end{tabular}} &  \\
\multirow{-4}{*}{\begin{tabular}[c]{@{}l@{}}Which Training Methods\\ for GANs do actually\\ converge? \citep{mescheder18icml}\end{tabular}} & \multirow{-4}{*}{{\color[HTML]{010066} \begin{tabular}[c]{@{}l@{}}Non-convergence of unregularized \\ GANs and WGAN-GP on \\ non-overlapping manifolds\end{tabular}}} & \multirow{-4}{*}{C$_2$} & {\color[HTML]{010066} \begin{tabular}[c]{@{}l@{}}Propose simplified version\\ of the above and prove \\ convergence\end{tabular}} & \multirow{-4}{*}{\begin{tabular}[c]{@{}l@{}}S$_1$(iii)\\ S$_4$(iii)\end{tabular}} \\ \hline \hline
 & {\color[HTML]{34696D} Mode Collapse} &  & {\color[HTML]{34696D} Introduce regularization} &  \\
 & {\color[HTML]{000000} \begin{tabular}[c]{@{}l@{}}Non-convergence of alternative\\ gradient descent\end{tabular}} &  & {\color[HTML]{000000} \begin{tabular}[c]{@{}l@{}}View the GAN optimization\\ as regret minimization.\end{tabular}} &  \\
 & {\color[HTML]{000000} \begin{tabular}[c]{@{}l@{}}GANs learn swiss roll distribution\\ despite vanishing gradients\end{tabular}} &  & {\color[HTML]{000000} \begin{tabular}[c]{@{}l@{}}Prove convergence for the \\ convex-concave case.\end{tabular}} &  \\
\multirow{-5}{*}{\begin{tabular}[c]{@{}l@{}}On Convergence and Stability\\ of GANs (DRAGAN)\\ \citep{DRAGAN}\end{tabular}} & {\color[HTML]{000000} \begin{tabular}[c]{@{}l@{}}WGAN-GP does not follow from\\ KR duality as WGAN does.\end{tabular}} & \multirow{-5}{*}{\begin{tabular}[c]{@{}c@{}}C$_1$\\ C$_2$\end{tabular}} & {\color[HTML]{000000} \begin{tabular}[c]{@{}l@{}}Converge to $\epsilon$- approximate equilibrium\\ in non-convex case\end{tabular}} & \multirow{-5}{*}{\begin{tabular}[c]{@{}l@{}}S$_1$(iii)\\ S$_4$(iii)\end{tabular}} \\ \hline \hline
 & {\color[HTML]{680100} \begin{tabular}[c]{@{}l@{}}Non-convergence of WGAN due\\ to biased gradient estimator\end{tabular}} &  & {\color[HTML]{680100} \begin{tabular}[c]{@{}l@{}}Propose Cramer distance with \\ unbiased sample gradients\end{tabular}} &  \\
\multirow{-3}{*}{\begin{tabular}[c]{@{}l@{}}Cramer Distance as a Solution\\ to Biased Wasserstein\\ Gradients \citep{cramer}\end{tabular}} & {\color[HTML]{CE6301} \begin{tabular}[c]{@{}l@{}}Powerful critic is needed and also \\ should not over-fit the empirical\\ distribution\end{tabular}} & \multirow{-2}{*}{C$_2$} & {\color[HTML]{CE6301} \begin{tabular}[c]{@{}l@{}}Cramer distance enables learning\\ without perfect critic\end{tabular}} & \multirow{-2}{*}{S$_1$(iv)} \\ \hline \hline
\begin{tabular}[c]{@{}l@{}}Learning Generative Models\\ with Sinkhorn Divergences\\ \citep{sinkhorn}\end{tabular} & {\color[HTML]{010066} \begin{tabular}[c]{@{}l@{}}Biased gradient estimator in GANs\\ No results on sample complexity\end{tabular}} & C$_2$ & {\color[HTML]{010066} \begin{tabular}[c]{@{}l@{}}Propose Sinkhorn divergence with\\ entropic smoothing to make it \\ differentiable.\\ It includes properties of MMD \\ distance metric which has\\ unbiased gradient estimator and\\ favourable sample complexity\end{tabular}} & \multicolumn{1}{c||}{S$_1$(iv)} \\ \hline \hline
 & {\color[HTML]{34696D} \begin{tabular}[c]{@{}l@{}}In WGAN, it is not possible to \\ optimize over all possible 1-Lipschitz \\ functions leading to imperfect critic \end{tabular}} &  & {\color[HTML]{34696D} \begin{tabular}[c]{@{}l@{}}Propose a new distance metric\\ Mini-batch energy distance does\\ not require Lipschitz assumption\end{tabular}} &  \\
\multirow{-3}{*}{\begin{tabular}[c]{@{}l@{}}Improving GANs using \\ Optimal Transport\\ OT_GAN \citep{ot}\end{tabular}} & {\color[HTML]{000000} \begin{tabular}[c]{@{}l@{}}Sinkhorn distance has biased\\ sample gradients\end{tabular}} & \multirow{-3}{*}{C$_2$} & {\color[HTML]{000000} \begin{tabular}[c]{@{}l@{}}Mini-batch energy distance uses \\ Sinkhorn distance along with\\ Generalized energy distance  \\ hence has unbiased estimator\end{tabular}} & \multirow{-3}{*}{S$_1$(iv)} \\ \hline \hline
\begin{tabular}[c]{@{}l@{}}Demystifying MMD GANs\\ \citep{mmd}\end{tabular} & {\color[HTML]{680100} \begin{tabular}[c]{@{}l@{}}Wasserstein distance leads to \\ biased sample gradients\end{tabular}} & C$_2$ & {\color[HTML]{680100} \begin{tabular}[c]{@{}l@{}}The natural maximum mean \\ discrepancy metric provides\\ unbiased gradients\end{tabular}} & \multicolumn{1}{c||}{S$_1$(iv)} \\ \hline \hline
\begin{tabular}[c]{@{}l@{}}Unrolled GANs\\ \citep{unrolled}\end{tabular} & {\color[HTML]{CE6301} \begin{tabular}[c]{@{}l@{}}Mode collapse as $D$ cannot \\ be trained till optimality \\ at every iteration. $G$ moves \\ mass to a single point and $D$\\ assigns lower probability to it.\end{tabular}} & C$_1$ & {\color[HTML]{CE6301} \begin{tabular}[c]{@{}l@{}}Introduce a surrogate loss which\\ in limit equals the optimal $D$\\ The $G$ is updated based on the\\ future update of $D$ hence \\ reducing mode collapse\end{tabular}} & \multicolumn{1}{c||}{S$_1$(v)} \\ \hline 
\end{tabular}
 }
 \end{adjustbox}
  %\caption{Summary}
\end{table}

\newpage
\begin{table}[!htb]
\centering
\begin{adjustbox}{max width =0.95\textwidth}
{\def\arraystretch{1.4}
\begin{tabular}{||l|l|c|l|c||}
\hline
\multicolumn{1}{||l|}{\emph{Paper}} & \multicolumn{1}{l|}{\emph{Concerns Raised}} & \multicolumn{1}{c|}{\emph{Challenge}} & \multicolumn{1}{l|}{\emph{Novel Approach}} & \emph{Solution} \\ \hline
\hline
\begin{tabular}[c]{@{}l@{}}Coloumb GANs: Provably \\ Optimal Nash Equilibria\\ via Potential Fields \\ \citep{coulomb}\end{tabular} & {\color[HTML]{010066} \begin{tabular}[c]{@{}l@{}}GANs converge to local Nash\\ equilibrium problem causing\\ mode collapse\end{tabular}} & \begin{tabular}[c]{@{}c@{}}C$_1$\\ C$_2$\end{tabular} & {\color[HTML]{010066} \begin{tabular}[c]{@{}l@{}}Coloumb GANs have unique\\ Nash equilibrium. Prove convergence \\ for non parametric settings\end{tabular}} & \multicolumn{1}{c||}{\begin{tabular}[c]{@{}c@{}}S$_1$(v)\\ S$_4$(ii)\end{tabular}} \\ \hline \hline
\begin{tabular}[c]{@{}l@{}}Unsupervised Representation \\ Learning with Deep Convolutional \\ GANs (DCGAN) \citep{dcgan}\end{tabular} & {\color[HTML]{34696D} Unstable training of Vanilla GANs} & C$_2$ & {\color[HTML]{34696D} \begin{tabular}[c]{@{}l@{}}Introduce convolutional layers \\ and other heuristics for stability\end{tabular}} & \multicolumn{1}{c||}{S$_2$} \\ \hline\hline
\begin{tabular}[c]{@{}l@{}}Mode Regularized GANs\\ \citep{modeReg}\end{tabular} & \begin{tabular}[c]{@{}l@{}}Mode collapse: large modes have\\ better discriminator gradients \\ hence for all $z$ the generator is\\ pushed towards major modes.\end{tabular} & C$_1$ & \begin{tabular}[c]{@{}l@{}}Introduce geometric metric\\ regularizer using an encoder \\ which maps all the $z$ to $x$\\ and $G$ maps $z$ to $\hat{x}$\end{tabular} & \multicolumn{1}{c||}{S$_2$} \\ \hline\hline
 & {\color[HTML]{680100} \begin{tabular}[c]{@{}l@{}}Difficult to balance the training\\ between $G$ and $D$ in WGAN\end{tabular}} &  & {\color[HTML]{680100} \begin{tabular}[c]{@{}l@{}}Propose an auto-encoder based\\ loss and objective for balance\end{tabular}} & \multicolumn{1}{c||}{} \\
 & {\color[HTML]{CE6301} Mode Collapse} &  & {\color[HTML]{CE6301} \begin{tabular}[c]{@{}l@{}}Introduce diversity ratio to \\ generate diverse samples\end{tabular}} & \multicolumn{1}{c||}{} \\
\multirow{-4}{*}{\begin{tabular}[c]{@{}l@{}}Boundary Equilibrium GANs\\ (BEGAN) \citep{began}\end{tabular}} & {\color[HTML]{010066} \begin{tabular}[c]{@{}l@{}}Cannot know convergence based\\ on loss plots during training\end{tabular}} & \multirow{-4}{*}{\begin{tabular}[c]{@{}c@{}}C$_1$\\ C$_2$\end{tabular}} & {\color[HTML]{010066} \begin{tabular}[c]{@{}l@{}}Introduce a convergence \\ measure for better training\end{tabular}} & \multicolumn{1}{c||}{\multirow{-4}{*}{S$_2$}} \\ \hline\hline
& {\color[HTML]{34696D} \begin{tabular}[c]{@{}l@{}}With binary logistic loss, the \\ gradients for different samples \\ are not orthogonal which \\ results in inefficient training\end{tabular}} &  & {\color[HTML]{34696D} \begin{tabular}[c]{@{}l@{}}Auto-encoder based reconstruction\\ loss which provides gradients in\\ different directions for each sample\\ for efficient training\end{tabular}} &  \\
\multirow{-4}{*}{\begin{tabular}[c]{@{}l@{}}Energy-Based GAN\\ (EBGAN) \citep{eb-gan}\end{tabular}} & {\color[HTML]{000000} Mode Collapse} & \multirow{-4}{*}{\begin{tabular}[c]{@{}c@{}}C$_1$\\ C$_2$\end{tabular}} & {\color[HTML]{000000} \begin{tabular}[c]{@{}l@{}}Repelling regularizer to \\ orthogonalize sample representations\end{tabular}} & \multirow{-4}{*}{S$_2$(i)} \\ \hline \hline
\begin{tabular}[c]{@{}l@{}}VEEGAN: Reducing Mode \\ Collapse  in GANs Using \\ Implicit Variational Learning \\ \citep{veegan}\end{tabular} & {\color[HTML]{680100} \begin{tabular}[c]{@{}l@{}}The vanilla GAN objective suffers\\ from mode collapse i.e., \\ $p_g \neq p_d$\end{tabular}} & C$_1$ & {\color[HTML]{680100} \begin{tabular}[c]{@{}l@{}}Introduce a reconstructor\\ network\\ i) Maps $x$ to $z$ ii) Maps $\hat{x}$ to $z$\\ This is possible iff $p_g = p_d$\end{tabular}} & S$_2$(i) \\ \hline \hline
\begin{tabular}[c]{@{}l@{}}Variational approaches \\ for auto-encoding \\ GANs (AE-GAN) \citep{variational}\end{tabular} & {\color[HTML]{CE6301} \begin{tabular}[c]{@{}l@{}}Mode Collapse in GANs  but not\\ in VAE\end{tabular}} & C$_1$ & {\color[HTML]{CE6301} \begin{tabular}[c]{@{}l@{}}The authors propose a model\\ combining GANs and VAE\end{tabular}} & S$_2$(i) \\ \hline \hline
\begin{tabular}[c]{@{}l@{}}Multi-Agent Diverse GANs\\ (MAD-GANs) \citep{madgan}\end{tabular} & {\color[HTML]{010066} \begin{tabular}[c]{@{}l@{}}Mode collapse in vanilla GANs\\ with one $G$\end{tabular}} & C$_1$ & {\color[HTML]{010066} \begin{tabular}[c]{@{}l@{}}Propose multiple $G$ and force them\\ to learn different modes as the $D$\\ has to minimize BCE and also identify\\ the $G$ which generated the sample\end{tabular}} & S$_2$(ii) \\ \hline \hline
 & {\color[HTML]{34696D} \begin{tabular}[c]{@{}l@{}}Vanilla GAN does not guarantee \\ convergence in finite steps\end{tabular}} &  & {\color[HTML]{34696D} \begin{tabular}[c]{@{}l@{}}Propose a mixture model \\ (mixture of weak generators)\\ and provide sufficient and necessary \\ conditions for convergence\end{tabular}} &  \\
\multirow{-4}{*}{\begin{tabular}[c]{@{}l@{}}AdaGAN: Boosting Generative\\ Models \citep{adagan}\end{tabular}} & {\color[HTML]{000000} Mode collapse} & \multirow{-4}{*}{\begin{tabular}[c]{@{}c@{}}C$_1$\\ C$_2$\end{tabular}} & {\color[HTML]{000000} \begin{tabular}[c]{@{}l@{}}They sequentially combine generators\\ at every step such that the samples \\ missed are given more weightage\end{tabular}} & \multirow{-4}{*}{\begin{tabular}[c]{@{}c@{}}S$_2$(ii)\\ S$_4$(iii)\end{tabular}} \\ \hline \hline
Evolutionary GANs \citep{egan} & \begin{tabular}[c]{@{}l@{}}KL divergence suffers from\\ vanishing gradients\\ WGAN has non-convergent\\ limit cycles near equilibrium\end{tabular} & C$_2$ & \begin{tabular}[c]{@{}l@{}}Jointly optimize over $G$'s\\ trained on different objectives.\\ Overcome the weakness of each\\ by preserving the best $G$'s at every step\end{tabular} & $S_2$(ii) \\ \hline \hline
\begin{tabular}[c]{@{}l@{}}Dual Discriminator GANs\\ (D2GAN) \citep{d2gan}\end{tabular} & {\color[HTML]{9A0000} \begin{tabular}[c]{@{}l@{}}KL divergence metric covers all\\ modes but also allows unreal images \\ Reverse KL suffers mode \\ collapse\end{tabular}} & C$_1$ & {\color[HTML]{9A0000} \begin{tabular}[c]{@{}l@{}}Propose to combine two $D$'s\\ on trained on KL and other on\\ reverse KL and prove convergence\\ to Nash Equilibrium\end{tabular}} & \begin{tabular}[c]{@{}c@{}}$S_2$(ii)\\ $S_4$(ii)\end{tabular} \\ \hline \hline

\end{tabular}
}
\end{adjustbox}
 %\caption{Summary}
\end{table}

\newpage

\begin{table}[!htb]
\centering
\begin{adjustbox}{max width =0.95\textwidth}
{\def\arraystretch{1.4}
\begin{tabular}{||l|l|c|l|c||}
\hline
\multicolumn{1}{||l|}{\emph{Paper}} & \multicolumn{1}{l|}{\emph{Concerns Raised}} & \emph{Challenge} & \multicolumn{1}{l|}{\emph{Novel Approach}} & \emph{Solution} \\ \hline \hline
 
 & {\color[HTML]{CE6301} \begin{tabular}[c]{@{}l@{}}Unimodal $z$ to embed multiple\\ classes\end{tabular}} &  & {\color[HTML]{CE6301} \begin{tabular}[c]{@{}l@{}}Jointly learn continuous variable $z$\\ and class variable $c$ to enforce\\ discontinuity in classes\end{tabular}} & \multicolumn{1}{l||}{} \\
\multirow{-4}{*}{\begin{tabular}[c]{@{}l@{}}Memorization precedes generation: \\Learning unsupervised GANs with \\ Memory Networks \citep{memory}\end{tabular}} & {\color[HTML]{010066} \begin{tabular}[c]{@{}l@{}}$D$ forgets about the previous\\ samples generated which causes\\ divergence in training\end{tabular}} & \multirow{-4}{*}{C$_2$} & {\color[HTML]{010066} \begin{tabular}[c]{@{}l@{}}They propose a memory network \\ to mitigate the issue.\end{tabular}} & \multicolumn{1}{l||}{\multirow{-4}{*}{S$_2$(iii)}} \\ \hline \hline
\begin{tabular}[c]{@{}l@{}}Training GANs with Optimism \\\citep{optimism}\end{tabular} & {\color[HTML]{34696D} \begin{tabular}[c]{@{}l@{}}WGAN suffer from limit\\ cycling behavior at equilibrium\\ No variant of GD can converge \\ in terms of last weight even in\\ convex-concave setting\end{tabular}} & C$_2$ & {\color[HTML]{34696D} \begin{tabular}[c]{@{}l@{}}Propose optimistic mirror descent\\ which converges w.r.t. the last weight\\ for bilinear functions and faster rates\\ of convergence in terms of average\\ of weights\end{tabular}} & S$_3$ \\ \hline \hline
 & {\color[HTML]{000000} \begin{tabular}[c]{@{}l@{}}Vanishing gradient doesn't \\ occur when using Equation\\ \ref{eq:gan_loss2} for $G$\end{tabular}} &  & {\color[HTML]{000000} \begin{tabular}[c]{@{}l@{}}Show that GANs using Equation \\ \ref{eq:gan_loss2} do not minimize\\ divergence experimentally\end{tabular}} &  \\
\multirow{-4}{*}{\begin{tabular}[c]{@{}l@{}}Many Paths to Equilibrium: \\ Gans Do Not Need To\\ Decrease Divergence\\ At Every Step \citep{many}\end{tabular}} & {\color[HTML]{680100} \begin{tabular}[c]{@{}l@{}}Unstable training and sensitivity\\ to hyperparameters\end{tabular}} & \multirow{-4}{*}{C$_2$} & {\color[HTML]{680100} \begin{tabular}[c]{@{}l@{}}Show experimentally that gradient\\ penalties overcome the issues.\end{tabular}} & \multirow{-4}{*}{S$_4$ (iii)} \\ \hline\hline

 & {\color[HTML]{CE6301} \begin{tabular}[c]{@{}l@{}}Existence of pure strategy\\ Nash equilibrium is not\\ guaranteed but mixed strategy\\ Nash always exists\end{tabular}} &  & {\color[HTML]{CE6301} \begin{tabular}[c]{@{}l@{}}Introduce MIX-GAN that combines\\ multiple $G$'s and $D$'s, which \\ converges to the mixed strategy\\ Nash equilibrium\end{tabular}} & \multicolumn{1}{l||}{} \\
\multirow{-4}{*}{\begin{tabular}[c]{@{}l@{}}Generalization and Equilibrium\\ in GANs \citep{Arora03}\end{tabular}} & {\color[HTML]{010066} \begin{tabular}[c]{@{}l@{}}No generalization bounds on\\ vanilla GAN objective\end{tabular}} & \multirow{-4}{*}{C$_2$} & {\color[HTML]{010066} \begin{tabular}[c]{@{}l@{}}Introduce NN distance and prove\\ generalization bound that depends\\ on the number of parameters in $D$\end{tabular}} & \multicolumn{1}{l||}{\multirow{-4}{*}{\begin{tabular}[c]{@{}l@{}}S$_2$(ii)\\ S$_4$(i)\end{tabular}}} \\ \hline\hline
\begin{tabular}[c]{@{}l@{}}Do GANs Learn the Distribution?\\ Some Theory and Empirics \\ \citep{doarora}\end{tabular} & {\color[HTML]{34696D} \begin{tabular}[c]{@{}l@{}}There has been no ways to \\ quantify mode collapse i.e.,\\ the sample diversity\end{tabular}} & C$_1$ & {\color[HTML]{34696D} \begin{tabular}[c]{@{}l@{}}Introduce birthday paradox test\\ to measure sample diversity\end{tabular}} & \multicolumn{1}{l||}{\begin{tabular}[c]{@{}l@{}}S$_4$(i)\\ S$_4$(ii)\end{tabular}} \\ \hline \hline
\begin{tabular}[c]{@{}l@{}}Approximation and Convergence\\ Properties of Generative \\ Adversarial Learning \citep{app_conv}\end{tabular} & {\color[HTML]{000000} \begin{tabular}[c]{@{}l@{}}It is difficult to estimate the\\ number of parameters required\\ for convergence\end{tabular}} & C$_2$ & {\color[HTML]{000000} \begin{tabular}[c]{@{}l@{}}Define adversarial divergence a \\ generalized framework for GAN loss\\ and establish a relation between\\ number of parameters in $D$ and\\ convergence\end{tabular}} & \begin{tabular}[c]{@{}c@{}}S$_4$(i)\\ S$_4$(ii)\end{tabular} \\ \hline \hline
\begin{tabular}[c]{@{}l@{}}$f$GAN: Training Generative \\ Neural  Samplers using\\ Variational Divergence \\ Minimization \citep{fgan}\end{tabular} & {\color[HTML]{680100} \begin{tabular}[c]{@{}l@{}}Why does single step gradient \\ descent for $G$ and $D$ at each \\ iteration converges sometimes\end{tabular}} & C$_2$ & {\color[HTML]{680100} \begin{tabular}[c]{@{}l@{}}Generalize GAN objective to\\ variational divergence \\ minimization and propose \\ algorithms for which they prove \\ the  convergence to saddle point\\ under certain conditions\end{tabular}} & S$_4$(iii) \\ \hline \hline
 & {\color[HTML]{CE6301} \begin{tabular}[c]{@{}l@{}}Non-convergence in GANs\\ Convergence proofs have unrealistic\\ assumptions\\ Prove convergence for when \\ samples go to infinity (not practical)\end{tabular}} &  & {\color[HTML]{CE6301} \begin{tabular}[c]{@{}l@{}}Propose two time-scale update rule\\ Prove the convergence of loss when\\ estimated with minibatch of samples \\ to stationary local Nash equilibrium, \\ under assumptions\end{tabular}} &  \\
\multirow{-5}{*}{\begin{tabular}[c]{@{}l@{}}GANs Trained by a Two Time-Scale \\ Update Rule Converge to a Local \\ Nash Equilibrium \citep{ttur}\end{tabular}} & {\color[HTML]{010066} Mode Collapse} & \multirow{-5}{*}{\begin{tabular}[c]{@{}c@{}}C$_1$\\ C$_2$\end{tabular}} & {\color[HTML]{010066} \begin{tabular}[c]{@{}l@{}}Propose that Adam optimizer which\\ depends on average of past gradients\\ hence avoids local minima\end{tabular}} & \multirow{-5}{*}{\begin{tabular}[c]{@{}c@{}}S$_3$\\ S$_4$(iii)\end{tabular}} \\ \hline\hline
\begin{tabular}[c]{@{}l@{}}MMD GAN: Towards Deeper \\ Understanding of Moment Matching \\ Network \citep{mmd-gan}\end{tabular} & {\color[HTML]{34696D} \begin{tabular}[c]{@{}l@{}}No proper convergence theory for\\ GANs unlike in Generative \\ Moment Matching Networks (GMMN)\\ Although the empirical results from\\ GMMN are poorer than GANs\end{tabular}} & C$_2$ & {\color[HTML]{34696D} \begin{tabular}[c]{@{}l@{}}Propose to improve upon GMMNs\\ by introducing adversarial loss,\\ which requires an auto-encoder\\ architecture (MMD-GAN). WGAN is \\ a special case of MMD-GAN\end{tabular}} & S$_2$(i) \\ \hline
\end{tabular}
}
\end{adjustbox}
 \caption{Summary}\label{tab:summary}
\end{table}

\newpage

\begin{table}
\centering
\begin{adjustbox}{max width =0.95\textwidth}
{\def\arraystretch{1.4}
\begin{tabular}{|c|c|c|c|c|c|c|c|c|c|c|c|c|c|c|c|}
\hline
\emph{Paper} & C$_1$ & C$_2$ & S$_1$(i) & S$_1$(ii) & S$_1$(iii) & S$_1$(iv) & S$_1$(v) & S$_2$ & S$_2$(i) & S$_2$(ii) & S$_2$(iii) & S$_3$ & S$_4$(i) & S$_4$(ii) & S$_4$(iii) \\ \hline \hline
\cite{goodfellow16} & \checkmark & \checkmark  &   \checkmark &  &  &  & \checkmark  &  &  &  &  &  &  &  &    \\ \hline \hline
\cite{arjovsky01} &  & \checkmark &   \checkmark &  &  &  &  &  &  &  &  &  &    &  &  \\ \hline \hline
\cite{arjovskyWGAN} &  & \checkmark &   \checkmark &  &  &  &  &  &  &  &  &   &  &  &  \\ \hline \hline
\cite{wgan_ts} &  & \checkmark &   \checkmark &  &  &  &  &  &  &  &  &  &  &    &  \\ \hline \hline
\cite{lsgan} &  & \checkmark &   \checkmark &  &  &  &  &  &  &  &  &  &  &  &  \\ \hline \hline
\cite{ls-gan} &  & \checkmark &   \checkmark &  &  &  &  &  &  &  &  &    &  & \checkmark &  \\ \hline \hline
\cite{fgan_2} &  & \checkmark &   & \checkmark &  &  &  &  &  &  &  &   &  &  &  \\ \hline \hline
\cite{gularajani} &  & \checkmark &    & \checkmark &  &  &  &  &  &  &   &  &  &  &  \\ \hline \hline
\cite{imp_wgan_gp} &  & \checkmark &    & \checkmark &  &  &  &  &  &  &    &  &  &  &  \\ \hline\hline
\cite{reg_wgan} &  & \checkmark &   & \checkmark &  &  &  &  &  &  &  &    &  &  &  \\ \hline\hline
\cite{fisher} &  & \checkmark &    & \checkmark &  &  &  &  &  &  &  &    &  &  &  \\ \hline\hline
\cite{spectral} &  & \checkmark &   & \checkmark &  &  &  &  &  &  &   &  &  &  &  \\ \hline\hline
\cite{geiger06} &  & \checkmark &    &  & \checkmark &  &  &  &  &  &  & \checkmark &    &  & \checkmark \\ \hline\hline
\cite{CMU17} &  & \checkmark &    &  & \checkmark &  &  &  &  &  &  &  &  &    & \checkmark \\ \hline\hline
\cite{mescheder18icml} &  & \checkmark &    &  & \checkmark &  &  &  &  &  &  &  &  &  & \checkmark \\ \hline\hline
\cite{DRAGAN} & \checkmark & \checkmark &    &  & \checkmark &  &  &  &  &  &  &   &  &  & \checkmark \\ \hline\hline
\cite{cramer} &  & \checkmark &    &  &  & \checkmark &  &  &  &  &  &  &  &    &  \\ \hline\hline
\cite{sinkhorn} &  & \checkmark &    &  &  & \checkmark &  &  &  &  &  &  &   &  &  \\ \hline\hline
\cite{ot} &  & \checkmark &    &  &  & \checkmark &  &  &  &  &  &  &  &  &  \\ \hline\hline
\cite{mmd} &  & \checkmark &    &  &  & \checkmark &  &  &  &  &  &  &    &  &  \\ \hline\hline
\cite{unrolled} & \checkmark &    &  &  &  &  & \checkmark &  &  &  &  &    &  &  &  \\ \hline\hline
\cite{coulomb} & \checkmark & \checkmark &    &  &  &  & \checkmark &  &  &  &  &    &  & \checkmark &  \\ \hline\hline
\cite{dcgan} &  & \checkmark &    &  &  &  &  & \checkmark &  &  &  &  &  &    &  \\ \hline\hline
\cite{ali} &  &   &  &  &  &  &  &  & \checkmark &  &  &  &    &  &  \\ \hline\hline
\cite{modeReg} & \checkmark &    &  &  &  &  &  &  & \checkmark &  &  &  &    &  &  \\ \hline\hline
\cite{began} & \checkmark & \checkmark &    &  &  &  &  &  & \checkmark &  &    &  &  &  &  \\ \hline\hline
\cite{eb-gan} & \checkmark & \checkmark &    &  &  &  &  &  & \checkmark &  &   &  &  &  &  \\ \hline\hline
\cite{veegan} & \checkmark &  &   &  &  &  &  &  & \checkmark &  &  &  &  &    &  \\ \hline\hline
\cite{variational} & \checkmark &    &  &  &  &  &  &  & \checkmark &  &    &  &  &  &  \\ \hline\hline
\cite{mmd-gan} &  & \checkmark &    &  &  &  &  &  & \checkmark &  &  &  &  &    &  \\ \hline\hline
\cite{madgan} & \checkmark &  &    &  &  &  &  &  &  & \checkmark &  &    &  &  &  \\ \hline
\cite{adagan} & \checkmark & \checkmark &    &  &  &  &  &  &  & \checkmark &  &  &  &  & \checkmark \\ \hline\hline
\cite{egan} &  & \checkmark &  &   &  &  &  &  &  & \checkmark &  &  &  &    &  \\ \hline\hline
\cite{d2gan} & \checkmark &  &    &  &  &  &  &  &  & \checkmark &  &  &    & \checkmark &  \\ \hline\hline
\cite{memory} &  & \checkmark &    &  &  &  &  &  &  &  & \checkmark &  &    &  &  \\ \hline\hline
\cite{optimism} &  & \checkmark &    &  &  &  &  &  &  &  &  & \checkmark &    &  &  \\ \hline\hline
\cite{many} &  & \checkmark &    &  &  &  &  &  &  &  &  &  &  &    & \checkmark \\ \hline\hline
\cite{Arora03} &  & \checkmark &   &  &  &  &  &  &  & \checkmark &  &   & \checkmark &  &  \\ \hline\hline
\cite{doarora} & \checkmark &    &  &  &  &  &  &  &  &  &    &  & \checkmark & \checkmark &  \\ \hline\hline
\cite{convDual} &  & \checkmark  & \checkmark &  &  &  &  &  &  &  &    &  & \checkmark &  &  \\ \hline\hline
\cite{app_conv} &  & \checkmark &    &  &  &  &  &  &  &  &  &    & \checkmark & \checkmark &  \\ \hline\hline
\cite{fgan} &  & \checkmark &    &  &  &  &  &  &  &  &  &  &    &  & \checkmark \\ \hline\hline
\cite{ttur} & \checkmark & \checkmark &    &  &  &  &  &  &  &  &  & \checkmark &    &  & \checkmark \\ \hline\hline
\end{tabular}
}
\end{adjustbox}
\caption{Papers In a Glance}
\label{tab:glance}
\end{table}

%%%%%%%%%%%%%%%%%%%%%%%%%%%%%%%%%%%%%%%%%%%%%%%%%%%%%%%%%%%%%%%%%%%%%%%%%%%%%%%%%%%%%%%%%%%%%%%%%%%%%%%%%%%%%%
\section{Conclusion}
\label{sec:con}

GANs have opened up a new approach for generative modeling of data distribution. With simple implementation, it can produce qualitative results beating the existing state-of-art. Despite its major success, the training of GANs suffers from different challenges. Moreover, there is a lack of a proper theoretical framework, which explains the convergence of GANs satisfactorily. In this summary paper, we briefly discussed and categorized the recent papers which contribute to the theoretical understanding of GANs. We have categorized the papers based on the two primary issues of mode collapse and non-convergence. We also identified the four main types of solutions proposed and further categorized the papers based on this. Through this, we get an idea of the general approach undertaken by the community at large for solving the issues pertaining to GANs. 
Finally, we have also provided samples of images generated by a few of the approaches discussed to compare the results visually. We hope that this summary guides future research based on the areas which are still unexplored or explored less. It is also possible to combine various solution techniques proposed in different papers for better results.

\vskip 0.2in
\bibliography{ref}

\begin{thebibliography}{69}
\providecommand{\natexlab}[1]{#1}
\providecommand{\url}[1]{\texttt{#1}}
\expandafter\ifx\csname urlstyle\endcsname\relax
  \providecommand{\doi}[1]{doi: #1}\else
  \providecommand{\doi}{doi: \begingroup \urlstyle{rm}\Url}\fi

\bibitem[{Arjovsky} and {Bottou}(2017)]{arjovsky01}
M.~{Arjovsky} and L.~{Bottou}.
\newblock {Towards Principled Methods for Training Generative Adversarial
  Networks}.
\newblock \emph{ArXiv e-prints}, January 2017.

\bibitem[{Arjovsky} et~al.(2017){Arjovsky}, {Chintala}, and
  {Bottou}]{arjovskyWGAN}
M.~{Arjovsky}, S.~{Chintala}, and L.~{Bottou}.
\newblock {Wasserstein GAN}.
\newblock \emph{ArXiv e-prints}, January 2017.

\bibitem[{Arora} et~al.(2017){Arora}, {Ge}, {Liang}, {Ma}, and
  {Zhang}]{Arora03}
S.~{Arora}, R.~{Ge}, Y.~{Liang}, T.~{Ma}, and Y.~{Zhang}.
\newblock {Generalization and Equilibrium in Generative Adversarial Nets
  (GANs)}.
\newblock \emph{ArXiv e-prints}, March 2017.

\bibitem[Arora et~al.(2018)Arora, Risteski, and Zhang]{doarora}
Sanjeev Arora, Andrej Risteski, and Yi~Zhang.
\newblock Do {GAN}s learn the distribution? some theory and empirics.
\newblock In \emph{International Conference on Learning Representations}, 2018.
\newblock URL \url{https://openreview.net/forum?id=BJehNfW0-}.

\bibitem[Bellemare et~al.(2018)Bellemare, Danihelka, Dabney, Mohamed,
  Lakshminarayanan, Hoyer, and Munos]{cramer}
Marc~G. Bellemare, Ivo Danihelka, Will Dabney, Shakir Mohamed, Balaji
  Lakshminarayanan, Stephan Hoyer, and Remi Munos.
\newblock The cramer distance as a solution to biased wasserstein gradients,
  2018.
\newblock URL \url{https://openreview.net/forum?id=S1m6h21Cb}.

\bibitem[Berthelot et~al.(2017)Berthelot, Schumm, and Metz]{began}
David Berthelot, Tom Schumm, and Luke Metz.
\newblock Began: Boundary equilibrium generative adversarial networks.
\newblock \emph{ArXiv}, abs/1703.10717, 2017.

\bibitem[Bińkowski et~al.(2018)Bińkowski, Sutherland, Arbel, and
  Gretton]{mmd}
Mikołaj Bińkowski, Dougal~J. Sutherland, Michael Arbel, and Arthur Gretton.
\newblock Demystifying {MMD} {GAN}s.
\newblock In \emph{International Conference on Learning Representations}, 2018.
\newblock URL \url{https://openreview.net/forum?id=r1lUOzWCW}.

\bibitem[{Borji}(2018)]{proCon}
A.~{Borji}.
\newblock {Pros and Cons of GAN Evaluation Measures}.
\newblock \emph{ArXiv e-prints}, February 2018.

\bibitem[Che et~al.(2016)Che, Li, Jacob, Bengio, and Li]{modeReg}
Tong Che, Yanran Li, Athul~Paul Jacob, Yoshua Bengio, and Wenjie Li.
\newblock Mode regularized generative adversarial networks.
\newblock \emph{CoRR}, abs/1612.02136, 2016.
\newblock URL \url{http://arxiv.org/abs/1612.02136}.

\bibitem[Daskalakis et~al.(2018)Daskalakis, Ilyas, Syrgkanis, and
  Zeng]{optimism}
Constantinos Daskalakis, Andrew Ilyas, Vasilis Syrgkanis, and Haoyang Zeng.
\newblock Training {GAN}s with optimism.
\newblock In \emph{International Conference on Learning Representations}, 2018.
\newblock URL \url{https://openreview.net/forum?id=SJJySbbAZ}.

\bibitem[Dong et~al.(2018)Dong, Hsiao, Yang, and Yang]{musegan}
Hao-Wen Dong, Wen-Yi Hsiao, Li-Chia Yang, and Yi-Hsuan Yang.
\newblock Musegan: Multi-track sequential generative adversarial networks for
  symbolic music generation and accompaniment.
\newblock In \emph{Thirty-Second AAAI Conference on Artificial Intelligence},
  2018.

\bibitem[Dumoulin et~al.(2016)Dumoulin, Belghazi, Poole, Mastropietro, Lamb,
  Arjovsky, and Courville]{ali}
Vincent Dumoulin, Ishmael Belghazi, Ben Poole, Olivier Mastropietro, Alex Lamb,
  Martin Arjovsky, and Aaron Courville.
\newblock Adversarially learned inference.
\newblock \emph{arXiv preprint arXiv:1606.00704}, 2016.

\bibitem[Farnia and Tse(2018)]{convDual}
Farzan Farnia and David Tse.
\newblock A convex duality framework for gans.
\newblock In \emph{Proceedings of the 32Nd International Conference on Neural
  Information Processing Systems}, NIPS'18, pages 5254--5263, USA, 2018. Curran
  Associates Inc.
\newblock URL \url{http://dl.acm.org/citation.cfm?id=3327345.3327431}.

\bibitem[Fedus* et~al.(2018)Fedus*, Rosca*, Lakshminarayanan, Dai, Mohamed, and
  Goodfellow]{many}
William Fedus*, Mihaela Rosca*, Balaji Lakshminarayanan, Andrew~M. Dai, Shakir
  Mohamed, and Ian Goodfellow.
\newblock Many paths to equilibrium: {GAN}s do not need to decrease a
  divergence at every step.
\newblock In \emph{International Conference on Learning Representations}, 2018.
\newblock URL \url{https://openreview.net/forum?id=ByQpn1ZA-}.

\bibitem[Frid-Adar et~al.(2018)Frid-Adar, Diamant, Klang, Amitai, Goldberger,
  and Greenspan]{anomaly}
Maayan Frid-Adar, Idit Diamant, Eyal Klang, Michal Amitai, Jacob Goldberger,
  and Hayit Greenspan.
\newblock Gan-based synthetic medical image augmentation for increased cnn
  performance in liver lesion classification.
\newblock \emph{Neurocomputing}, 321:\penalty0 321--331, 2018.

\bibitem[Genevay et~al.(2018)Genevay, Peyre, and Cuturi]{sinkhorn}
Aude Genevay, Gabriel Peyre, and Marco Cuturi.
\newblock Learning generative models with sinkhorn divergences.
\newblock In Amos Storkey and Fernando Perez-Cruz, editors, \emph{Proceedings
  of the Twenty-First International Conference on Artificial Intelligence and
  Statistics}, volume~84 of \emph{Proceedings of Machine Learning Research},
  pages 1608--1617, Playa Blanca, Lanzarote, Canary Islands, 09--11 Apr 2018.
  PMLR.
\newblock URL \url{http://proceedings.mlr.press/v84/genevay18a.html}.

\bibitem[Ghosh et~al.(2018)Ghosh, Kulharia, Namboodiri, Torr, and
  Dokania]{madgan}
Arnab Ghosh, Viveka Kulharia, Vinay~P. Namboodiri, Philip~H.S. Torr, and
  Puneet~K. Dokania.
\newblock Multi-agent diverse generative adversarial networks.
\newblock In \emph{The IEEE Conference on Computer Vision and Pattern
  Recognition (CVPR)}, June 2018.

\bibitem[{Goodfellow} et~al.(2014){Goodfellow}, {Pouget-Abadie}, {Mirza}, {Xu},
  {Warde-Farley}, {Ozair}, {Courville}, and {Bengio}]{ganGoodfellow}
I.~J. {Goodfellow}, J.~{Pouget-Abadie}, M.~{Mirza}, B.~{Xu}, D.~{Warde-Farley},
  S.~{Ozair}, A.~{Courville}, and Y.~{Bengio}.
\newblock {Generative Adversarial Networks}.
\newblock \emph{ArXiv e-prints}, June 2014.

\bibitem[Gulrajani et~al.(2017)Gulrajani, Ahmed, Arjovsky, Dumoulin, and
  Courville]{gularajani}
Ishaan Gulrajani, Faruk Ahmed, Mart{\'{\i}}n Arjovsky, Vincent Dumoulin, and
  Aaron~C. Courville.
\newblock Improved training of wasserstein gans.
\newblock \emph{CoRR}, abs/1704.00028, 2017.

\bibitem[Heusel et~al.(2017)Heusel, Ramsauer, Unterthiner, Nessler, and
  Hochreiter]{ttur}
Martin Heusel, Hubert Ramsauer, Thomas Unterthiner, Bernhard Nessler, and Sepp
  Hochreiter.
\newblock Gans trained by a two time-scale update rule converge to a local nash
  equilibrium.
\newblock In I.~Guyon, U.~V. Luxburg, S.~Bengio, H.~Wallach, R.~Fergus,
  S.~Vishwanathan, and R.~Garnett, editors, \emph{Advances in Neural
  Information Processing Systems 30}, pages 6626--6637. Curran Associates,
  Inc., 2017.
\newblock URL
  \url{http://papers.nips.cc/paper/7240-gans-trained-by-a-two-time-scale-update-rule-converge-to-a-local-nash-equilibrium.pdf}.

\bibitem[Hinton(2010)]{dbn}
Geoffrey Hinton.
\newblock \emph{Deep Belief Nets}, pages 267--269.
\newblock Springer US, Boston, MA, 2010.
\newblock ISBN 978-0-387-30164-8.
\newblock \doi{10.1007/978-0-387-30164-8_208}.
\newblock URL \url{https://doi.org/10.1007/978-0-387-30164-8_208}.

\bibitem[Hinton et~al.(2006)Hinton, Osindero, and Teh]{hinton2006}
Geoffrey~E Hinton, Simon Osindero, and Yee-Whye Teh.
\newblock A fast learning algorithm for deep belief nets.
\newblock \emph{Neural computation}, 18\penalty0 (7):\penalty0 1527--1554,
  2006.

\bibitem[{Hitawala}(2018)]{hitawala}
S.~{Hitawala}.
\newblock {Comparative Study on Generative Adversarial Networks}.
\newblock \emph{ArXiv e-prints}, January 2018.

\bibitem[Jaiswal et~al.(2019)Jaiswal, AbdAlmageed, Wu, and Natarajan]{bigan}
Ayush Jaiswal, Wael AbdAlmageed, Yue Wu, and Premkumar Natarajan.
\newblock Bidirectional conditional generative adversarial networks.
\newblock In C.~V. Jawahar, Hongdong Li, Greg Mori, and Konrad Schindler,
  editors, \emph{Computer Vision -- ACCV 2018}, pages 216--232, Cham, 2019.
  Springer International Publishing.

\bibitem[Jayathilaka(2018)]{loss_curves}
Mirantha Jayathilaka.
\newblock {Understanding and optimizing GANs (Going back to first principles)}.
\newblock
  \url{https://towardsdatascience.com/understanding-and-optimizing-gans-going-back-to-first-principles-e5df8835ae18},
  2018.

\bibitem[Karras et~al.(2017)Karras, Aila, Laine, and Lehtinen]{progressive}
Tero Karras, Timo Aila, Samuli Laine, and Jaakko Lehtinen.
\newblock Progressive growing of gans for improved quality, stability, and
  variation.
\newblock \emph{arXiv preprint arXiv:1710.10196}, 2017.

\bibitem[Kim et~al.(2017)Kim, Cha, Kim, Lee, and Kim]{discogan}
Taeksoo Kim, Moonsu Cha, Hyunsoo Kim, Jung~Kwon Lee, and Jiwon Kim.
\newblock Learning to discover cross-domain relations with generative
  adversarial networks.
\newblock In Doina Precup and Yee~Whye Teh, editors, \emph{Proceedings of the
  34th International Conference on Machine Learning}, volume~70 of
  \emph{Proceedings of Machine Learning Research}, pages 1857--1865,
  International Convention Centre, Sydney, Australia, 06--11 Aug 2017. PMLR.
\newblock URL \url{http://proceedings.mlr.press/v70/kim17a.html}.

\bibitem[Kim et~al.(2018)Kim, Kim, and Kim]{memory}
Youngjin Kim, Minjung Kim, and Gunhee Kim.
\newblock Memorization precedes generation: Learning unsupervised gans with
  memory networks.
\newblock In \emph{6th International Conference on Learning Representations,
  {ICLR} 2018, Vancouver, BC, Canada, April 30 - May 3, 2018, Conference Track
  Proceedings}, 2018.
\newblock URL \url{https://openreview.net/forum?id=rkO3uTkAZ}.

\bibitem[Kingma and Ba(2015)]{adam}
Diederik~P. Kingma and Jimmy Ba.
\newblock Adam: A method for stochastic optimization.
\newblock \emph{CoRR}, abs/1412.6980, 2015.

\bibitem[Kingma and Welling(2014)]{kingma2014}
Diederik~P Kingma and Max Welling.
\newblock Stochastic gradient vb and the variational auto-encoder.
\newblock In \emph{Second International Conference on Learning Representations,
  ICLR}, 2014.

\bibitem[Kodali et~al.(2017)Kodali, Abernethy, Hays, and Kira]{DRAGAN}
Naveen Kodali, Jacob~D. Abernethy, James Hays, and Zsolt Kira.
\newblock How to train your {DRAGAN}.
\newblock \emph{CoRR}, abs/1705.07215, 2017.

\bibitem[Ledig et~al.(2017)Ledig, Theis, Husz{\'a}r, Caballero, Cunningham,
  Acosta, Aitken, Tejani, Totz, Wang, et~al.]{srgan}
Christian Ledig, Lucas Theis, Ferenc Husz{\'a}r, Jose Caballero, Andrew
  Cunningham, Alejandro Acosta, Andrew Aitken, Alykhan Tejani, Johannes Totz,
  Zehan Wang, et~al.
\newblock Photo-realistic single image super-resolution using a generative
  adversarial network.
\newblock In \emph{Proceedings of the IEEE conference on computer vision and
  pattern recognition}, pages 4681--4690, 2017.

\bibitem[Li et~al.(2017)Li, Chang, Cheng, Yang, and P{\'o}czos]{mmd-gan}
Chun-Liang Li, Wei-Cheng Chang, Yu~Cheng, Yiming Yang, and Barnab{\'a}s
  P{\'o}czos.
\newblock Mmd gan: Towards deeper understanding of moment matching network.
\newblock In \emph{Advances in Neural Information Processing Systems}, pages
  2203--2213, 2017.

\bibitem[{Li} et~al.(2017){Li}, {Liang}, {Wei}, {Xu}, {Feng}, and
  {Yan}]{perceptual}
J.~{Li}, X.~{Liang}, Y.~{Wei}, T.~{Xu}, J.~{Feng}, and S.~{Yan}.
\newblock Perceptual generative adversarial networks for small object
  detection.
\newblock In \emph{2017 IEEE Conference on Computer Vision and Pattern
  Recognition (CVPR)}, pages 1951--1959, July 2017.
\newblock \doi{10.1109/CVPR.2017.211}.

\bibitem[Li et~al.(2015)Li, Swersky, and Zemel]{gmmn}
Yujia Li, Kevin Swersky, and Rich Zemel.
\newblock Generative moment matching networks.
\newblock In \emph{International Conference on Machine Learning}, pages
  1718--1727, 2015.

\bibitem[Liu et~al.(2018)Liu, GU, and Samaras]{wgan_ts}
Huidong Liu, Xianfeng GU, and Dimitris Samaras.
\newblock A two-step computation of the exact {GAN} {W}asserstein distance.
\newblock In Jennifer Dy and Andreas Krause, editors, \emph{Proceedings of the
  35th International Conference on Machine Learning}, volume~80 of
  \emph{Proceedings of Machine Learning Research}, pages 3159--3168,
  Stockholmsmässan, Stockholm Sweden, 10--15 Jul 2018. PMLR.
\newblock URL \url{http://proceedings.mlr.press/v80/liu18d.html}.

\bibitem[Liu et~al.(2017{\natexlab{a}})Liu, Bousquet, and Chaudhuri]{app_conv}
Shuang Liu, Olivier Bousquet, and Kamalika Chaudhuri.
\newblock Approximation and convergence properties of generative adversarial
  learning.
\newblock In \emph{Proceedings of the 31st International Conference on Neural
  Information Processing Systems}, NIPS'17, pages 5551--5559, USA,
  2017{\natexlab{a}}. Curran Associates Inc.
\newblock ISBN 978-1-5108-6096-4.
\newblock URL \url{http://dl.acm.org/citation.cfm?id=3295222.3295306}.

\bibitem[Liu et~al.(2017{\natexlab{b}})Liu, Qin, Luo, and Wang]{paint}
Yifan Liu, Zengchang Qin, Zhenbo Luo, and Hua Wang.
\newblock Auto-painter: Cartoon image generation from sketch by using
  conditional generative adversarial networks.
\newblock \emph{arXiv preprint arXiv:1705.01908}, 2017{\natexlab{b}}.

\bibitem[Mao et~al.(2016)Mao, Li, Xie, Lau, and Wang]{lsgan}
Xudong Mao, Qing Li, Haoran Xie, Raymond Y.~K. Lau, and Zhen Wang.
\newblock Multi-class generative adversarial networks with the {L2} loss
  function.
\newblock \emph{CoRR}, abs/1611.04076, 2016.

\bibitem[Mescheder et~al.(2018)Mescheder, Nowozin, and Geiger]{mescheder18icml}
Lars Mescheder, Sebastian Nowozin, and Andreas Geiger.
\newblock Which training methods for gans do actually converge?
\newblock In \emph{International Conference on Machine Learning (ICML)}, 2018.

\bibitem[Mescheder et~al.(2017)Mescheder, Nowozin, and Geiger]{geiger06}
Lars~M. Mescheder, Sebastian Nowozin, and Andreas Geiger.
\newblock The numerics of gans.
\newblock \emph{CoRR}, abs/1705.10461, 2017.

\bibitem[Metz et~al.(2016)Metz, Poole, Pfau, and Sohl{-}Dickstein]{unrolled}
Luke Metz, Ben Poole, David Pfau, and Jascha Sohl{-}Dickstein.
\newblock Unrolled generative adversarial networks.
\newblock \emph{CoRR}, abs/1611.02163, 2016.
\newblock URL \url{http://arxiv.org/abs/1611.02163}.

\bibitem[Miyato et~al.(2018)Miyato, Kataoka, Koyama, and Yoshida]{spectral}
Takeru Miyato, Toshiki Kataoka, Masanori Koyama, and Yuichi Yoshida.
\newblock Spectral normalization for generative adversarial networks.
\newblock In \emph{International Conference on Learning Representations}, 2018.
\newblock URL \url{https://openreview.net/forum?id=B1QRgziT-}.

\bibitem[{Mohamed} and {Lakshminarayanan}(2016)]{review}
S.~{Mohamed} and B.~{Lakshminarayanan}.
\newblock {Learning in Implicit Generative Models}.
\newblock \emph{ArXiv e-prints}, October 2016.

\bibitem[Mroueh and Sercu(2017)]{fisher}
Youssef Mroueh and Tom Sercu.
\newblock Fisher gan.
\newblock In I.~Guyon, U.~V. Luxburg, S.~Bengio, H.~Wallach, R.~Fergus,
  S.~Vishwanathan, and R.~Garnett, editors, \emph{Advances in Neural
  Information Processing Systems 30}, pages 2513--2523. Curran Associates,
  Inc., 2017.
\newblock URL \url{http://papers.nips.cc/paper/6845-fisher-gan.pdf}.

\bibitem[Nagarajan and Kolter(2017)]{CMU17}
Vaishnavh Nagarajan and J.~Zico Kolter.
\newblock Gradient descent {GAN} optimization is locally stable.
\newblock \emph{CoRR}, abs/1706.04156, 2017.
\newblock URL \url{http://arxiv.org/abs/1706.04156}.

\bibitem[Neumann(1928)]{neumann1928}
J~v Neumann.
\newblock Zur theorie der gesellschaftsspiele.
\newblock \emph{Mathematische annalen}, 100\penalty0 (1):\penalty0 295--320,
  1928.

\bibitem[Nguyen et~al.(2017)Nguyen, Le, Vu, and Phung]{d2gan}
Tu~Dinh Nguyen, Trung Le, Hung Vu, and Dinh Phung.
\newblock Dual discriminator generative adversarial nets.
\newblock In \emph{Proceedings of the 31st International Conference on Neural
  Information Processing Systems}, NIPS'17, pages 2667--2677, USA, 2017. Curran
  Associates Inc.
\newblock ISBN 978-1-5108-6096-4.
\newblock URL \url{http://dl.acm.org/citation.cfm?id=3294996.3295027}.

\bibitem[{Nowozin} et~al.(2016){Nowozin}, {Cseke}, and {Tomioka}]{fgan}
S.~{Nowozin}, B.~{Cseke}, and R.~{Tomioka}.
\newblock {f-GAN: Training Generative Neural Samplers using Variational
  Divergence Minimization}.
\newblock \emph{ArXiv e-prints}, June 2016.

\bibitem[Petzka et~al.(2018)Petzka, Fischer, and Lukovnikov]{reg_wgan}
Henning Petzka, Asja Fischer, and Denis Lukovnikov.
\newblock On the regularization of wasserstein {GAN}s.
\newblock In \emph{International Conference on Learning Representations}, 2018.
\newblock URL \url{https://openreview.net/forum?id=B1hYRMbCW}.

\bibitem[Qi(2017)]{ls-gan}
Guo{-}Jun Qi.
\newblock Loss-sensitive generative adversarial networks on lipschitz
  densities.
\newblock \emph{CoRR}, abs/1701.06264, 2017.

\bibitem[Radford et~al.(2015)Radford, Metz, and Chintala]{dcgan}
Alec Radford, Luke Metz, and Soumith Chintala.
\newblock Unsupervised representation learning with deep convolutional
  generative adversarial networks.
\newblock \emph{arXiv preprint arXiv:1511.06434}, 2015.

\bibitem[Rosca et~al.(2017)Rosca, Lakshminarayanan, Warde-Farley, and
  Mohamed]{variational}
Mihaela Rosca, Balaji Lakshminarayanan, David Warde-Farley, and Shakir Mohamed.
\newblock Variational approaches for auto-encoding generative adversarial
  networks.
\newblock \emph{ArXiv}, abs/1706.04987, 2017.

\bibitem[Roth et~al.(2017)Roth, Lucchi, Nowozin, and Hofmann]{fgan_2}
Kevin Roth, Aur{\'{e}}lien Lucchi, Sebastian Nowozin, and Thomas Hofmann.
\newblock Stabilizing training of generative adversarial networks through
  regularization.
\newblock \emph{CoRR}, abs/1705.09367, 2017.

\bibitem[Salimans et~al.(2016)Salimans, Goodfellow, Zaremba, Cheung, Radford,
  and Chen]{goodfellow16}
Tim Salimans, Ian~J. Goodfellow, Wojciech Zaremba, Vicki Cheung, Alec Radford,
  and Xi~Chen.
\newblock Improved techniques for training gans.
\newblock \emph{CoRR}, abs/1606.03498, 2016.

\bibitem[Salimans et~al.(2018)Salimans, Zhang, Radford, and Metaxas]{ot}
Tim Salimans, Han Zhang, Alec Radford, and Dimitris Metaxas.
\newblock Improving {GAN}s using optimal transport.
\newblock In \emph{International Conference on Learning Representations}, 2018.
\newblock URL \url{https://openreview.net/forum?id=rkQkBnJAb}.

\bibitem[Srivastava et~al.(2017)Srivastava, Valkov, Russell, Gutmann, and
  Sutton]{veegan}
Akash Srivastava, Lazar Valkov, C.~Bradley Russell, Michael~U. Gutmann, and
  Charles~A. Sutton.
\newblock Veegan: Reducing mode collapse in gans using implicit variational
  learning.
\newblock In \emph{NIPS}, 2017.

\bibitem[Subramanian et~al.(2017)Subramanian, Rajeswar, Dutil, Pal, and
  Courville]{advnl}
Sandeep Subramanian, Sai Rajeswar, Francis Dutil, Christopher~Joseph Pal, and
  Aaron~C. Courville.
\newblock Adversarial generation of natural language.
\newblock In \emph{Rep4NLP@ACL}, 2017.

\bibitem[Tolstikhin et~al.(2017)Tolstikhin, Gelly, Bousquet, Simon-Gabriel, and
  Sch{\"o}lkopf]{adagan}
Ilya~O Tolstikhin, Sylvain Gelly, Olivier Bousquet, Carl-Johann Simon-Gabriel,
  and Bernhard Sch{\"o}lkopf.
\newblock Adagan: Boosting generative models.
\newblock In \emph{Advances in Neural Information Processing Systems}, pages
  5424--5433, 2017.

\bibitem[Unterthiner et~al.(2018)Unterthiner, Nessler, Seward, Klambauer,
  Heusel, Ramsauer, and Hochreiter]{coulomb}
Thomas Unterthiner, Bernhard Nessler, Calvin Seward, Günter Klambauer, Martin
  Heusel, Hubert Ramsauer, and Sepp Hochreiter.
\newblock Coulomb {GAN}s: Provably optimal nash equilibria via potential
  fields.
\newblock In \emph{International Conference on Learning Representations}, 2018.
\newblock URL \url{https://openreview.net/forum?id=SkVqXOxCb}.

\bibitem[Wang et~al.(2019)Wang, Xu, Yao, and Tao]{egan}
Chaoyue Wang, Chang Xu, Xin Yao, and Dacheng Tao.
\newblock Evolutionary generative adversarial networks.
\newblock \emph{IEEE Transactions on Evolutionary Computation}, 2019.

\bibitem[Wei et~al.(2018)Wei, Liu, Wang, and Gong]{imp_wgan_gp}
Xiang Wei, Zixia Liu, Liqiang Wang, and Boqing Gong.
\newblock Improving the improved training of wasserstein {GAN}s.
\newblock In \emph{International Conference on Learning Representations}, 2018.
\newblock URL \url{https://openreview.net/forum?id=SJx9GQb0-}.

\bibitem[Wu et~al.(2017)Wu, Zheng, Zhang, and Huang]{imageblend}
Huikai Wu, Shuai Zheng, Junge Zhang, and Kaiqi Huang.
\newblock Gp-gan: Towards realistic high-resolution image blending.
\newblock 03 2017.

\bibitem[Yoo et~al.(2016{\natexlab{a}})Yoo, Kim, Park, Paek, and
  Kweon]{pix2pix}
Donggeun Yoo, Namil Kim, Sunggyun Park, Anthony~S. Paek, and In~So Kweon.
\newblock Pixel-level domain transfer.
\newblock In Bastian Leibe, Jiri Matas, Nicu Sebe, and Max Welling, editors,
  \emph{Computer Vision -- ECCV 2016}, pages 517--532, Cham,
  2016{\natexlab{a}}. Springer International Publishing.

\bibitem[Yoo et~al.(2016{\natexlab{b}})Yoo, Kim, Park, Paek, and
  Kweon]{pixeldtgan}
Donggeun Yoo, Namil Kim, Sunggyun Park, Anthony~S. Paek, and In~So Kweon.
\newblock Pixel-level domain transfer.
\newblock In Bastian Leibe, Jiri Matas, Nicu Sebe, and Max Welling, editors,
  \emph{Computer Vision -- ECCV 2016}, pages 517--532, Cham,
  2016{\natexlab{b}}. Springer International Publishing.

\bibitem[Yu et~al.(2017)Yu, Zhang, Wang, and Yu]{relgan}
L~Yu, W~Zhang, J~Wang, and Y~Yu.
\newblock Seqgan: sequence generative adversarial nets with policy gradient.
\newblock 08 2017.

\bibitem[Yu et~al.(2016)Yu, Zhang, Wang, and Yu]{seqgan}
Lantao Yu, Weinan Zhang, Jun Wang, and Yong Yu.
\newblock Seqgan: Sequence generative adversarial nets with policy gradient.
\newblock \emph{CoRR}, abs/1609.05473, 2016.
\newblock URL \url{http://arxiv.org/abs/1609.05473}.

\bibitem[Zhao et~al.(2016)Zhao, Mathieu, and LeCun]{eb-gan}
Junbo~Jake Zhao, Micha{\"{e}}l Mathieu, and Yann LeCun.
\newblock Energy-based generative adversarial network.
\newblock \emph{CoRR}, abs/1609.03126, 2016.

\bibitem[Zhu et~al.(2017)Zhu, Park, Isola, and Efros]{cyclegan}
Jun-Yan Zhu, Taesung Park, Phillip Isola, and Alexei~A Efros.
\newblock Unpaired image-to-image translation using cycle-consistent
  adversarial networks.
\newblock In \emph{Computer Vision (ICCV), 2017 IEEE International Conference
  on}, 2017.

\end{thebibliography}

\end{document}